\documentclass[10pt,twocolumn,letterpaper]{article}

\usepackage[pagenumbers]{cvpr}

\usepackage{cuted} 

\usepackage{currfile} 

\usepackage{caption} 

\ifdefined\XeTeXversion
    \usepackage[T1]{fontenc}
\fi

\usepackage{tabularx}
\usepackage{float}
\usepackage{colortbl}
\usepackage{pifont}
\usepackage{multirow}
\usepackage{tikz}
\usetikzlibrary{
    arrows.meta,
    positioning,
    fit,
    calc,
    backgrounds
}
\usepackage{listings}
\usepackage{enumitem}
\usepackage{xurl}

\definecolor{NWBlue}{HTML}{2A78D6}
\definecolor{NWOrange}{HTML}{EB6834}
\definecolor{NWGreen}{HTML}{1BAF7A}
\definecolor{NWGray}{RGB}{110,110,110}

\definecolor{CapGreen}{HTML}{2BAA79}
\definecolor{CapRed}{HTML}{E39AA2}
\definecolor{CapAmber}{HTML}{C99543}
\definecolor{CapOurs}{HTML}{F1F3F5}

\definecolor{cvprblue}{rgb}{0.21,0.49,0.74}

\usepackage[
    pagebackref,
    breaklinks,
    colorlinks,
    allcolors=cvprblue
]{hyperref}

\hypersetup{
    pdftitle={NeoWorld-3: From Pixels to Code-Defined Physical Worlds},
    pdfauthor={}
}

\lstdefinestyle{spec}{
    basicstyle=\ttfamily\footnotesize,
    keywordstyle=\bfseries\color{NWBlue},
    commentstyle=\itshape\color{NWGray},
    showstringspaces=false,
    columns=fullflexible,
    keepspaces=true,
    frame=single,
    framerule=0.3pt,
    rulecolor=\color{black!25},
    xleftmargin=4pt,
    xrightmargin=4pt,
    breaklines=true,
}
\newcommand{\model}{NeoWorld-3}

\newcommand{\SDK}{NeoSDK}
\newcommand{\MCP}{NeoMCP}

\newcommand{\code}[1]{\texttt{#1}}

\def\confName{CVPR}
\def\confYear{2027}

\expandafter\ifx\csname iflatexml\endcsname\relax
  \expandafter\newif\csname iflatexml\endcsname
\fi

\newcommand{\projectlinks}[1]{%
Code: \url{https://github.com/NeoWorldProject/NeoWorldStudio}#1
Blog: \url{https://neoworldproject.github.io/Studio}#1
Dataset:
\href{https://huggingface.co/datasets/Doxi-AI/NeoWorld-10K}{Hugging Face}
\quad
\href{https://modelscope.cn/datasets/Doxi-AI/NeoWorld-10K}{ModelScope}%
}

\title{
    NeoWorld-3: From Pixels to Code-Defined Physical Worlds
}

\author{
\textbf{Eachen~Soong}$^{1,2}$
\qquad
\textbf{Zijian~Niu}$^{1,2}$
\qquad
\textbf{Haoran~Xu}$^{1,2}$
\qquad
\textbf{Haoan~Yu}$^{1}$
\\
\textbf{Yihong~Dong}$^{1}$\thanks{Corresponding authors.}
\qquad
\textbf{Yunbo~Wang}$^{1,2*}$
\qquad
\textbf{Xiaokang Yang}$^{1}$
\\
$^{1}$Shanghai Jiao Tong University
\quad
$^{2}$Doxi AI
\iflatexml\else
\\
\projectlinks{\\}
\fi
}

\begin{document}

\maketitle

\iflatexml
\begin{center}
\projectlinks{\quad}
\end{center}
\fi


\begin{strip}
    \vspace{-45pt}
    \centering
    \includegraphics[width=\linewidth]{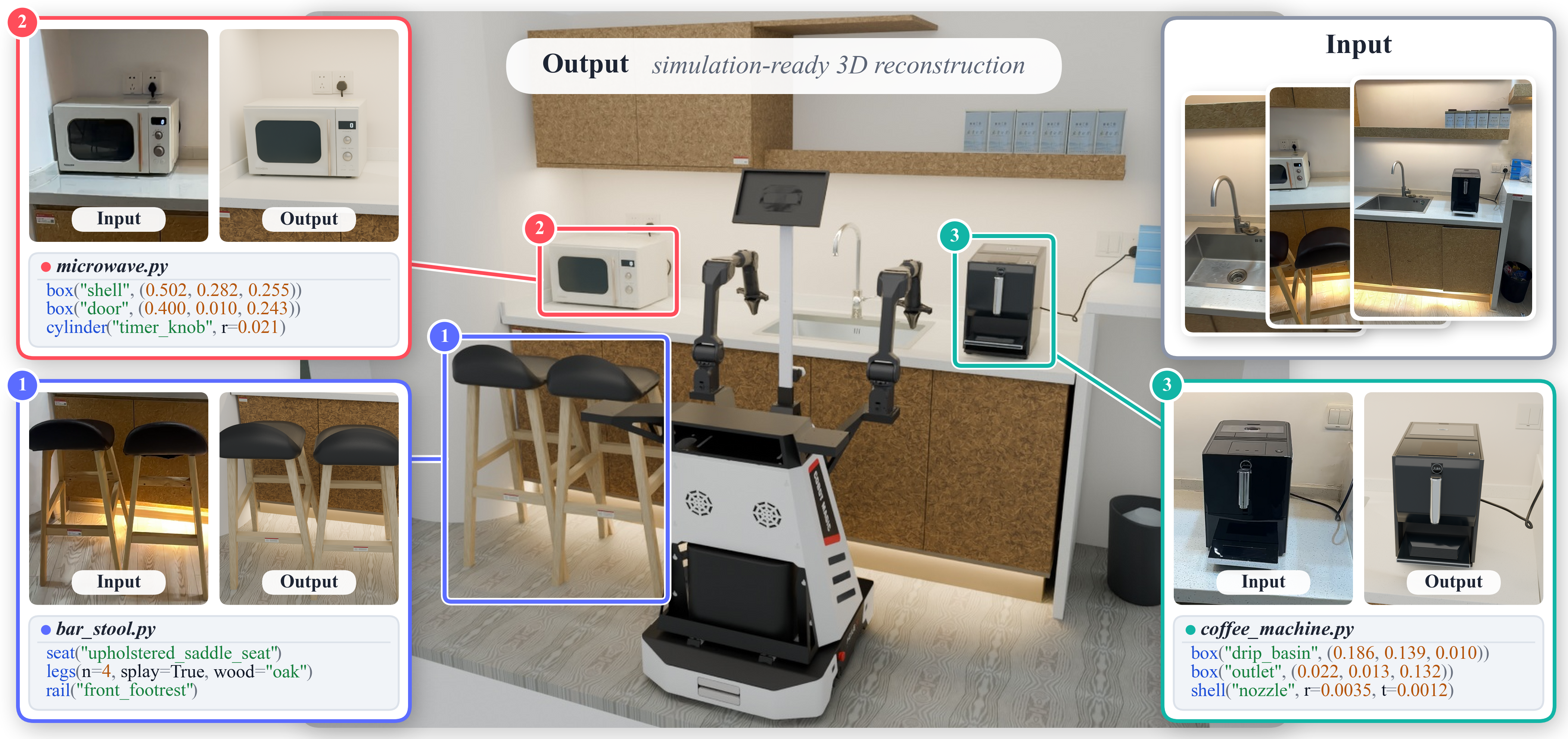}
    \captionsetup{hypcap=false}
    \captionof{figure}{
        \textbf{Code-defined physical worlds with \model{}.}
        Unlike black-box generation or unvalidated code synthesis, \model{} reconstructs executable worlds via two core designs: (1) a \textit{hierarchical scene program tree} for planner-guided, parallel reconstruction of scenes and articulated objects, and (2) a \textit{physics-in-the-loop} workflow of NeoSDK that couples parallel GPU fitting with automated physical repair. The resulting twins power zero-shot (without human demonstrations or teleoperation) Real-Sim-Real robotic manipulation.
    }
    \label{fig:teaser}
    \captionsetup{hypcap=true}
\end{strip}


\begin{abstract}

Existing 3D reconstruction prioritizes visual fidelity over physical interactivity, whereas cutting-edge MLLMs can synthesize geometry programs via tools like Blender yet lack effective physical validation. 
Unlike prior coding agents, \textbf{NeoWorld-3} integrates editable scene reconstruction, physics-in-the-loop optimization, and embodied validation within a unified framework, transforming visual observations into executable physical worlds.
We make two technical contributions. First, we introduce a structured, editable scene representation based on a hierarchical scene program tree, enabling planner-guided decomposition and parallel reconstruction of composed scenes and articulated objects.
Second, we develop \textbf{NeoSDK}, a physics-in-the-loop optimization workflow that couples parallel GPU fitting of program parameters and object poses with automated geometric and physical repair.
NeoWorld-3 surpasses coding-agent baselines on object geometry benchmarks, raises scene-level DINOv3 similarity on RoboCasa365 from $0.785$ to $0.863$, and boosts \textbf{zero-shot real-robot success} on multiple manipulation tasks from $12.5\%$ to $52.5\%$. We also release \textbf{NeoWorld-10K}, comprising $10$K articulated assets across $912$ categories and $19$ domains.

%


\end{abstract}


\section{Introduction}
\label{sec:intro}

\begin{table*}[t]
\centering
\small
\caption{
\textbf{Capabilities of representative 3D coding agents.}
\textit{Scope} indicates support for objects, scenes, and articulation.
\textit{Agent architecture} describes how agents divide and coordinate modeling work.
\textit{Geometry alignment} concerns matching object shapes and positions to images.
\textit{Mesh integrity} covers closure, connectivity, overlap checks, and repair.
\textit{Joint reliability} describes how collisions are checked as joints move.
\textit{Physics role} describes simulation use for motion, interaction, or reconstruction correction.
}
\label{tab:capabilities}
\vspace{-5pt}
\setlength{\tabcolsep}{5pt}
\renewcommand{\arraystretch}{1.1}
\begin{tabular*}{\textwidth}{
    >{\bfseries}l@{\extracolsep{\fill}}
    l l >{\columncolor{NWBlue!5}}l}
\toprule
 & \multicolumn{1}{c}{\textbf{VIGA}~\citep{viga}}
 & \multicolumn{1}{c}{\textbf{Articraft}~\citep{articraft}}
 & \multicolumn{1}{>{\columncolor{NWBlue!12}}c}{\textcolor{NWBlue!75!black}{\textbf{\model{} (Ours)}}} \\
\midrule
Scope
 & Scenes; no articulation
 & Articulation; no scenes
 & Scenes + articulation \\
Agent architecture
 & Generator--verifier loop
 & One agent per object
 & Planner $\to$ parallel region owners \\
\midrule
Geometry alignment
 & Alignment through code edits
 & Alignment through code edits
 & Batched parameter fitting + pose search \\
Mesh integrity
 & No dedicated closure checks
 & Connectivity + overlaps
 & Detects and repairs open meshes \\
Joint reliability
 & No dedicated joint checks
 & Agent tests at selected poses
 & Collision checks across joint ranges \\
Physics role
 & Dynamic physics simulation
 & Asset simulation
 & Corrects physically invalid placement \\
\bottomrule
\end{tabular*}
\vspace{-5pt}
\end{table*}

Reconstructing simulation-ready 3D worlds from visual observations is a foundational requirement for embodied intelligence. Interactive simulation demands editable geometry, explicit articulation, and physical validity, which black-box generative 3D models~\citep{poole2023dreamfusion,xiang2024trellis,pixal3d} do not directly provide.
While coding agents offer a promising alternative by synthesizing 3D worlds as editable programs~\citep{hu2024scenecraft,huang2024blenderalchemy,articraft,blendermcp}, existing approaches face three core limitations at scene scale.
First, \textit{imprecise numerical values}: autoregressive token prediction lacks the continuous precision needed for tight geometric and pose alignment with visual observations. Second, \textit{uncoordinated scene-level editing}: a single agent reasoning over an entire scene suffers from cross-object interference and lacks a visual arbiter to revert edits that contradict image evidence. Third, \textit{partial physical checks}: validating assets only at rest or sparse poses overlooks open meshes, joint collisions across motion ranges, and physically invalid object placements.

To overcome these bottlenecks, we present \textbf{\model{}}, a unified framework that transforms visual observations into executable physical worlds through editable scene reconstruction, physics-in-the-loop optimization, and embodied validation (Fig.~\ref{fig:teaser}).
Specifically, we organize scene composition, object geometry, and kinematic structure into a hierarchical \textit{scene program tree}, enabling a planner to decompose complex environments into subtrees for parallel reconstruction while reverting edits contradicted by visual observations.
Underneath, \textbf{\SDK{}} provides a \textit{physics-in-the-loop} optimization workflow (exposed via standardized \MCP{} tools) that aligns program parameters and object poses to visual observations through parallel GPU fitting. It then enforces simulation readiness via automated geometric and physical repair, including watertight mesh closure, collision checks across full joint ranges, and correction of physically invalid placements.
%
%
Table~\ref{tab:capabilities} contrasts \model{} with representative 3D coding agents~\citep{viga,articraft} across modeling scope, numerical alignment, and physical verification.

We evaluate \model{} across object, articulated-object, and scene reconstruction benchmarks as well as downstream real-world robotic manipulation.
On RoboCasa365~\citep{robocasa365}, \model{} improves scene-level DINOv3 similarity from $0.785$ to $0.863$ over the strongest coding-agent baseline. On 3DCodeBench, \model{} achieves a mesh closure rate of $99.1\%$ among delivered objects compared with $69.8\%$ for Articraft~\citep{articraft}.
Crucially, in zero-shot real-robot manipulation tasks adapted from RoboDojo~\citep{chen2026robodojo}, task programs developed in our reconstructed twins and executed on the physical robot achieve a $52.5\%$ success rate without task demonstrations, outperforming the generalist VLA baseline $\pi_{0.5}$~\citep{black2025pi05} ($12.5\%$).
%

Our primary contributions are summarized as follows:
\begin{itemize}[leftmargin=*]
  \item \textbf{\textit{Structured and editable scene representation.}} We introduce a scene-program-tree representation that organizes scene composition, object geometry, and articulation into hierarchical, executable, and editable programs. This representation supports planner-guided decomposition and parallel reconstruction of complex scenes.
  \item \textbf{\textit{Physics-in-the-loop program optimization.}} We develop NeoSDK, a unified workflow for optimizing scene programs against visual observations. It combines parallel GPU fitting of geometry parameters and object poses with physical validation and repair, including mesh integrity checks, joint collision checks across motion ranges, and correction of physically invalid object placement.
  \item \textbf{\textit{Real-Sim-Real embodied validation.}} We demonstrate that reconstructed worlds can support zero-shot real-robot manipulation, an aspect underexplored in prior 3D reconstruction methods.
  \item \textbf{\textit{NeoWorld-10K open dataset.}}
  Using \model{}, we release a dataset of 10K articulated 3D assets spanning $912$ categories and $19$ domains, supporting simulation-ready scene construction and embodied AI research.
\end{itemize}

\section{Related work}
\label{sec:related}

3D reconstruction has made significant progress in recent years, yet existing approaches still face important limitations in reconstructing simulation-ready environments. Neural rendering methods, including NeRF~\citep{mildenhall2020nerf} and 3D Gaussian Splatting~\citep{kerbl2023gaussians}, model scene appearance through radiance fields and Gaussian primitives, but do not directly provide geometric structures suitable for physical interaction. Generative 3D reconstruction methods, such as DreamFusion~\citep{poole2023dreamfusion}, TRELLIS~\citep{xiang2024trellis}, SAM 3D~\citep{sam3d}, and Pixal3D~\citep{pixal3d}, leverage learned priors to generate 3D assets, but generally lack the articulated structures required for interactive simulation. Articulated reconstruction methods, including SINGAPO~\citep{singapo}, SPARK~\citep{spark}, PhysX-Anything~\citep{physxanything}, and URDF-Anything~\citep{urdfanything}, further recover part geometry and kinematic structures, but physical validity and simulation stability remain challenging.

Beyond these approaches, programmatic reconstruction is becoming an established direction in 3D reconstruction.
Early LLM- and VLM-based methods, including 3D-GPT~\citep{sun2023threedgpt}, BlenderAlchemy~\citep{huang2024blenderalchemy}, and Real2Code~\citep{real2code}, demonstrated the feasibility of generating and refining 3D programs, although reconstruction precision remained limited compared with dedicated reconstruction pipelines. More recent agentic approaches have substantially advanced this direction through iterative reasoning, tool use, and visual feedback: Articulate-Anything~\citep{articulateanything}, VIGA~\citep{viga}, and Articraft~\citep{articraft} iteratively construct and refine structured 3D assets, while 3D-CoS~\citep{3dcos} further explores code-based reconstruction as a general approach. However, discrete program edits can limit geometric precision, while visual feedback alone cannot guarantee physical validity. To address these limitations, \model{} combines editable program construction with parallel program-space fitting and physics-in-the-loop validation to improve visual and geometric fidelity while producing simulation-ready 3D assets.

\begin{figure*}[t]
\vspace{-10pt}
    \centering
    \includegraphics[width=0.99\textwidth]{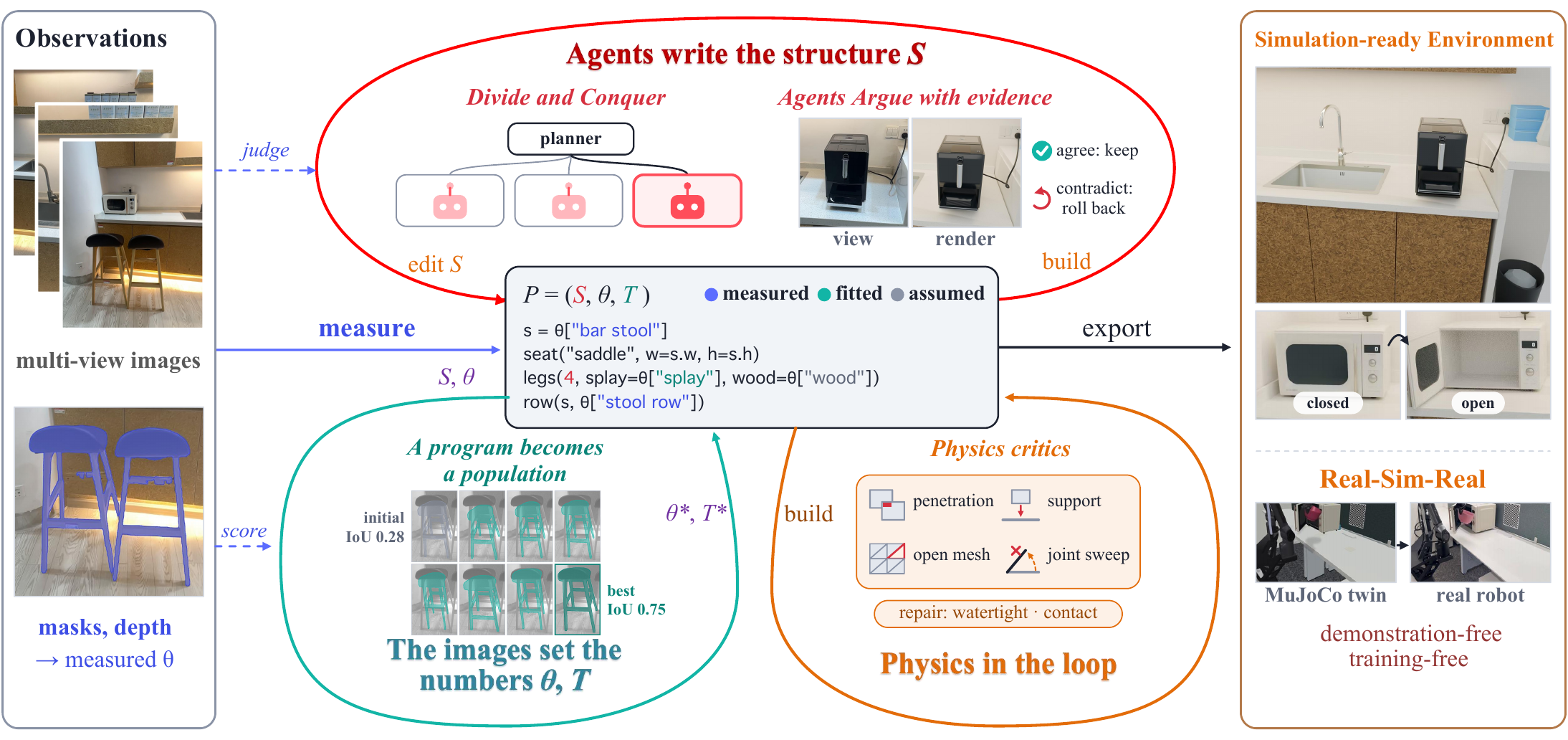}
    \vspace{-8pt}
    \caption{
    \textbf{Overview of \model{}.}
        A scene is a tree of grounded programs $P=(S,\theta,T)$ (center, Sec.~\ref{sec:method-editable-main}), revised in three loops.
        Modeling agents write the structure $S$: a planner divides the tree into regions, one agent per region works in parallel, and the views decide which edits are kept (top, Sec.~\ref{sec:method-reconstruction}).
        \SDK{} optimizes the programs with physics in the loop (Sec.~\ref{sec:method-refinement-main}): parallel GPU fitting sets $\theta$ and $T$ against the views (bottom left), and physics critics find penetrating, unsupported, or open parts and blocked joints, repair what they can, and return the rest to the owning agents (bottom right).
        The fitting panel starts from a perturbed estimate and reports silhouette IoU against the masks in the photograph.
        The exported scene is ready for simulation and Real-Sim-Real robotic tasks (right, Sec.~\ref{sec:method-robotics-main}).
    }
    \vspace{-10pt}
    \label{fig:architecture}
\end{figure*}
\section{NeoWorld-3}
\label{sec:method}



\model{} is a unified framework that transforms visual observations into executable physical worlds by integrating editable scene reconstruction, physics-in-the-loop optimization, and embodied validation (Fig.~\ref{fig:architecture}).
First, we organize scene layout, part-level geometry, and articulation into a hierarchical \emph{scene program tree} of grounded prototypes (Sec.~\ref{sec:method-editable-main}).
Second, a hierarchical agentic harness coordinates modeling agents across subtrees of this program tree via standardized \MCP{} tools, enforcing parallel reconstruction with observation-grounded rollback (Sec.~\ref{sec:method-reconstruction}).
Third, \SDK{} executes a \emph{physics-in-the-loop} optimization workflow that aligns continuous parameters and object poses through parallel GPU fitting while validating and repairing mesh integrity, full-range joint collisions, and physical object placement (Sec.~\ref{sec:method-refinement-main}).
Finally, the resulting simulation-ready digital twins directly support zero-shot real-sim-real robotic manipulation without human demonstrations or teleoperation data (Sec.~\ref{sec:method-robotics-main}).

\subsection{Structured \& Editable Scene Program Tree}
\label{sec:method-editable-main}
To unify scene composition, object geometry, and kinematic structure into a single executable and editable representation, \model{} models each environment as a hierarchical \emph{scene program tree}.
The root encodes room-level architecture such as walls and floors; internal nodes group semantically or spatially coupled objects into subtrees, such as a counter run with its appliances; and each leaf node is an editable, grounded prototype $P=(S,\theta,T)$, consisting of a Python construction program $S$, continuous parameters $\theta\in\mathbb R^d$, and a scene pose $T$.
Executing $S$ yields the geometry $\mathcal G=S(\theta)$, which $T$ places into the scene.
Within $S$, parts are connected by explicit joints specifying joint type, axis, and motion range, allowing a single tree to carry scene layout, part geometry, and articulation and export directly to physics simulation (Appendix~\ref{sec:physics}); repeated instances share one program $S$ and parameters $\theta$, differing only in $T$.

While a vision-language agent can infer discrete structure (\textit{e.g.}, that a cabinet has three drawers), metric dimensions such as a $61$,cm width require numerical grounding.
We partition the parameters of each prototype into three:
\[
\theta=(\theta^{\mathrm{meas}},\ \theta^{\mathrm{fit}},\ \theta^{\mathrm{asm}}).
\]
Here, \textbf{\textit{measured values}} ($\theta^{\mathrm{meas}}$) are extracted directly from the observations before modeling: each object is segmented in every view, lifted to 3D using MoGe-3 depth~\citep{kong2026moge3finedetailmonoculargeometry} and camera parameters, and carved using free space visible in other views; repeated instances are grouped into arrays with a measured count and spacing, and surface colors are read from the images.
\textbf{\textit{Fitted values} }($\theta^{\mathrm{fit}}$) are optimized numerically against multi-view observations (Sec.~\ref{sec:method-refinement-main}).
\textbf{\textit{Assumed values}} ($\theta^{\mathrm{asm}}$), such as the occluded interior color of a microwave, retain the agent's semantic priors.
Because $S$ references all parameters by name, the agent defines the structural logic without hard-coding fragile metric values. 

\vspace{-8pt}
\paragraph{Parameter sharing across parts.} 
Part dimensions and relative positions are expressed in terms of $\theta$, so a shared parameter $\theta_k$ controls several related parts and preserves the common dimensions, symmetries, and attachments encoded in $S$.
A measured width therefore determines the widths of all dependent drawer fronts, and a compact set of coupled parameters leaves fewer values to fit and less risk of overfitting limited views.
Revision takes two forms: \emph{structural editing} changes $S$ itself by adding, removing, or reorganizing primitives, Boolean operations, or kinematic parent--child relations (Sec.~\ref{sec:method-reconstruction}), and \emph{fitting} updates the values with $S$ fixed: \emph{parameter fitting} sets $\theta$ on a twin of the program and \emph{pose search} sets $T$ on a twin of the scene (Sec.~\ref{sec:method-refinement-main}).
Agents specify program structure, while image-based fitting estimates the corresponding parameter values.

\subsection{Hierarchical Agentic Harness}
\label{sec:method-reconstruction}

Scene-scale modeling is a sequence of edits in which any modification can invalidate earlier ones. \model{} orchestrates this process through a hierarchical agentic harness over the scene program tree, where agents act through \SDK{} and the standardized tools of \MCP{} (Appendices~\ref{sec:sdk} and~\ref{sec:mcp}).
For each object, a modeling agent synthesizes the program structure from input views, compares rendered views against observations, and revises structure while numerical fitting sets view-constrained parameters.
Three core principles govern the harness.

\vspace{-8pt}
\paragraph{Divide and conquer.}
A global planner inventories objects from visual observations and partitions the scene program tree into disjoint subtrees, grouping repeated objects under shared prototypes and assigning each subtree to a modeling agent.
Each agent observes the full scene to maintain global context but restricts edits to its assigned subtree, requesting cross-boundary adjustments from responsible agents.
Subtree agents execute in parallel while the composed scene is evaluated globally, tracing any regression to the specific object node that caused it (Appendix~\ref{sec:scene}).

\vspace{-8pt}
\paragraph{Evidence-grounded arbitration.}
Because a coding agent may favor its own edits and a modification that improves one viewpoint can degrade another, in \model{}, every edit and objection must cite supporting views: any edit contradicted by multi-view observations is rolled back, regardless of which agent proposed it.
An MLLM critic flags discrepancies (missing objects, wrong categories or orientations, and misplaced faces), each tied to specific views. The owning agent revises the program or cites views refuting the issue, resolving semantic and topological errors before metric fitting so missing parts are never compensated by distorted dimensions.

\vspace{-8pt}
\paragraph{Closed-loop physical feedback.}
A drawer can match every input view and still pass through its cabinet, so physical verification acts as a second critic.
After each build, \SDK{} reports penetrating, unsupported, or open parts and sweeps each declared joint across its motion range (Sec.~\ref{sec:method-refinement-main}).
Findings are routed back to the owning subtree agent with visual and physical evidence to guide the next round of structural or pose refinement.

\begin{figure*}[t]
    \centering
    \vspace{-5pt}
    \includegraphics[width=0.95\textwidth]{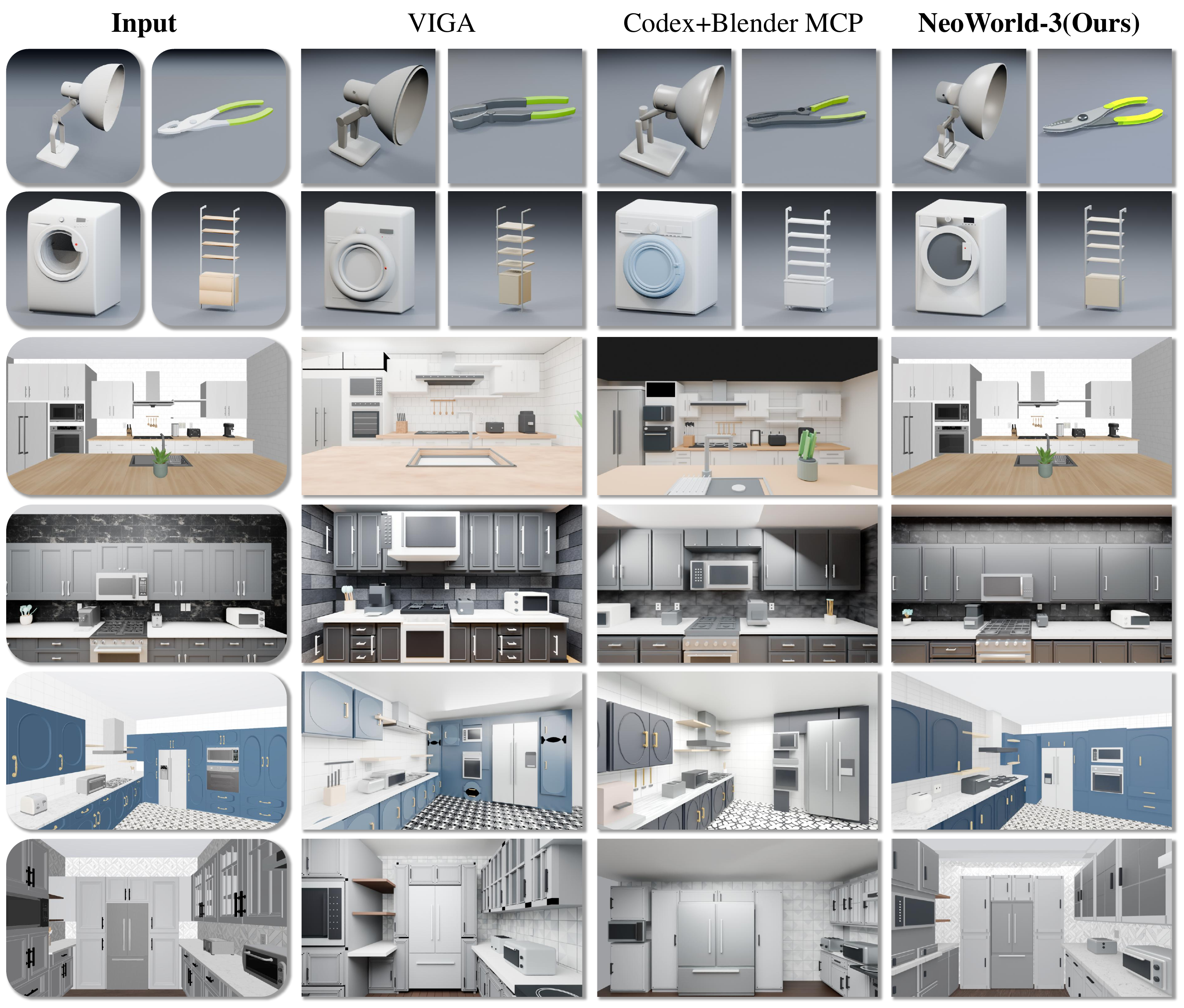}
    \vspace{-10pt}
    \caption{
        \textbf{Qualitative comparison of object- and scene-level reconstructions.}
        We compare \model{} with two baselines: VIGA and Codex+BlenderMCP.
        \model{} achieves higher geometric and visual fidelity at the object level, along with more accurate spatial layouts, object correspondences, proportions, materials, and colors at the scene level.
    }
    \label{fig:comp}
    \vspace{-10pt}
\end{figure*}

\subsection{Program Optimization with \SDK{}}
\label{sec:method-refinement-main}

While agents specify program topology and parameter semantics, \SDK{} optimizes the scene program tree against visual observations with physics in the loop.
After each build, \SDK{} fits continuous values on GPU twins of the programs and scene, validates geometric and physical soundness, repairs solvable defects, and returns remaining issues with evidence to the owning agents.
\MCP{} exposes these operations as callable tools.

\vspace{-8pt}
\paragraph{Parallel GPU fitting.}
Let $\mathcal E_S(\theta)$ denote the tensor geometry produced by batched execution of $S$ on the GPU, let $\mathcal V$ index the views of the observation set $\mathcal O$, and let $\mathcal R_v$ render the geometry under poses $T$ from view $v$.
With $S$, $T$, and the cameras fixed, we seek
\begin{equation}
\theta^\star
=
\arg\min_{\theta\in\mathcal B}
\mathcal L_{\mathrm{obs}}\!\left(
\left\{
\mathcal R_v\!\left(\mathcal E_S(\theta),T\right)
\right\}_{v\in\mathcal V},
\mathcal O
\right),
\label{eq:program-refinement}
\end{equation}
where $\mathcal L_{\mathrm{obs}}$ combines silhouette, contour, and depth evidence.
The program twin replays $S$ for a whole batch of parameter vectors on every fresh build and rasterizes them in one call~\citep{laine2020nvdiffrast}; the cross-entropy method samples the feasible set $\mathcal B$ and refits its sampling distribution to the best candidates of each generation.
A direction $s\in\{-1,+1\}$ of $\theta_i$ is searched only if a step $s\,\varepsilon_i$ changes the visible surface in at least $\tau$ pixels of some view; ranges never cross zero, orientation is estimated before shape, and a change is kept only if the input views support it.
Object poses $T$ are fitted on a twin of the whole scene: after every build, each instance of a repeated object is placed by an exhaustive search over yaw and planar offset, scored against its depth and masks, and an agent may request a local pose fit for an object it owns (Appendix~\ref{sec:object-proposals}).

\begin{table*}[!t]
    \centering
    \small
    \caption{
        \textbf{Comparison of object-level reconstruction on 3DCodeBench.}
        Within this comparison, \model{} obtains the best visual and geometric scores.
        IoU: Framed IoU; 3D--3D: Uni3D point-cloud--point-cloud; Image--3D: image--point-cloud cosine similarities.
    }
    \vspace{-5pt}
    \label{tab:object_comparison}


    \setlength{\tabcolsep}{4pt}
    \begin{tabular}{lcccccccc}
        \toprule
        \textbf{Method}
        & \textbf{Exec (\%) $\uparrow$}
        & \textbf{SigLIP-2 $\uparrow$}
        & \textbf{DINOv3 $\uparrow$}
        & \textbf{IoU $\uparrow$}
        & \textbf{CD $\downarrow$}
        & \textbf{3D--3D $\uparrow$}
        & \textbf{Image--3D $\uparrow$}
        & \textbf{Closed (\%) $\uparrow$} \\
        \midrule

        Direct GPT~\citep{openaigpt}
        & $94.81$ & $0.8252$ & $0.6316$ & $0.4776$
        & $0.0690$ & $0.5550$ & $0.2967$ & $36.3$ \\

        Codex~\citep{codexcli}
        & $96.70$ & $0.8505$ & $0.6502$ & $0.4870$
        & $0.0701$ & $0.5697$ & $0.3096$ & $33.7$ \\

        Codex + BlenderMCP~\citep{codexcli,blendermcp}
        & $98.58$ & $0.8591$ & $0.6457$ & $0.4886$
        & $0.0717$ & $0.5810$ & $0.3115$ & $49.3$ \\

        VIGA~\citep{viga}
        & $98.58$ & $0.8712$ & $0.6930$ & $0.5870$
        & $0.0602$ & $0.5756$ & $0.3072$ & $40.3$ \\

        Articraft~\citep{articraft}
        & $\mathbf{100.00}$ & $0.8409$ & $0.6086$ & $0.4971$
        & $0.0694$ & $0.5616$ & $0.3113$ & $69.8$ \\

        \midrule
        \textbf{\model{} (Ours)}
        & $\mathbf{100.00}$
        & $\mathbf{0.8930}$
        & $\mathbf{0.7139}$
        & $\mathbf{0.7310}$
        & $\mathbf{0.0380}$
        & $\mathbf{0.6215}$
        & $\mathbf{0.3188}$
        & $\mathbf{99.1}$ \\

        \bottomrule
    \end{tabular}%
\end{table*}

\begin{table*}[!t]
    \centering
    \small
    \caption{
        \textbf{Comparison of articulated-object reconstruction on selected PartNet-Mobility assets.}
        Across the reported metrics, \model{} ranks first among the evaluated methods.
        Axis Error is in radians; Pivot Error is in normalized reference units.
    }
    \label{tab:articulation_comparison}
    \vspace{-5pt}
    \setlength{\tabcolsep}{5pt}

    \begin{tabular}{lccccccc}
        \toprule
        \textbf{Method}
        & \textbf{AS-gIoU $\downarrow$}
        & \textbf{AS-CD $\downarrow$}
        & \textbf{AOR $\downarrow$}
        & \textbf{Type (\%) $\uparrow$}
        & \textbf{Axis Error $\downarrow$}
        & \textbf{Pivot Error $\downarrow$}
        & \textbf{Closed (\%) $\uparrow$} \\
        \midrule

        Direct GPT~\citep{openaigpt}
        & $0.8799$
        & $0.2355$
        & $0.0023$
        & $87.50$
        & $0.4036$
        & $0.1734$
        & $76.6$ \\

        Codex~\citep{codexcli}
        & $0.9131$
        & $0.2529$
        & $0.0011$
        & $89.86$
        & $0.3228$
        & $0.1595$
        & $85.7$ \\

        Codex + BlenderMCP~\citep{codexcli,blendermcp}
        & $0.8959$
        & $0.2510$
        & $0.0010$
        & $90.00$
        & $0.2932$
        & $0.2044$
        & $83.1$ \\

        Articraft~\citep{articraft}
        & $0.8732$
        & $0.1890$
        & $0.0001$
        & $89.33$
        & $0.3629$
        & $0.1427$
        & $98.7$ \\

        \midrule
        \textbf{\model{} (Ours)}
        & $\mathbf{0.7773}$
        & $\mathbf{0.1480}$
        & $\mathbf{0.0000}$
        & $\mathbf{91.45}$
        & $\mathbf{0.2687}$
        & $\mathbf{0.0996}$
        & $\mathbf{100.0}$ \\

        \bottomrule
    \end{tabular}%
    \vspace{-5pt}
\end{table*}

\vspace{-8pt}
\paragraph{Physical validation and repair.}
A drawer can match every view and still pass through its cabinet, so after each build \SDK{} checks three properties.
\emph{Mesh integrity:} each link mesh must be a closed, consistently oriented manifold, as solid collision and volume-based physical properties require; open meshes are repaired by default before export, keeping moving links separate.
\emph{Joint collisions across motion ranges:} each declared joint is swept over its range, and the sampled states are checked for link collisions and detachment; only joints that no sweep refutes are exported (Appendix~\ref{sec:object-articulation}).
\emph{Object placement:} penetrating or unsupported objects receive bounded minimum-motion corrections, admitted only if the views still support them (Appendix~\ref{sec:scene-physics}).
Exported kinematics are compared with native playback (Appendix~\ref{sec:physics}), and findings that repair cannot resolve go to the owning agent with their evidence.

\subsection{Real-Sim-Real Robotic Manipulation}
\label{sec:method-robotics-main}

We leverage \model{} as the core of a real-sim-real robotic manipulation pipeline, and validate the practical utility of reconstructed worlds on a physical robot, which is under-explored in prior 3D reconstruction work.
\model{} constructs a MuJoCo~\citep{todorov2012mujoco} twin from workspace observations, synthesizes and refines manipulation policies in simulation, and uses contact feedback during physical execution (see Appendix~\ref{sec:robotics} for details).

\vspace{-8pt}
\paragraph{Reality to Simulation.}
A fixed RGB-D camera and wrist-mounted cameras observe the workspace, from which \model{} reconstructs the editable scene twin as described above.
Hand-eye calibration via a calibration board registers the twin's metric frame to the robot arms, and each movable body is realigned using a fresh wrist scan immediately prior to manipulation (Appendix~\ref{sec:robotics-real}).

\vspace{-8pt}
\paragraph{Simulation-based policy RSI.}
Within the reconstructed twin, a coding agent writes each task as a state machine over modular control primitives, where inverse kinematics computes grasps and trajectories that are checked for collisions.
Before deployment, the agent iteratively refines each state machine in the twin by rolling out episodes and revising failure modes using recorded execution traces.

\vspace{-8pt}
\paragraph{Policy deployment.}
The refined state machines then execute on the physical robot after brief real-robot refinement (Appendix~\ref{sec:robotics-eval}): a pre-grasp wrist scan realigns the target pose, and contact force estimated from motor currents enables articulated parts such as doors to move under compliant control without requiring an exact analytical mechanism model (Appendix~\ref{sec:robotics-compliant}).
Policy synthesis and execution are strictly zero-shot, which requires no human demonstrations, teleoperation data, or task-specific training.
The evaluation layouts in Section~\ref{sec:val_rsr} remain unseen during development.

\section{Experiments}
\label{sec:evaluation}

\begin{table*}[t]
    \centering
    \small
    \caption{
        \textbf{Comparison of scene reconstruction on RoboCasa365}
        (scenes $02$-$10$).
        Among the compared methods, the full model achieves the best placement, visual, and MLLM scores, with $100.0\%$ watertight surface area.
        OCD is measured in centimeters (cm).
        Closed is the surface-area-weighted closure percentage (\%), averaged across scenes.
    }
    \vspace{-5pt}
    \label{tab:scene_results}

    \setlength{\tabcolsep}{2pt}

    \begin{tabular*}{\textwidth}{@{\extracolsep{\fill}}lcccccccc@{}}
        \toprule
        \textbf{Method}
        & \textbf{Obj@$25\%$ $\uparrow$}
        & \textbf{Pair@$10$cm $\uparrow$}
        & \textbf{OCD $\downarrow$}
        & \textbf{DINOv3 $\uparrow$}
        & \textbf{Complete $\uparrow$}
        & \textbf{Layout $\uparrow$}
        & \textbf{Function $\uparrow$}
        & \textbf{Closed $\uparrow$} \\
        \midrule

        \multicolumn{9}{@{}l}{\textit{External agents and methods}} \\

        Direct GPT~\citep{openaigpt}
        & $0.118$ & $0.007$ & $109.1$ & $0.662$
        & $3.37$ & $3.30$ & $3.56$ & $98.3$ \\

        VIGA~\citep{viga}
        & $0.288$ & $0.018$ & $42.5$ & $0.785$
        & $4.48$ & $4.81$ & $4.59$ & $91.5$ \\

        \midrule
        \multicolumn{9}{@{}l}{\textit{Tool interface: BlenderMCP vs.\ \MCP{}}} \\
        Codex~\citep{codexcli}
        & $0.270$ & $0.012$ & $50.6$ & $0.743$
        & $4.15$ & $4.04$ & $4.19$ & $82.5$ \\

        Codex + BlenderMCP~\citep{codexcli,blendermcp}
        & $0.197$ & $0.005$ & $56.9$ & $0.761$
        & $4.22$ & $4.07$ & $4.70$ & $99.9$ \\
        
        Codex + NeoMCP~\citep{codexcli}
        & $0.412$ & $0.069$ & $27.8$ & $0.846$
        & $4.74$ & $5.19$ & $5.37$ & $96.2$ \\
        
        \midrule
        \multicolumn{9}{@{}l}{\textit{Divide and conquer, fitting and validity}} \\
        \model{}, single scene agent
        & $0.419$ & $0.072$ & $28.0$ & $0.809$
        & $4.30$ & $4.85$ & $4.78$ & $99.1$ \\

        \model{} w/o fitting
        & $0.640$ & $0.233$ & $19.4$ & $0.832$
        & $4.44$ & $4.74$ & $5.07$ & $100.0$ \\

        \model{} w/o repair
        & $0.685$ & $0.284$ & $15.6$ & $0.863$
        & $4.85$ & $5.37$ & $5.78$ & $77.9$ \\

        \textbf{\model{} (full)}
        & $\mathbf{0.685}$
        & $\mathbf{0.284}$
        & $\mathbf{15.6}$
        & $\mathbf{0.863}$
        & $\mathbf{4.85}$
        & $\mathbf{5.37}$
        & $\mathbf{5.78}$
        & $\mathbf{100.0}$ \\

        \bottomrule
    \end{tabular*}
\end{table*}

\subsection{Experimental Setup}
We evaluate \model{} through three main comparisons:
\emph{object reconstruction quality}, \emph{articulation quality},
and \emph{scene quality}.
For object reconstruction, we use $212$ instances from
3DCodeBench~\citep{codebench3d}; for articulated objects,
$77$ selected PartNet-Mobility assets~\citep{sapien} with
SINGAPO annotations~\citep{singapo}; and for scenes,
nine held-out environments from RoboCasa365~\citep{robocasa365}.
We compare \model{} against Direct GPT~\citep{openaigpt}, Codex~\citep{codexcli},
Codex + BlenderMCP~\citep{blendermcp}, VIGA~\citep{viga}, and Articraft~\citep{articraft}.
All methods use the same underlying LLM and input images within each setting.
For real-robot manipulation, we compare with $\pi_{0.5}$~\citep{black2025pi05} and SIMPACT~\citep{liu2026simpact} on four tasks adapted from RoboDojo~\citep{chen2026robodojo}.

\vspace{-8pt}
\paragraph{Evaluation metrics.}
We assess object reconstruction quality using these metrics: execution success (Exec), SigLIP-2~\citep{siglip2}, DINOv3~\citep{dinov3}, Framed IoU, Chamfer distance, and Uni3D similarity~\citep{uni3d}. Articulation quality follows the SINGAPO protocol~\citep{singapo}, using AS-gIoU, AS-CD, average overlapping ratio (AOR)~\citep{cage}, joint-type accuracy, axis error, and pivot error. Scene quality is measured by DINOv3 similarity, MLLM-based completeness (Complete), layout accuracy (Layout), functional plausibility (Function), Object Center Distance (OCD), and placement accuracies (Obj@$25\%$ and Pair@$10$cm). Watertightness is assessed using Closed. Real-robot manipulation is evaluated using successes out of ten layouts per task and mean score. Detailed metric definitions and evaluation protocols are provided in Appendix~\ref{sec:experiment-appendix}.



\subsection{Object-Level Comparisons}
\label{sec:object_comparisons}

\paragraph{Geometric reconstruction quality.}
Table~\ref{tab:object_comparison} compares object reconstruction
performance on 3DCodeBench. \model{} consistently outperforms the
compared baselines across visual and geometric metrics while
achieving a $100\%$ execution rate. Except for Exec, all metrics are
computed over successfully executed assets.
Compared with VIGA, the strongest baseline on both metrics, \model{}
raises Framed IoU from $0.5870$ to $0.7310$ and reduces Chamfer distance
from $0.0602$ to $0.0380$. Higher Uni3D, SigLIP-2, and DINOv3 scores further
indicate greater 3D and visual similarity to the references.
Closed reaches $99.1\%$, versus $69.8\%$ for Articraft, the strongest baseline on this metric.
As shown in Figure~\ref{fig:comp}, \model{} also achieves higher
geometric fidelity and visual quality, more accurately recovering
fine-grained geometric details and overall object shapes.

\vspace{-8pt}
\paragraph{Articulated reconstruction quality.}
Table~\ref{tab:articulation_comparison} evaluates articulated geometry
and joint parameters on $77$ selected PartNet-Mobility assets. \model{} achieves the lowest
AS-gIoU ($0.7773$) and AS-CD ($0.1480$), indicating higher geometric fidelity
across sampled motion states. Joint-type accuracy reaches $91.45\%$
versus $90.00\%$ for Codex + BlenderMCP, with the lowest axis and normalized pivot
errors of $0.2687$\,rad and $0.0996$, indicating better joint-type prediction and localization. Closed reaches $100.0\%$, compared with Articraft's $98.7\%$. AOR ($0.0000$) measures sibling-part
bounding-box overlap at the five protocol states. A supplementary
evaluation tests solid intersections for all link pairs, including
parent--child pairs, over sampled declared-range sweeps and joint
combinations; limit IoU assesses agreement with reference joint ranges
(Appendix~\ref{sec:articulation-penetration}).

\begin{figure*}[t]
\centering
\includegraphics[width=0.99\linewidth]{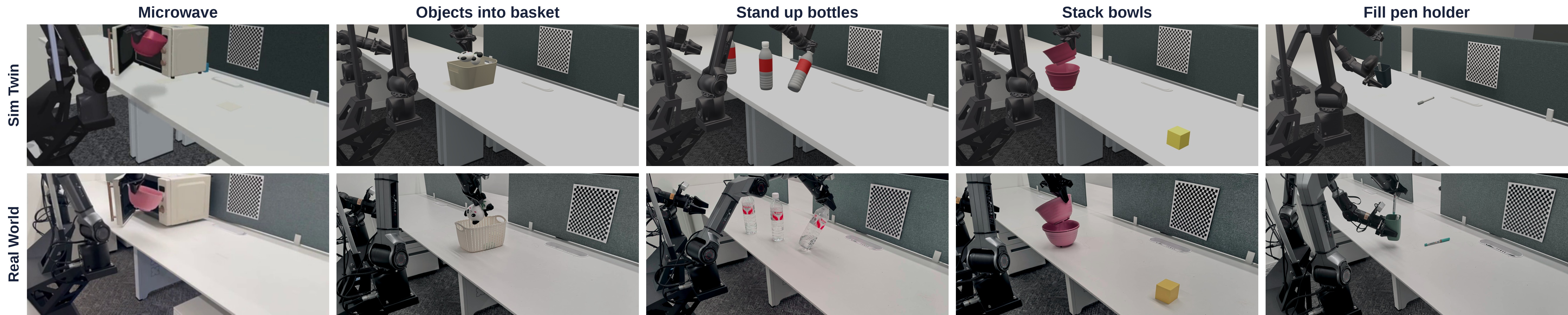}
\vspace{-5pt}
\caption{\textbf{Zero-shot real-sim-real robotic manipulation.} \textbf{Top:} task policies optimized inside the reconstructed digital twin. \textbf{Bottom:} matching execution stages transferred zero-shot to the physical dual-arm robot without human demonstrations or teleoperation data
}
\vspace{-5pt}
\label{fig:real-sim-robotics}
\end{figure*}

\subsection{Scene-Level Comparisons}
\label{sec:scene_comparisons}

Table~\ref{tab:scene_results} reports scene reconstruction results on RoboCasa365 scenes $02$--$10$. Guided by the hierarchical scene program tree, \model{} places objects substantially more accurately than all external baselines: it places $68.5\%$ of objects within $25\%$ of their true size (Obj@25\%) and $28.4\%$ of neighboring pairs within $10$,cm of their ground-truth offset (Pair@10cm), compared with $28.8\%$ and $1.8\%$ for VIGA~\citep{viga}, the strongest external baseline on placement, while reducing OCD from $42.5$ to $15.6$,cm. \model{} also achieves the highest DINOv3 and MLLM scores, outperforming Direct GPT and Codex. As shown in Figure~\ref{fig:comp}, our framework recovers more accurate spatial layouts and object correspondences alongside higher-fidelity object proportions, materials, and colors, confirming its effectiveness on complex compositional scenes.

\vspace{-8pt}
\paragraph{Ablation studies on \SDK{} and \MCP{}.}
A single Direct GPT call performs similarly to multiple Codex calls on objects, while Codex's more accurate scene placement suggests that the agent framework may benefit scene reconstruction.
With the agent unchanged, replacing BlenderMCP with \MCP{} raises Obj@25\% from $0.197$ to $0.412$ and DINOv3 from $0.761$ to $0.846$ for Codex.

\vspace{-8pt}
\paragraph{Ablation studies on divide-and-conquer.}
A \emph{single scene agent} with the same tools (Sec.~\ref{sec:method-reconstruction}) reaches Obj@25\% 0.419 and Pair@10cm $0.072$ against $0.685$ and $0.284$; at equal output tokens, round~1 without the critic already reaches $0.580$ and $0.145$ (Table~\ref{tab:scene-decomposition}).

\vspace{-8pt}
\paragraph{Ablation studies on the numerical fitting pipeline.}
Removing fitting (pose search, numeric readings, and parameter fitting; Sec.~\ref{sec:method-refinement-main}) lowers Obj@25\% from $0.685$ to $0.640$ and Pair@10cm from $0.284$ to $0.233$.
Watertight repair raises the closed-area fraction from $77.9\%$ in the \emph{w/o repair} variant to $100.0\%$, without changing reconstruction scores.









\subsection{Real-Sim-Real Robot Manipulation Results}
\label{sec:val_rsr}

As a downstream use of MuJoCo twins of the reconstructed scenes, we compare \model{} with two methods on four tabletop tasks adapted from RoboDojo~\citep{chen2026robodojo}: stacking three bowls, standing up three fallen bottles, putting three objects into a basket, and filling a pen holder with three pens.
Each task has ten layouts generated with a fixed seed and shared by all methods; each method runs once per layout, totaling $120$ trials, with a $900$\,s limit.
A human scores each trial using the task's scoring rubric ($0$--$1$); only a full score counts as a success, and any human intervention counts as a failure.
$\pi_{0.5}$~\citep{black2025pi05} runs closed-loop using public weights trained on RoboTwin~2.0~\citep{chen2025robotwin2}, mapped to the Piper arms and given the same three views; SIMPACT~\citep{liu2026simpact} builds a simulation from one RGB-D image, plans in it with a vision-language model, and executes open-loop.
No method uses demonstrations of these tasks; \model{}'s programs were refined on the robot beforehand (Appendix~\ref{sec:robotics-eval}).

\begin{table}[t]
\centering
\footnotesize
\caption{\textbf{Real-robot comparison.} Successes and mean score (0--1) over ten layouts per task. No method uses demonstrations of these tasks, which are adapted from RoboDojo~\citep{chen2026robodojo}. Scores follow the per-task scoring rules in Table~\ref{tab:rsr-tasks}.}
\label{tab:rsr}
\vspace{-5pt}
\setlength{\tabcolsep}{3pt}
\begin{tabular*}{\linewidth}{@{\extracolsep{\fill}}lcccccc@{}}
\toprule
 & \multicolumn{2}{c}{$\pi_{0.5}$} & \multicolumn{2}{c}{SIMPACT} & \multicolumn{2}{c}{\textbf{\model{}}} \\
\cmidrule(lr){2-3}\cmidrule(lr){4-5}\cmidrule(lr){6-7}
Task & Succ. & Score & Succ. & Score & Succ. & Score \\
\midrule
Stack bowls & 1/10 & 0.19 & 0/10 & 0.09 & \textbf{5/10} & \textbf{0.56} \\
Stand up bottles & 1/10 & 0.12 & 1/10 & 0.14 & \textbf{7/10} & \textbf{0.77} \\
Objects into basket & 3/10 & 0.30 & 2/10 & 0.20 & \textbf{5/10} & \textbf{0.50} \\
Fill pen holder & 0/10 & 0.00 & 0/10 & 0.08 & \textbf{4/10} & \textbf{0.46} \\
\midrule
All & 5/40 & 0.15 & 3/40 & 0.13 & \textbf{21/40} & \textbf{0.57} \\
\bottomrule
\end{tabular*}
\vspace{-5pt}
\end{table}

Table~\ref{tab:rsr} reports successes and mean score.
\model{} succeeds in $21$ of $40$ trials ($52.5\%$), against $5$ for $\pi_{0.5}$ ($12.5\%$) and $3$ for SIMPACT ($7.5\%$), and has the highest mean score on every task.
The largest margin is on standing bottles up (7 of 10, against at most 1 for the baselines), and only \model{} succeeds on filling the pen holder (4 of 10).
In the development demonstration of Figure~\ref{fig:real-sim-robotics} (left), one arm opens a microwave with admittance control that follows the measured force rather than a hinge model; the other places a cube into a bowl and the bowl into the microwave, and the door is then closed, with a minimum sequence duration of $109$\,s.

\section{NeoWorld-10K: An Open-Source Collection of Simulation-Ready Articulated Assets}
\label{sec:dataset}



Building on this pipeline, we leverage \model{} to construct \textbf{NeoWorld-10K}, a large-scale dataset of $10$K articulated 3D assets tailored for simulation-ready scene construction and embodied interaction. Its household-centered taxonomy spans $19$ domains, $219$ families, $536$ subfamilies, and $912$ leaf categories.

Starting from curated object specifications, we synthesize a reference image and three complementary views, then use \model{} to reconstruct part-structured geometry programs and attach explicit joint descriptions. Quality control combines \model{}'s geometry and articulation validation with three-stage human screening of the taxonomy, synthesized images, and reconstructed assets. Final screening inspects each asset from multiple viewpoints, exercises every movable joint over its full available range, and applies drop, inclined-plane, and ground-shaking tests to assess simulated behavior and assembly integrity. The resulting programs retain editable geometry, materials, and kinematics for reuse in interactive environments. We publicly release NeoWorld-10K. Appendix~\ref{app:neoworld} details the taxonomy, construction pipeline, and screening procedures.

\section{Conclusions and Limitations}
\label{sec:conclusion}

We presented \model{}, which reconstructs simulation-ready worlds as grounded program prototypes: agents define the structure, and the images determine the values through measurement and fitting, and \SDK{} and \MCP{} make the result valid for simulation.
\model{} achieves high geometric fidelity on objects and scenes and accurate joint localization on articulated objects. Programs developed in its twins achieve substantially higher success rates than the compared baselines in real-robot manipulation without task-specific demonstrations.
These results link visual reconstruction to interactive simulation and real-robot program development.


Despite these advances, passive visual observations cannot resolve non-visual physical properties---such as internal mass distributions, center-of-mass offsets, contact friction, or occluded joint limits---even under high visual fidelity. Incorporating uncertainty-driven active perception and physical probing (\textit{e.g.}, exploratory robotic pushes or tactile sensing) offers a promising avenue to disambiguate latent dynamics and refine physical parameters via closed-loop feedback.


\iftoggle{cvprfinal}{%
    \section*{Acknowledgments}
We thank Jiajun Ma, Jinghui Zhong, Ning He, Sijia Li, and Haoran Zhao
for their contributions to the construction of the NeoWorld-10K dataset.

}{}




{
    \small
    \bibliographystyle{ieeenat_fullname}
    \bibliography{references}

@misc{blender,
  author       = {{Blender Foundation}},
  title        = {Blender: a Free and Open Source 3D Creation Suite},
  year         = {2025},
  howpublished = {\url{https://docs.blender.org/manual/en/4.5/}},
  note         = {Version 4.5}
}

@misc{blendermcp,
  author       = {Ahuja, Siddharth},
  title        = {{MCP for Blender}: Community Plugin to Control Blender with Language Models},
  year         = {2025},
  howpublished = {\url{https://github.com/ahujasid/mcp-for-blender}},
  note         = {Accessed: 2026-09-26}
}

@misc{manifold,
  author       = {Lalish, Emmett and contributors},
  title        = {Manifold: A Geometry Library for Topological Robustness},
  year         = {2025},
  howpublished = {\url{https://github.com/elalish/manifold}}
}

@misc{mildenhall2020nerf,
      title={NeRF: Representing Scenes as Neural Radiance Fields for View Synthesis}, 
      author={Ben Mildenhall and Pratul P. Srinivasan and Matthew Tancik and Jonathan T. Barron and Ravi Ramamoorthi and Ren Ng},
      year={2020},
      eprint={2003.08934},
      archivePrefix={arXiv},
      primaryClass={cs.CV},
      url={https://arxiv.org/abs/2003.08934}, 
}

@misc{sun2023threedgpt,
  author       = {Sun, Chunyi and Han, Junlin and Deng, Weijian and Wang, Xinlong and Qin, Zishan and Gould, Stephen},
  title        = {{3D-GPT}: Procedural 3D Modeling with Large Language Models},
  year         = {2023},
  howpublished = {arXiv:2310.12945}
}

@inproceedings{hu2024scenecraft,
  author    = {Hu, Ziniu and Iscen, Ahmet and Jain, Aashi and Kipf, Thomas and Yue, Yisong and Ross, David A. and Schmid, Cordelia and Fathi, Alireza},
  title     = {{SceneCraft}: An {LLM} Agent for Synthesizing 3D Scenes as Blender Code},
  booktitle = {International Conference on Machine Learning (ICML)},
  year      = {2024}
}

@inproceedings{huang2024blenderalchemy,
  author    = {Huang, Ian and Yang, Guandao and Guibas, Leonidas},
  title     = {{BlenderAlchemy}: Editing 3D Graphics with Vision-Language Models},
  booktitle = {European Conference on Computer Vision (ECCV)},
  year      = {2024}
}

@misc{kong2026moge3finedetailmonoculargeometry,
      title={MoGe-3: Fine-Detail Monocular Geometry Estimation with Self-Guided Sparse Volumetric Refinement}, 
      author={Lingyu Kong and Ruicheng Li and Ruicheng Wang and Sicheng Xu and Chengtang Yao and Jianfeng Xiang and Jiaolong Yang},
      year={2026},
      eprint={2607.17967},
      archivePrefix={arXiv},
      primaryClass={cs.CV},
      url={https://arxiv.org/abs/2607.17967}, 
}

@inproceedings{poole2023dreamfusion,
  author    = {Poole, Ben and Jain, Ajay and Barron, Jonathan T. and Mildenhall, Ben},
  title     = {{DreamFusion}: Text-to-3D Using 2D Diffusion},
  booktitle = {International Conference on Learning Representations (ICLR)},
  year      = {2023}
}

@misc{xiang2024trellis,
  author       = {Xiang, Jianfeng and Lv, Zelong and Xu, Sicheng and Deng, Yu and Wang, Ruicheng and Zhang, Bowen and Chen, Dong and Tong, Xin and Yang, Jiaolong},
  title        = {Structured 3D Latents for Scalable and Versatile 3D Generation},
  year         = {2024},
  howpublished = {arXiv:2412.01506}
}

@article{laine2020nvdiffrast,
  author  = {Laine, Samuli and Hellsten, Janne and Karras, Tero and Seol, Yeongho and Lehtinen, Jaakko and Aila, Timo},
  title   = {Modular Primitives for High-Performance Differentiable Rendering},
  journal = {ACM Transactions on Graphics},
  volume  = {39},
  number  = {6},
  pages   = {194:1--194:14},
  year    = {2020}
}

@article{kerbl2023gaussians,
  author  = {Kerbl, Bernhard and Kopanas, Georgios and Leimk{\"u}hler, Thomas and Drettakis, George},
  title   = {3D Gaussian Splatting for Real-Time Radiance Field Rendering},
  journal = {ACM Transactions on Graphics},
  volume  = {42},
  number  = {4},
  articleno = {139},
  pages   = {1--14},
  year    = {2023},
  doi     = {10.1145/3592433}
}

@article{articraft,
  title   = {Articraft: An Agentic System for Scalable Articulated 3D Asset Generation},
  author  = {Zhou, Matt and Li, Ruining and Lyu, Xiaoyang and Song, Zhaomou
             and Huang, Zhening and Zheng, Chuanxia and Rupprecht, Christian
             and Vedaldi, Andrea and Wu, Shangzhe},
  journal = {arXiv preprint arXiv:2605.15187},
  year    = {2026}
}

@misc{codebench3d,
  title = {{3DCodeBench}: Benchmarking Agentic Procedural {3D} Modeling Via Code},
  author = {Yipeng Gao and Lei Shu and Genzhi Ye and Xi Xiong and Ameesh Makadia and Meiqi Guo and Laurent Itti and Jindong Chen},
  year = {2026},
  eprint = {2606.01057},
  archivePrefix = {arXiv},
  primaryClass = {cs.CV},
  url = {https://arxiv.org/abs/2606.01057}
}

@misc{sapien,
  title = {{SAPIEN}: A {SimulAted Part-based Interactive ENvironment}},
  author = {Fanbo Xiang and Yuzhe Qin and Kaichun Mo and Yikuan Xia and Hao Zhu and Fangchen Liu and Minghua Liu and Hanxiao Jiang and Yifu Yuan and He Wang and Li Yi and Angel X. Chang and Leonidas J. Guibas and Hao Su},
  year = {2020},
  eprint = {2003.08515},
  archivePrefix = {arXiv},
  primaryClass = {cs.CV},
  url = {https://arxiv.org/abs/2003.08515}
}

@inproceedings{singapo,
  title = {{SINGAPO}: Single Image Controlled Generation of Articulated Parts in Objects},
  author = {Jiayi Liu and Denys Iliash and Angel X. Chang and Manolis Savva and Ali Mahdavi-Amiri},
  booktitle = {International Conference on Learning Representations},
  year = {2025},
  url = {https://arxiv.org/abs/2410.16499}
}

@misc{codexcli,
  title = {{Codex CLI}},
  author = {{OpenAI}},
  year = {2026},
  howpublished = {\url{https://developers.openai.com/codex/cli/}},
  url = {https://developers.openai.com/codex/cli/}
}

@misc{openaigpt,
  title = {Models: {OpenAI API} Documentation},
  author = {{OpenAI}},
  year = {2026},
  howpublished = {OpenAI API documentation. \url{https://developers.openai.com/api/docs/models}},
  url = {https://developers.openai.com/api/docs/models}
}

@article{deboer2005crossentropy,
  title = {A Tutorial on the Cross-Entropy Method},
  author = {de Boer, Pieter-Tjerk and Kroese, Dirk P. and Mannor, Shie and Rubinstein, Reuven Y.},
  journal = {Annals of Operations Research},
  volume = {134},
  number = {1},
  pages = {19--67},
  year = {2005},
  doi = {10.1007/s10479-005-5724-z}
}

@inproceedings{todorov2012mujoco,
  title = {{MuJoCo}: A Physics Engine for Model-Based Control},
  author = {Todorov, Emanuel and Erez, Tom and Tassa, Yuval},
  booktitle = {IEEE/RSJ International Conference on Intelligent Robots and Systems (IROS)},
  pages = {5026--5033},
  year = {2012},
  doi = {10.1109/IROS.2012.6386109}
}

@misc{viga,
  title = {Vision-as-Inverse-Graphics Agent via Interleaved Multimodal Reasoning},
  author = {Shaofeng Yin and Jiaxin Ge and Zora Zhiruo Wang and Chenyang Wang and Xiuyu Li and Michael J. Black and Trevor Darrell and Angjoo Kanazawa and Haiwen Feng},
  year = {2026},
  eprint = {2601.11109},
  archivePrefix = {arXiv},
  primaryClass = {cs.CV},
  url = {https://arxiv.org/abs/2601.11109}
}

@misc{siglip2,
  title = {{SigLIP 2}: Multilingual Vision-Language Encoders with Improved Semantic Understanding, Localization, and Dense Features},
  author = {Michael Tschannen and Alexey Gritsenko and Xiao Wang and Muhammad Ferjad Naeem and Ibrahim Alabdulmohsin and Nikhil Parthasarathy and Talfan Evans and Lucas Beyer and Ye Xia and Basil Mustafa and Olivier H{\'e}naff and Jeremiah Harmsen and Andreas Steiner and Xiaohua Zhai},
  year = {2025},
  eprint = {2502.14786},
  archivePrefix = {arXiv},
  primaryClass = {cs.CV},
  url = {https://arxiv.org/abs/2502.14786}
}

@misc{dinov3,
  title = {{DINOv3}},
  author = {Oriane Sim{\'e}oni and Huy V. Vo and Maximilian Seitzer and Federico Baldassarre and Maxime Oquab and Cijo Jose and Vasil Khalidov and Marc Szafraniec and Seungeun Yi and Micha{\"e}l Ramamonjisoa and Francisco Massa and Daniel Haziza and Luca Wehrstedt and Jianyuan Wang and Timoth{\'e}e Darcet and Th{\'e}o Moutakanni and Leonel Sentana and Claire Roberts and Andrea Vedaldi and Jamie Tolan and John Brandt and Camille Couprie and Julien Mairal and Herv{\'e} J{\'e}gou and Patrick Labatut and Piotr Bojanowski},
  year = {2025},
  eprint = {2508.10104},
  archivePrefix = {arXiv},
  primaryClass = {cs.CV},
  url = {https://arxiv.org/abs/2508.10104}
}

@misc{uni3d,
  title = {{Uni3D}: Exploring Unified {3D} Representation at Scale},
  author = {Junsheng Zhou and Jinsheng Wang and Baorui Ma and Yu-Shen Liu and Tiejun Huang and Xinlong Wang},
  year = {2023},
  eprint = {2310.06773},
  archivePrefix = {arXiv},
  primaryClass = {cs.CV},
  url = {https://arxiv.org/abs/2310.06773}
}

@misc{pointset,
  title = {A Point Set Generation Network for {3D} Object Reconstruction from a Single Image},
  author = {Haoqiang Fan and Hao Su and Leonidas Guibas},
  year = {2016},
  eprint = {1612.00603},
  archivePrefix = {arXiv},
  primaryClass = {cs.CV},
  url = {https://arxiv.org/abs/1612.00603}
}

@misc{giou,
  title = {Generalized Intersection over Union: A Metric and A Loss for Bounding Box Regression},
  author = {Hamid Rezatofighi and Nathan Tsoi and JunYoung Gwak and Amir Sadeghian and Ian Reid and Silvio Savarese},
  year = {2019},
  eprint = {1902.09630},
  archivePrefix = {arXiv},
  primaryClass = {cs.CV},
  url = {https://arxiv.org/abs/1902.09630}
}

@inproceedings{cage,
  title = {{CAGE}: Controllable Articulation {GEneration}},
  author = {Jiayi Liu and Hou In Ivan Tam and Ali Mahdavi-Amiri and Manolis Savva},
  booktitle = {Proceedings of the IEEE/CVF Conference on Computer Vision and Pattern Recognition},
  year = {2024},
  pages = {17880--17889},
  url = {https://arxiv.org/abs/2312.09570}
}

@inproceedings{robocasa365,
  title={RoboCasa365: A Large-Scale Simulation Framework for Training and Benchmarking Generalist Robots},
  author={Soroush Nasiriany and Sepehr Nasiriany and Abhiram Maddukuri and Yuke Zhu},
  booktitle={International Conference on Learning Representations (ICLR)},
  year={2026}
}

@inproceedings{articulateanything,
  title     = {Articulate-Anything: Automatic Modeling of Articulated Objects via a Vision-Language Foundation Model},
  author    = {Le, Long and Xie, Jason and Liang, William and Wang, Hung-Ju and Yang, Yue and Ma, Yecheng Jason and Vedder, Kyle and Krishna, Arjun and Jayaraman, Dinesh and Eaton, Eric},
  booktitle = {International Conference on Learning Representations (ICLR)},
  year      = {2025}
}

@article{sam3d,
  title        = {SAM 3D: 3Dfy Anything in Images},
  author       = {SAM 3D Team and Chen, Xingyu and Chu, Fu-Jen and Gleize, Pierre
                  and Liang, Kevin J. and Sax, Alexander and Tang, Hao
                  and Wang, Weiyao and Guo, Michelle and Hardin, Thibaut
                  and Li, Xiang and Lin, Aohan and Liu, Jiawei and Ma, Ziqi
                  and Sagar, Anushka and Song, Bowen and Wang, Xiaodong
                  and Yang, Jianing and Zhang, Bowen and Doll{\'a}r, Piotr
                  and Gkioxari, Georgia and Feiszli, Matt and Malik, Jitendra},
  journal      = {arXiv preprint arXiv:2511.16624},
  year         = {2025}
}

@article{pixal3d,
  title   = {Pixal3D: Pixel-Aligned 3D Generation from Images},
  author  = {Li, Dong-Yang and Zhao, Wang and Chen, Yuxin and Hu, Wenbo
             and Guo, Meng-Hao and Zhang, Fang-Lue and Shan, Ying
             and Hu, Shi-Min},
  journal = {arXiv preprint arXiv:2605.10922},
  year    = {2026}
}

@inproceedings{spark,
  title     = {SPARK: Sim-ready Part-level Articulated Reconstruction
               with VLM Knowledge},
  author    = {He, Yumeng and Jiang, Ying and Lu, Jiayin and Yang, Yin
               and Jiang, Chenfanfu},
  booktitle = {Proceedings of the IEEE/CVF Conference on Computer Vision
               and Pattern Recognition},
  year      = {2026}
}

@inproceedings{physxanything,
  title     = {PhysX-Anything: Simulation-Ready Physical 3D Assets
               from Single Image},
  author    = {Cao, Ziang and Hong, Fangzhou and Chen, Zhaoxi
               and Pan, Liang and Liu, Ziwei},
  booktitle = {Proceedings of the IEEE/CVF Conference on Computer Vision
               and Pattern Recognition},
  year      = {2026}
}

@inproceedings{urdfanything,
  title     = {URDF-Anything: Constructing Articulated Objects with
               3D Multimodal Language Model},
  author    = {Li, Zhe and Bai, Xiang and Zhang, Jieyu and Wu, Zhuangzhe
               and Xu, Che and Li, Ying and Hou, Chengkai and Zhang, Shanghang},
  booktitle = {Advances in Neural Information Processing Systems},
  volume    = {38},
  year      = {2025}
}

@inproceedings{real2code,
  title     = {Real2Code: Reconstruct Articulated Objects via Code Generation},
  author    = {Mandi, Zhao and Weng, Yijia and Bauer, Dominik and Song, Shuran},
  booktitle = {International Conference on Learning Representations},
  year      = {2025}
}

@article{3dcos,
  title   = {3D-CoS: A New 3D Reconstruction Paradigm Based on VLM Code Synthesis},
  author  = {Wang, Yuhao and Wang, Puyi and Li, Linjie and Yang, Zhengyuan
             and Lin, Kevin Qinghong and Cheng, Yu},
  journal = {arXiv preprint arXiv:2606.10478},
  year    = {2026}
}

@misc{carion2025sam3,
  author       = {Carion, Nicolas and Gustafson, Laura and others},
  title        = {{SAM 3}: Segment Anything with Concepts},
  year         = {2025},
  howpublished = {arXiv:2511.16719}
}

@misc{wang2026vggtomega,
  author       = {Wang, Jianyuan and Chen, Minghao and Zhang, Shangzhan and Karaev, Nikita and Sch{\"o}nberger, Johannes and Labatut, Patrick and Bojanowski, Piotr and Novotny, David and Vedaldi, Andrea and Rupprecht, Christian},
  title        = {{VGGT-$\Omega$}},
  year         = {2026},
  howpublished = {arXiv:2605.15195}
}

@misc{black2025pi05,
  author       = {{Physical Intelligence} and Black, Kevin and Brown, Noah and others},
  title        = {{$\pi_{0.5}$}: a Vision-Language-Action Model with Open-World Generalization},
  year         = {2025},
  howpublished = {arXiv:2504.16054}
}

@inproceedings{liu2026simpact,
  author    = {Liu, Haowen and Yao, Shaoxiong and Chen, Haonan and Gao, Jiawei and Mao, Jiayuan and Huang, Jia-Bin and Du, Yilun},
  title     = {{SIMPACT}: Simulation-Enabled Action Planning using Vision-Language Models},
  booktitle = {IEEE/CVF Conference on Computer Vision and Pattern Recognition (CVPR)},
  year      = {2026}
}

@misc{chen2025robotwin2,
  author       = {Chen, Tianxing and Chen, Zanxin and Chen, Baijun and Cai, Zijian and Liu, Yibin and others},
  title        = {{RoboTwin 2.0}: A Scalable Data Generator and Benchmark with Strong Domain Randomization for Robust Bimanual Robotic Manipulation},
  year         = {2025},
  howpublished = {arXiv:2506.18088}
}

@misc{chen2026robodojo,
  author       = {Chen, Tianxing and Chen, Yue and Li, Zixuan and Tang, Junyuan and Su, Kailun and others},
  title        = {{RoboDojo}: A Unified Sim-and-Real Benchmark for Comprehensive Evaluation of Generalist Robot Manipulation Policies},
  year         = {2026},
  howpublished = {arXiv:2607.04434}
}

@inproceedings{wen2024foundationpose,
  author    = {Wen, Bowen and Yang, Wei and Kautz, Jan and Birchfield, Stan},
  title     = {{FoundationPose}: Unified {6D} Pose Estimation and Tracking of Novel Objects},
  booktitle = {IEEE/CVF Conference on Computer Vision and Pattern Recognition (CVPR)},
  year      = {2024}
}
}


\clearpage
\raggedbottom
\onecolumn
\appendix
\begingroup
\newcolumntype{N}{c}
\newcommand{\appbold}[1]{{\bfseries\boldmath #1}}
\makeatletter
\renewcommand{\cvprsection}{\@startsection{section}{1}{\z@}
  {-14pt plus -3pt minus -2pt}{8pt}{\normalfont\large\bfseries}}
\renewcommand{\cvprsubsection}{\@startsection{subsection}{2}{\z@}
  {-11pt plus -2pt minus -2pt}{5pt}{\normalfont\fontsize{10.95}{13}\selectfont\bfseries}}
\renewcommand{\cvprsubsubsection}{\@startsection{subsubsection}{3}{\z@}
  {-8pt plus -2pt minus -1pt}{4pt}{\normalfont\normalsize\bfseries}}
\renewcommand{\paragraph}{\@startsection{paragraph}{4}{\z@}
  {6pt plus 2pt minus 1pt}{-0.8em}{\normalfont\normalsize\bfseries}}
\newcommand{\appneedspace}[1]{%
  \par\begingroup\@tempdima=#1\relax
  \ifdim\dimexpr\pagegoal-\pagetotal\relax<\@tempdima\newpage\fi
  \endgroup}
\@addtoreset{table}{section}
\@addtoreset{figure}{section}
\@addtoreset{equation}{section}
\renewcommand{\thetable}{\Alph{section}.\arabic{table}}
\renewcommand{\thefigure}{\Alph{section}.\arabic{figure}}
\renewcommand{\theequation}{\Alph{section}.\arabic{equation}}
\renewcommand{\theHtable}{appendix.\Alph{section}.\arabic{table}}
\renewcommand{\theHfigure}{appendix.\Alph{section}.\arabic{figure}}
\renewcommand{\theHequation}{appendix.\Alph{section}.\arabic{equation}}
\makeatother

\setlength{\parindent}{1em}
\setlength{\parskip}{0pt}
\setlength{\textfloatsep}{12pt plus 3pt minus 2pt}
\setlength{\floatsep}{10pt plus 2pt minus 2pt}
\setlength{\intextsep}{10pt plus 2pt minus 2pt}
\renewcommand{\topfraction}{0.9}
\renewcommand{\bottomfraction}{0.8}
\renewcommand{\textfraction}{0.08}
\renewcommand{\floatpagefraction}{0.75}
\setcounter{topnumber}{3}
\setcounter{bottomnumber}{2}
\setcounter{totalnumber}{5}
\captionsetup{font=small,labelfont=bf,labelsep=period,
  justification=raggedright,singlelinecheck=false,skip=5pt}
\setlist[itemize]{leftmargin=1.4em,itemsep=2pt,topsep=4pt,parsep=0pt,partopsep=0pt}
\setlist[enumerate]{leftmargin=1.7em,itemsep=3pt,topsep=4pt,parsep=0pt,partopsep=0pt}
\setlist[description]{leftmargin=1.2em,labelindent=0pt,itemsep=3pt,topsep=4pt,parsep=0pt}
\AtBeginEnvironment{table}{\small\setlength{\tabcolsep}{6pt}\renewcommand{\arraystretch}{1.16}}
\AtBeginEnvironment{table*}{\small\setlength{\tabcolsep}{6pt}\renewcommand{\arraystretch}{1.16}}
\AtBeginEnvironment{tabular}{\small}
\AtBeginEnvironment{tabular*}{\small}
\AtBeginEnvironment{tabularx}{\small}

\section*{Appendix contents}
\label{sec:appendix-contents}
\noindent Implementation details, dataset construction, and evaluation protocols.
\medskip

\newcommand{\appentry}[3]{%
  \noindent\hspace*{#1}\hyperref[#2]{\ref*{#2}\quad #3}%
  \nobreak\leaders\hbox{.}\hfill\nobreak\hyperref[#2]{\pageref*{#2}}\par
}

\newcommand{\appsectionentry}[2]{\appentry{0pt}{#1}{\textbf{#2}}}

\begingroup
\small
\setlength{\parskip}{2.5pt}
\medskip

\appsectionentry{sec:method-details}{Framework design and reconstruction}
\appentry{1em}{sec:design-principles}{Design principles}
\appentry{1em}{sec:scene}{Scene reconstruction along the program tree}
\appentry{1em}{sec:object}{Object modeling and numerical refinement}
\appentry{1em}{sec:physics}{Physics, articulation, and geometry validity}
\appentry{1em}{sec:predicted-depth}{Predicted depth and scale alignment}
\medskip

\appsectionentry{sec:robotic-deployment}{Robotic manipulation and deployment}
\appentry{1em}{sec:robotics}{Connecting reconstructed worlds to Real-Sim-Real robotics}
\appentry{1em}{sec:robotics-real}{Physical registration and re-posing from wrist scans}
\appentry{1em}{sec:robotics-compliant}{Compliant execution on the physical arms}
\medskip

\appsectionentry{sec:sdk-mcp-implementation}{NeoSDK and NeoMCP implementation}
\appentry{1em}{sec:sdk}{NeoSDK: source programs and numerical execution}
\appentry{1em}{sec:mcp}{NeoMCP: agent-facing operations}
\medskip

\appsectionentry{app:neoworld}{NeoWorld-10K: taxonomy and construction details}
\appentry{1em}{app:neoworld-scope}{Dataset scope}
\appentry{1em}{app:neoworld-taxonomy}{Hierarchical semantic organization}
\appentry{1em}{app:neoworld-reference-construction}{Dataset construction}
\appentry{1em}{app:neoworld-evaluation}{Three-stage human screening}
\appentry{1em}{app:neoworld-organization}{Dataset organization and annotations}
\medskip

\appsectionentry{sec:experiment-appendix}{Supplementary experiments and evaluation protocols}
\appentry{1em}{sec:appendix-evaluation-datasets}{Datasets}
\appentry{1em}{sec:appendix-evaluation-baselines}{Baselines and comparison settings}
\appentry{1em}{sec:appendix-evaluation-metrics}{Evaluation metrics}
\appentry{1em}{sec:appendix-twin-evidence}{Program-twin coverage, fidelity, and efficiency}
\appentry{1em}{sec:scene-compute}{Scene compute and controlled ablations}
\appentry{1em}{sec:articulation-penetration}{Supplementary articulation evaluation}
\appentry{1em}{sec:watertight-evaluation}{Watertightness evaluation and ablations}
\appentry{1em}{sec:robotics-eval}{Real-robot comparison: protocol and additional results}
\medskip


\endgroup
\clearpage

\section{Framework design and reconstruction}
\label{sec:method-details}
\label{sec:executable-reconstruction}

\subsection{Design principles}
\label{sec:design-principles}

This appendix follows the system from the top down. It starts with the scene
route (Section~\ref{sec:scene}) and the object route
(Section~\ref{sec:object}). It then covers the physical and articulation checks
(Section~\ref{sec:physics}), and the robotics handoff
(Section~\ref{sec:robotics}), and ends with the two layers they all share,
\SDK{} (Section~\ref{sec:sdk}) and \MCP{} (Section~\ref{sec:mcp}).
Figure~\ref{fig:architecture} gives the overall loop. Seven principles recur
throughout, and most mechanisms below exist to uphold one of them.

\begin{description}
\setlength{\itemsep}{2pt}
\item[The source is the state.] A reconstruction is an executable program
$P=(S,\theta,T)$, not a mesh detached from its construction. Structural changes
edit the source, numerical fitting moves named constants and poses, and every
retained result can be re-executed and continued (Sections~\ref{sec:sdk}
and~\ref{sec:object-transactions}).
\item[Values are grounded.] Every value in a program is measured from the
observations, fitted where some view can decide it, or assumed. Programs read
measured values by name and can ask where each one comes from. A fitted
measurement lives outside the program text and overrides the measured value in
the next build; a number the program writes itself, as a named constant or an
inline literal, is fitted in place in the source, and every such write is
receipted (Sections~\ref{sec:scene-measurement}, \ref{sec:scene-tests}
and~\ref{sec:object-proposals}).
\item[Agents argue with evidence.] Agents write source, request fits and ask to
finish; every edit, objection and rebuttal must cite the views. A deterministic
harness owns schemas, budgets and the admission of every candidate, rolls back
changes that the views contradict, and never edits a semantic value to make a
program pass. Critics' objections and the system's own fits answer to the same
views (Sections~\ref{sec:scene-tests} and~\ref{sec:mcp-harness}).
\item[Divide and conquer.] The planner divides a scene into regions along its
program tree. Each object has one owner, who reads the whole scene but writes
only what it owns; anything else it notices reaches its owner as a request.
Owners work in parallel, and a regression in the composed scene is traced to the
object that caused it (Section~\ref{sec:scene-ownership}).
\item[Search in the twin, admit in Blender.] Fitting is batched GPU search on
twins: population search on a tensor twin of the modeling
vocabulary for program values, and pose search on a twin of the scene. Blender
builds the baseline and
materializes only the chosen candidates for admission. A construction the twin
cannot replay is a typed failure, never a per-candidate Blender fallback
(Sections~\ref{sec:sdk} and~\ref{sec:object-proposals}).
\item[Checks report only what they measured.] Unobserved joint fields,
unmeasurable parts and unavailable measurements stay typed as unknown, and a
check that finds no refutation reports \emph{unrefuted}, not verified
(Sections~\ref{sec:scene-articulation} and~\ref{sec:object-articulation}).
\item[Everything is bounded.] The host counts model calls, wall time, images
sent to the model and child-call allowances. Refusals, failures and budget
exhaustion are recorded as distinct outcomes (Sections~\ref{sec:mcp-registry},
\ref{sec:scene-ownership} and~\ref{sec:scene-object-workers}).
\end{description}

\subsection{Scene reconstruction along the program tree}
\label{sec:scene}
\label{sec:overview}

A scene such as a kitchen holds dozens of objects that must stay one coherent
world: they meet, rest on each other, and are seen together by every camera.
\model{} reconstructs a scene as a tree of grounded prototypes
(Section~\ref{sec:method-editable-main}) and models it in rounds. A planning call
lists the objects, and the harness measures them without any model call.
Modeling sub-agents build and repair the regions they own. After every round,
the harness tests the integrated scene against the input views, reverts what
made it worse, and turns what it finds into issues for the owners
(Section~\ref{sec:method-reconstruction}). This section follows that route in
order. Section~\ref{sec:scene-live} describes the single-agent live-scene
route, whose agent calls the \MCP{} verbs of Table~\ref{tab:verbs} directly;
the sub-agents of this route see a narrower surface built on those verbs
(\emph{Tools} in Section~\ref{sec:scene-ownership}).

\subsubsection{Inputs, plan, and measurement}
\label{sec:scene-measurement}

\paragraph{Inputs.}
In the image-only RoboCasa setting, a scene is given as $16$ RGB views. Cameras
are estimated with VGGT-$\Omega$~\citep{wang2026vggtomega}, and depth is
predicted with MoGe-3~\citep{kong2026moge3finedetailmonoculargeometry}. No
ground-truth camera, depth, or layout enters a run.

\paragraph{Plan.}
One planning call (\code{gpt-6-sol}) sees all views at $768$\,px and returns the
room and its objects, grouped into regions (two in the reported runs). Each
object carries:
\begin{itemize}
\setlength{\itemsep}{1pt}\setlength{\parskip}{0pt}
\item a name and a one-sentence description;
\item a segmentation prompt that tells it apart from its neighbors;
\item a kind, a common noun that the export uses as its category;
\item the objects it depends on, and its region;
\item how it is measured: as one object, as a horizontal top, or as a repeated kind.
\end{itemize}

\paragraph{Measurement.}
The harness then measures every planned object without a model call.
SAM~$3$~\citep{carion2025sam3} segments the object's prompt in every view, and the
mask is lifted to 3D with the predicted depth through the estimated cameras. A
box lifted this way is inflated by whatever the mask spills onto, so it is
carved by what the views see past it:
\begin{itemize}
\setlength{\itemsep}{1pt}\setlength{\parskip}{0pt}
\item Points are sampled on a $2$\,cm grid inside the box.
\item A view votes for a point when the point projects into the object's mask.
It votes against the point when the point projects outside the mask and the
observed surface lies at or beyond it. It abstains when the observed surface is
nearer, since occlusion is not absence.
\item Points with at least two votes against and a majority are removed. The box
keeps the slices where at least $35\%$ of the best slice survives.
\item A measurement that loses more than $40\%$ of its box is marked untrusted.
The build card then tells the agent to judge the size from the images.
\end{itemize}
A repeated kind is inventoried over all views and grouped into rows with a
count, a spacing, and a facing. Each copy gets a position and a yaw. The yaw is
measured where the row's front is seen; otherwise it is assumed toward the
middle of the rows and marked as inferred. A horizontal top gets its footprint
from a sweep of masks, and a single object also gets a color read inside its
masks. Planes measured from the depth give the floor, the walls, and the ceiling
that every object shares.

\paragraph{Facts.}
Measurements are stored per object, both as data and as a readable evidence
file. A program reads them through one entry point, \code{facts()}, which also
answers where a value comes from: how it was measured, and whether it was seen
or inferred. Every top-level fact a program reads is recorded, and these are the
values a later fit may adjust. Fitted facts go to a file that only the harness
writes and that overrides the facts in the next build, so fitting a fact never
edits the program text; the numbers a program writes itself are fitted in place
(Section~\ref{sec:scene-tests}).

\subsubsection{Modeling sub-agents and ownership}
\label{sec:scene-ownership}

\paragraph{One folder per object.}
The tree keeps one folder per object, holding its program, its evidence and
facts, and its fitted values. Shared programs for the room frame and the
materials run before every object. Repeated objects of one kind share one
program, and each copy takes its pose from the measured rows.

\paragraph{Owners and budgets.}
Each round runs one modeling sub-agent per region and a structure specialist for
the room shell, all of them Codex CLI~\citep{codexcli} sessions with \code{gpt-6-sol}. A
sub-agent owns its region's folders. It reads the whole tree and works in a
snapshot of the scene in which the objects its region depends on already stand,
but it writes only its own folders. What it notices about another object goes
to that object's inbox as a request. A region's sub-agent has at most $80$ calls
and $2$\,$400$\,s, and a single-object owner such as the room has $40$ calls and
$1$\,$500$\,s; $300$\,s of each budget are reserved for finishing.

\paragraph{Tools.}
A sub-agent does not see the verbs of Table~\ref{tab:verbs}. The tree registers
its own operations on the \MCP{} registry and implements them with those verbs:
\code{build} runs the programs through \code{execute\_blender\_code} and
\code{edit\_code} in the live Blender, \code{measure} calls \code{measure} and
\code{read\_observation}, and \code{look} renders through
\code{render\_calibrated}; the other verbs stay registered for the harness.
Besides reading, writing, and editing files in the tree, a sub-agent has these
operations:
\begin{itemize}
\setlength{\itemsep}{1pt}\setlength{\parskip}{0pt}
\item \code{measure(object, prompt, how)} measures an object, including one the
agent adds.
\item \code{build(object)} runs the shared programs and the object's program and
returns a status card. The card compares the built count with the measured count
first, then the built box with the measured box. It also lists coincident faces,
which render black: faces of two meshes of the object that are parallel, face
the same way, lie within $1$\,mm, and overlap by at least $20$\,cm$^2$. Last come the
program's own messages.
\item \code{look(object, views)} shows each image beside the render through the
same camera.
\item \code{fit\_pose(object, parts, views)} searches a planar move of up to
$0.3$\,m and a turn of up to $15$$^\circ$ of the object's built parts against up to
six views, by a local coordinate descent on depth error whose candidates are
rendered on the scene twin (see \emph{Copies} below). It reports the move with
its per-view error, and the agent decides whether to apply it.
\item \code{plan(band)} shows the room from above in a horizontal band, with the
surfaces the photographs' depth saw and the model's objects.
\item \code{scene\_planes()} returns the planes fitted to the depth: floor,
walls, and counters.
\item \code{finish} ends the session and states what remains uncertain.
\end{itemize}
None of these operations accepts, refuses, or blocks a change: the card reports,
and the agent decides.

\paragraph{In-build review.}
While a sub-agent builds, every object it builds or edits is reviewed in the
background by the checks the round runs afterwards, restricted to that object.
\begin{itemize}
\setlength{\itemsep}{1pt}\setlength{\parskip}{0pt}
\item The live scene is saved between two tool calls and rendered through the
review cameras in a separate Blender.
\item Zero-model checks look for collisions, geometry outside the room, and
attached parts that stand off their hosts.
\item A vision-language critic reviews the object's
structure.
\end{itemize}
The notes, most severe first and with up to two of the critic's crops, are
attached to the agent's next tool result. A review that cannot run says so.
Like the card, a review only reports.

\paragraph{Copies.}
After a build, each photographed copy of a repeated kind is posed from its own
views, with one built copy as the template. The search runs on the scene twin, a
resident GPU raster of the integrated scene that renders any pose of any object
through the estimated cameras without Blender.
\begin{itemize}
\setlength{\itemsep}{1pt}\setlength{\parskip}{0pt}
\item \emph{Points.} The copy's masks, lifted with depth (nearest layer only),
give its points.
\item \emph{Search.} Every $5$$^\circ$ of yaw, with offsets on a $4$\,cm grid within
$0.2$\,m of the copy's measured position.
\item \emph{Score.} The depth gap plus one minus the silhouette agreement. The
depth gap is the mean distance of the points to the template surface, capped at
$0.15$\,m and divided by that cap. The silhouette agreement is the IoU of the
projected template against the copy's masks, outside occluded pixels.
\end{itemize}
The copy is snapped to the best pose, which is kept with its margin over the
best pose at least $45$$^\circ$ away; a margin of $0.1$ or more counts as a decided
turn.

\subsubsection{Rounds, tests, and issues}
\label{sec:scene-tests}

\paragraph{Rounds.}
Round~$1$ builds every object from the measurements, with nothing standing yet.
Rounds $2$--$6$ repair. Before a repair round's sub-agents start, the harness:
\begin{itemize}
\setlength{\itemsep}{1pt}\setlength{\parskip}{0pt}
\item reverts the regressions of the previous round;
\item checks the cameras;
\item writes new issues into the owners' inboxes;
\item fits each object's measured facts on the program twin (below);
\item measures newly found objects.
\end{itemize}
After the sub-agents finish, their folders are collected and the scene is
integrated without a model call. Objects build in dependency order with hosts
first, and a program that calls another program's function runs after it.

\paragraph{Fitting measured facts.}
Before a repair round's sub-agents start, each object's measured facts and the
numbers written in its file are fitted on the program twin
(Section~\ref{sec:object-proposals}) to the object's SAM~$3$ masks, bound to it
view by view, with the rest of the scene held fixed. A direction is searched
only when a $2$\,cm step changes the object's silhouette IoU in some view by at
least $0.002$. Fitted facts are written to the fit file by their paths and fitted
numbers at their exact positions in the source; rebuilding from both reproduces
the searched geometry within $0.5$\,mm. One inbox line per object reports the
change, and the next round's frame check admits or reverts it. No model judges
a fit: in a ten-object test, gating each fit by a vision-language model's
approval, or by the agreement of two half-view searches, kept less of the
improvement than fitting without a gate.

\paragraph{Frame check.}
Every view renders the integrated scene through its camera. The view's residual
is the median absolute difference between the observed and the rendered depth,
over the pixels where both see a surface within $4$\,m. It is computed overall and per object, the
object being the one that owns the pixel. A view is worse when its residual
exceeds the last accepted scene's by at least $2$\,cm and by at least $8\%$.

Worse views are grouped. In each group, every object that changed in the round
and appears there is swapped back to its accepted version, one at a time, and
the share of the worsening the swap undoes is that object's responsibility.
\begin{itemize}
\setlength{\itemsep}{1pt}\setlength{\parskip}{0pt}
\item An object responsible for at least half of it, whose revert makes no other
view worse by $2$\,cm, is reverted to its accepted version. Its inbox receives the
image, the accepted render, and the regressed render.
\item Otherwise the group becomes an issue for the objects that undo part of it.
\end{itemize}
Views whose residual stands at three times the scene's median or more in two
consecutive rounds, whatever changed, point to a camera problem rather than a
shape problem. A separate camera check compares each view's own floor and walls
with the scene's; views it marks untrusted are left out of the copy fit and the
frame check.

\paragraph{Issues.}
Four checks raise issues against the integrated scene:
\begin{itemize}
\setlength{\itemsep}{1pt}\setlength{\parskip}{0pt}
\item \emph{Shape discovery.} Regions where the scene and the views' depth
disagree go to a critic agent with three tools: the views
that see a region or a point, a side-by-side look at one view's image and model
(optionally cropped), and a report. It may browse beyond the regions. It reports
named issues, each with an owner (an existing object, or a new object with a
segmentation prompt) and the two to four views that show it best. A new object
is measured with those views and admitted to the plan when they hold it.
\item \emph{Silhouette audit.} SAM~$3$ masks of each object's prompt are bound to
the object by identity, and a box fit on the scene's twin reads its width along
its nearest wall on held-out views. A width that differs from the model's by
more than $12\%$ becomes an issue. Height and depth are not reported, because the
model's own walls and neighboring masks confound them.
\item \emph{Edge readings.} Each face of a singular object is triangulated from
the object's mask edges:
\begin{itemize}
\setlength{\itemsep}{0pt}
\item A view contributes one ray when half of its image rows agree on the edge
within $2$\,cm. It then sees a straight edge, not a handle.
\item A row counts only where the $2$\,cm just outside the edge lie in front of
every other model surface.
\item Rays from directions at least $12$$^\circ$ apart meet by least squares. Three
or more rays must meet within $3$\,cm; two-view readings are kept but flagged as
unchecked.
\end{itemize}
A face more than $3$\,cm from the model's face becomes one issue for that face.
\item \emph{Escapes.} A view that looks out of the model where the images see a
surface raises an issue for the room. The issue states where the views' depth
puts that surface. An escape counts as fixed only when the model there agrees
with the views' depth within $0.1$\,m, not when the view is merely blocked.
\end{itemize}

\paragraph{Issue ledger.}
Every issue gets an identifier and a state, and it is matched across rounds: by
object and axis for measured issues, and by object and wording for the critic's.
\begin{itemize}
\setlength{\itemsep}{1pt}\setlength{\parskip}{0pt}
\item A measured issue closes only when the next round's check of the rebuilt
scene no longer finds it. A critic issue closes when the critic no longer raises
it.
\item An answer of \emph{disputed} is kept, with the views it cites.
\item An object that a check finds off, but that was not off one round earlier,
is opened as a regression of that round's changes.
\item Inboxes list the open entries, longest open first. An entry open for two
rounds is marked unresolved.
\end{itemize}
Agents answer every identifier. Cards list semantic issues (missing, extra,
kind, count, facing) before metric ones.

\subsubsection{Finishing the scene}
\label{sec:scene-finish}
After the last round, the integrated scene is saved and exported, with each
object's kind as its category. A final appearance pass fits lights and colors to
the images through the method's own cameras:
\begin{itemize}
\setlength{\itemsep}{1pt}\setlength{\parskip}{0pt}
\item \emph{Lights.} Area lights hang just under the ceiling over the observed
floor. A single power per square metre of floor is fitted so that the scene,
rendered with Cycles, is as bright as the images: the median over views of the
mean luminance.
\item \emph{Colors.} Each material's base color is scaled by the ratio of the
images' median color to the render's, inside the material's eroded pixels. The
scaling applies only when the views agree, with a log-brightness spread of at
most $0.35$ over at least three views, and when the total change stays within a
factor of $0.2$--$5$. The lights are then refitted.
\end{itemize}
The agreement rule refuses objects built too large, whose pixels cover the
surfaces behind them.

\subsubsection{Single-agent live-scene route}
\label{sec:scene-live}
\model{} can also model a scene as one persistent episode of a single scene
agent in a live Blender, through the \MCP{} surface of Table~\ref{tab:verbs}. The
agent can hand individual bodies to the object route
(Section~\ref{sec:object}). A returned body is integrated only after the harness
has checked that it is no worse than the part it replaces and that nothing else
moved.


\begin{table*}[t]
\centering
\footnotesize
\setlength{\tabcolsep}{6pt}
\caption{\textbf{The scene agent's tools: the \MCP{} live-scene surface.} Each
verb carries a strict schema and returns a typed record
(Section~\ref{sec:mcp}). Verbs marked $\dagger$ are mounted only when their
option is enabled; each run records the tools it actually mounted. Diagnostic
labels and admission reports are not independent reconstruction-quality
verdicts.}
\label{tab:verbs}
\begin{tabularx}{\textwidth}{@{}lX@{}}
\toprule
\textbf{Verb} & \textbf{What the call returns} \\
\midrule
\multicolumn{2}{@{}l}{\textit{Observation and measurement}}\\
\code{read\_observation} & one photograph at full upright resolution with its raw camera and depth \\
\code{measure}           & text-prompted masks with their metric lift: bounds, centroid, principal normal, floor footprint \\
\code{sample\_depth}     & world points under named pixels, with per-pixel validity \\
\code{scene\_planes}     & measured planes of the capture: normal, offset, extent, support \\
\code{triangulate}$^\dagger$ & a world point from $2$--$64$ calibrated views of one agent-supplied landmark, with residuals \\
\addlinespace
\multicolumn{2}{@{}l}{\textit{Live scene}}\\
\code{inspect}           & scene summary, or per object: transform, bounds, materials, mesh and manifold counts \\
\code{execute\_blender\_code} & result of running a labeled program in the live scene; names persist across calls \\
\code{edit\_code}        & result of an exact-anchor edit, re-run after deleting exactly what the previous version created \\
\code{read\_source}      & a kept label's source with line numbers, the text \code{edit\_code} anchors against \\
\code{view\_image}       & an earlier image again, by evidence identifier \\
\addlinespace
\multicolumn{2}{@{}l}{\textit{Calibrated comparison and review}}\\
\code{render\_view}       & a shaded view through a recorded or free camera, for inspection \\
\code{render\_calibrated} & photograph / render / difference panels through the real cameras, with agreement per view \\
\code{review\_views}$^\dagger$ & ranked suggestions from independent critics over those panels; no images returned \\
\code{scan\_discrepancies}$^\dagger$ & missing, extra, seen-through, hidden, and never-visible geometry against every frame's depth \\
\addlinespace
\multicolumn{2}{@{}l}{\textit{Numerical refinement, placement, and delegation}}\\
\code{observe\_parameters} & named quantities, the views that can see them, and proposed fitting groups for supplied targets \\
\code{probe\_refinement} & bounded read-only twin perturbations, changed members, and per-view measurement deltas \\
\code{refine}            & the fitted result, applied only if the harness admits it, with before/after per view \\
\code{assemble}          & body placements fitted against photographic masks, with retained and refused moves \\
\code{repose}$^\dagger$  & one rigid body moved to where new calibrated RGB-D views place it: a split-half-checked depth proposal, admitted or refused by the pose rules \\
\code{inspect\_physics}$^\dagger$ & convex-proxy conflicts of declared rigid bodies against their fixed neighborhood \\
\code{refine\_object}$^\dagger$ & one body modeled by the object specialist and returned as its own label, with the admission receipt \\
\addlinespace
\multicolumn{2}{@{}l}{\textit{Diagnosis, articulation, termination}}\\
\code{audit}             & a ranked part-discrepancy work list with a paste-ready action per item \\
\code{analyze}           & the output of bounded read-only code over registered evidence, outside the scene \\
\code{joints}            & read-only geometric sweeps of source-declared joints: admitted, refuted, or undeterminable \\
\code{finish}            & a completion request; in two-phase runs, the first request returns an assembly and audit review \\
\bottomrule
\end{tabularx}
\end{table*}

\paragraph{One agent, one live scene.}
\label{sec:scene-audit-trajectory}
The scene agent writes Blender programs in a live, owned Blender through the
\MCP{} surface of Table~\ref{tab:verbs}. It looks at and measures the
observations, writes geometry, compares the result with the photographs through
the real cameras, moves named quantities and bodies until they agree, and
diagnoses what remains. Concretely, it measures named things with SAM~$3$
(\code{measure}), scans the whole scene against every frame's depth for missing
and extra geometry (\path{scan_discrepancies}), asks independent critics to
review photograph/render panels (\path{review_views}), and checks declared rigid
bodies for conflicts (\path{inspect_physics}, Section~\ref{sec:physics}). Runs
of this route use \code{gpt-6-sol} at high effort, with a
$10$\,$800$\,s wall-time limit.
The \code{audit} work list (Section~\ref{sec:mcp-audit}) ranks misshapen parts
by their photograph error and fits constants that only move a part before
constants that deform it.

\paragraph{Background and object bodies.}
Each \code{GROUPS} body declares \code{"role": "background"} or
\code{"role": "object"}. Scene \code{refine} fits background constants and body
poses; a constant whose effect reaches an object-role body is excluded, and the
receipt names the \path{refine_object} call that models that body.

\paragraph{Handing a body to the specialist.}
\label{sec:scene-object-workers}
\path{refine_object(target, brief)} hands one body, of either role, to the object
route (Section~\ref{sec:object}); a body linked by \code{JOINTS} goes with its
jointed component. The harness packages the body's evidence:
\begin{itemize}
\setlength{\itemsep}{1pt}\setlength{\parskip}{0pt}
\item \emph{Masks.} A stored SAM~$3$ mask binds to the body when at least half of
its visible pixels fall inside the mask and the body is the mask's largest
owner. SAM~$3$ measures frames where the body covers at least $1\%$ of the image and
no mask binds.
\item \emph{Occluders.} Nearer scene-model surfaces and other photograph masks.
\item \emph{Views.} Crops are turned upright with $K_{\mathrm{up}}=AKR^{\top}$ and
ordered greedily by coverage${}\times\min(1,\theta/30^\circ)$, with $\theta$ the
smallest viewpoint angle to the views already chosen.
\end{itemize}
The specialist writes a fresh program; the blockout is a hint, used as the
starting source only when every member is exactly its box.

\paragraph{Bringing the asset back.}
The asset returns as its own label: names carry the label prefix, \code{GROUPS}
rows declare \code{role: object}, and \code{JOINTS} stay in the object frame under
one label-wide transform, which \path{joints_of} applies to origin and axis
together. Integration runs in a separate staging runtime restored from the
parent's saved world; only when every check below holds does the parent apply
the same change:
\begin{itemize}
\setlength{\itemsep}{1pt}\setlength{\parskip}{0pt}
\item \emph{Pose, per degree of freedom.} $z$ and scale are never adopted,
flagged corrections are not adopted, and $x$, $y$ and yaw go through the scene's
own pose admission.
\item \emph{Joints.} Unresolved joints are refused; a joint whose only unobserved
field is its range keeps the declared range, labeled as a prior.
\item \emph{Parity and preservation.} The delivered mesh must match (at most
$10^{-6}$\,m per vertex, the same triangles and material names), and unowned
objects must be preserved. A failed apply is rolled back.
\item \emph{Quality.} Integration compares the delivered asset with the
blockout on the body's bound views (silhouette IoU against each view's bound
mask) and admits it only if it is no worse: its mean IoU over those views does
not fall, and its weakest-view IoU falls by at most refine's worst-view
tolerance ($0.02$); otherwise the call returns \code{rejected} with the failing
clause.
\end{itemize}

\paragraph{Concurrency, refusals, and budget.}
Several calls in one reply run concurrently; integration is serialized in
completion order and re-resolves the bodies on the label's current version
(\code{conflict} when an earlier one retired or regrouped them). Besides
\code{conflict}, a call can end \code{failed}, \path{split_required},
\code{rejected} (a failed admission check, named in the receipt) or
\code{invalid}. In this route each delegation reserves $16$ child calls from an
allowance of $48$, and unused calls are refunded. An optional
harness pass (\path{harness_delegation}, off by default) sends the top hints at
the first finish: non-background bodies ranked by visible
pixels${}\times(1-\mathrm{IoU})$.

\paragraph{Visual feedback.}
The input views' recorded cameras render the complete assembled scene. Matched
capture/render views expose omitted entities, incorrect extents, misplaced
instances, and interface errors. Original camera calibration and display-upright
transformations remain separate; lighting changes are explicit so appearance
differences can be interpreted alongside geometry.

\paragraph{Instances and export.}
Repeated instances declared in one \code{GROUPS} prototype row share one
construction, each with its own rigid pose. Finalization saves the native scene
and a GLB; a separate export writes USD, which may approximate procedural materials.

\subsection{Object modeling and numerical refinement}
\label{sec:object}
\label{sec:reconstruction-loop}

The object lane is both a standalone route for single-object benchmarks and the
specialist the scene delegates to (Section~\ref{sec:scene-object-workers}), so it
must work from very different evidence while keeping one implementation.
Object reconstruction alternates between inspecting photographs, editing
source, and inspecting native candidates. Named dimensions and shared
expressions preserve relationships that would be lost by adjusting unrelated
mesh vertices. Numerical fitting searches those controls; a missing profile,
part, or construction dependency requires a source edit.

\subsubsection{Evidence and camera profiles}
\label{sec:object-evidence}

\paragraph{Photograph-driven sessions.}
The workspace retains task photographs and their evidence bindings, the agent's
candidates, native previews, and explicit selection; ground-truth programs and
meshes remain evaluation inputs only. Masks are inferred or taken from image
alpha with the rule recorded, and camera convention, mask provenance, and
display orientation belong to one of three input profiles:
\begin{description}
\setlength{\itemsep}{1pt}
\item[3DCodeBench.] Four photographs with declared orbit cameras of unknown
distance and look-at. For each built candidate the harness fits one shared
camera distance and look-at together with the object's yaw against the bound
silhouettes; the modeled scale and position fix the gauge, and near-tied yaws are
resolved toward the declared front with the tie reported. This registration
recovers neither physical scale nor true orientation.
\item[PartNet.] Three RGB views of an articulated object without stated
cameras. 
\item[Calibrated specialist.] How the scene reaches this lane:
\code{refine\_object} packages one body's evidence (Section~\ref{sec:scene-object-workers}) and
states every view's pinhole camera in the object frame, so there is no yaw
registration or rig fit. By default a rigid pose correction ($x$, $y$, yaw) is
fitted per preview and refinement round and reported, never written into the
source.
\end{description}
Depth serves only as occluder evidence: outside the mask, a pixel nearer than
the rendered candidate counts as neither over- nor under-coverage, and such
pixels inside the mask are only reported. Views whose masks are known to be cut
or contradicted are excluded from search and admission and reported.

\subsubsection{Source editing, previews, and selection}
\label{sec:object-transactions}
The object session exposes source creation and exact-anchor editing, native
preview, candidate selection, and finish. A session is one persistent
episode, and source or materialization failures return diagnostic feedback. A
preview is not a selection: candidate identity and selection are tracked
separately. A continuation run starts a fresh episode whose baseline is the
previous run's selected asset.

\subsubsection{Tensor search and native materialization}
\label{sec:object-proposals}
For source $S$ and selected constants $\theta$, the twin produces
\begin{equation}
 (V(\theta),F,\mathcal{S})=\operatorname{Trace}(S,\theta),
 \label{eq:trace}
\end{equation}
where $\mathcal{S}$ records vertex ownership. Supported operations replay in
Torch, nvdiffrast rasterizes batches~\citep{laine2020nvdiffrast}, and shared
expressions remain live. Baseline twin/Blender parity (bounding boxes within
$5$\,mm, silhouette IoU at least $0.97$) precedes search, and Blender materializes
baselines and selected candidates, not every sampled parameter vector
(Section~\ref{sec:sdk}). The twin replays bevel and weighted-normal modifiers; a program with any other modifier is refused (\code{TwinUnsupported}) and keeps the values its author wrote.
Search is bounded cross-entropy
sampling~\citep{deboer2005crossentropy} over silhouette, contour, and
applicable depth and interior terms, with admission's clauses priced as hinges.
Constants with a declared \code{PARAMS} range are searched within it; others
use the SDK trust interval $v\pm\max(0.25|v|,\,25\,\mathrm{mm})$. Branches the
source takes on a searched constant are recorded at calibration, each constant
is bounded to the interval that keeps them on their calibrated side, and a
constant that cannot move without flipping one keeps its value and is reported.


\paragraph{Scene lane.}
The scene lane extends the variables of Eq.~\eqref{eq:program-refinement}, which
fixes $T$ and the cameras. One twin search can carry a label's constants, the
$x$, $y$, and yaw of its declared rigid bodies (part of $T$; by default within
$\pm0.2$\,m and $\pm20$$^\circ$ of the built pose), and, on request, per-frame
camera corrections (within $\pm2$$^\circ$ and $\pm3$\,cm per axis), so shape,
pose, and cameras are fitted jointly by the same search. \code{refine} also accepts two
whole-source forms. \code{label:$\langle$label$\rangle$} fits every named
constant of a label in one call, grouped by the members each one moves; without
explicit targets, the harness binds masks to the label's \code{GROUPS} bodies,
and inline geometry numbers are named first.
\code{literal:$\langle$label$\rangle$:$\langle$line$\rangle$:$\langle$col$\rangle$}
first names one inline number. A hoist is kept only if the rebuilt objects match the replaced
ones to $10^{-9}$\,m in names, types, world matrices, vertices, triangle counts,
and material slots; otherwise the previous source is re-run and the hoist is
refused. Both forms keep every touching pair touching. Constants whose twin
reach moves a body declared as an object are excluded and left to
\code{refine\_object}.

\paragraph{Object lane.}
The object refiner reuses the same implementation with the builder vocabulary
and photograph masks, and searches constants only. With estimated orbit
cameras, each round first fits the rig to the retained build and then searches
the constants under those cameras, for up to three rounds while a round admits
a change or widens a range; at stated cameras only the pose correction is
re-fitted. A constant that ends within $2\%$ of an end of its searched range
receives one widened interval in the next round, twice the span, centered on the
bound it reached, and the widening never changes sign; constants still at a bound
after the last round get one more round re-centered at their value. A declared range that
no longer brackets an admitted value is re-bracketed, and a changed program
becomes a new native candidate.

\paragraph{Local reach and representation changes.}
Constants whose traced reach touches no target are not searched, and each
effect is split into translation and deformation (Section~\ref{sec:mcp-audit}).
These local diagnostics do not prove that a family can represent the surface;
adding a bend, profile, or connection remains a source operation.

\paragraph{Well-posed fitting.}
\label{sec:object-observability}
The feasible set of Eq.~\eqref{eq:program-refinement} admits only what the views
determine. The reported object runs keep the fit well-posed with four rules:
\begin{itemize}
\setlength{\itemsep}{1pt}\setlength{\parskip}{0pt}
\item \emph{Identifiability.} A view sees a constant when a step of it
changes the visible surface in at least $\tau=20$ pixels of that view's search
raster: a surface appears, disappears, or crosses the image. The step is at most
a tenth of the constant's own value, taken on either side of it. A search over
constants with visual evidence holds each constant that no view sees at its value, at stated and
at fitted orbit cameras alike, so it keeps its measured or assumed value;
constants the modeling agent names are searched regardless. The scene's searches
over those constants of a label read each direction separately and hold a
constant seen on one side only on the other side.
\item \emph{Admissibility.} A range the harness gives never crosses zero.
\item \emph{Coarse to fine.} At fitted orbit cameras, a build turned away from its
declared front ends the pass before any search, and the turn is reported to the
agent.
\item \emph{Parsimony.} After a search, each changed constant is set back
alone; a change whose removal costs the twin's mean silhouette IoU less than
$0.002$ returns to its retained value.
\end{itemize}

\subsubsection{Candidate retention and delivery}
\label{sec:object-admission}
Numerical admission compares a realized candidate with the retained baseline
under the same evidence, so a fit is kept only when the photographs, not the
optimizer's loss, say it is better. Both lanes use the tolerances owned by the
SDK's admission module (\code{neosdk.refine.twin.admission}):
\begin{itemize}
\setlength{\itemsep}{1pt}\setlength{\parskip}{0pt}
\item the mean silhouette IoU over bound target--view pairs rises by at least
$0.002$;
\item the weakest view falls by at most $0.02$, and any overlapping view by at
most $0.05$;
\item the remaining contour, depth, scene, and mesh-stretch clauses hold where
measured.
\end{itemize}
The object lane drops only the scene clauses (void ratio, depth nearer/farther
fractions, per-mask coverage), which protect a moved part's surroundings. Refusal preserves the retained state and records the
failing clause; twin failures and budget exhaustion remain separate outcomes.

Selection is explicit and belongs to the session. If the budget ends with
nothing selected, the harness selects the latest completed native preview and
records itself as the selector (\code{harness\_budget\_fallback}); motion
artifacts and their fallback are described in
Section~\ref{sec:object-articulation}. The 3DCodeBench delivery re-executes the selected program in a
clean Blender and checks that it rebuilds the selected native geometry;
executable export does not establish agreement with unobserved geometry.
Integration of a delegated object back into a scene is described in
Section~\ref{sec:scene-live}.

\subsection{Physics, articulation, and geometry validity}
\label{sec:physics}
\label{sec:simulation-features}

\model{} separates link-level geometry inspection, declared-joint checks, and
simulator-specific physicalization. These stages address mesh connectivity,
kinematic consistency, and interaction under configured collision and body
properties. Scene finalization proposes bounded rigid corrections while
protecting the visual reconstruction; articulation routes inspect declared
motion; the SDK supplies physical-carrier operations.
Physics-in-the-loop is enabled in the scene pipeline (WebUI and CLI scene
tasks), while articulation stays optional. Figure~\ref{fig:geometry-articulation-detail}
maps the geometry and articulation paths. Table~\ref{tab:physical-routes}
summarizes their interfaces and configurations.

\begin{table}[t]
\centering
\small
\caption{
    \textbf{Physical-asset interfaces and validation checks.}
    Summary of the four physical-asset routes, their input/output
    representations, and the corresponding validation procedures
    and configuration options.
}
\label{tab:physical-routes}

\setlength{\tabcolsep}{7pt}
\renewcommand{\arraystretch}{1.18}

\begin{tabularx}{\linewidth}{
    @{}
    >{\raggedright\arraybackslash}p{0.19\linewidth}
    >{\raggedright\arraybackslash}p{0.28\linewidth}
    >{\raggedright\arraybackslash}X
    @{}
}
\toprule
\textbf{Route}
& \textbf{Input and Output}
& \textbf{Checks and Configuration} \\
\midrule

\textbf{Visual kinematic export}
& Declared links and joints $\rightarrow$ visual URDF
& Joint sweeps and native/URDF playback agreement;
visual meshes and kinematic frames. \\

\addlinespace[6pt]

\textbf{Benchmark physics export}
& Delivered scene $\rightarrow$ USD with UsdPhysics
& Rest contact, body properties, and native joints. Indexed
closure checks and repair are enabled by default (\code{--watertight});
wedge collision decomposition is optional (off by default). \\

\addlinespace[6pt]

\textbf{Robot twin}
& Reconstructed parts $\rightarrow$ configured simulator world
& Collision proxies, body properties, and robot-frame
alignment, as specified in Section~\ref{sec:robotics}. \\

\addlinespace[6pt]

\textbf{SDK solid adaptation}
& Metric indexed mesh and adaptation plan
$\rightarrow$ certified physical carriers
& Closure, manifoldness, self-intersection, orientation,
and positive volume. \\

\bottomrule
\end{tabularx}
\end{table}


\subsubsection{Bounded rigid-body scene correction}
\label{sec:scene-physics}

The scene pipeline enables physics-in-the-loop with an explicit task switch to
disable it; the choice is frozen at launch. Source declarations distinguish rigid,
fixed, and deformable groups; unassigned or deformable geometry in a reachable
contact neighborhood remains unresolved rather than silently becoming fixed. The
modeling agent can inspect conflicts, and the end of each build invokes a bounded rigid candidate
transaction that derives minimum-motion proposals in the contact neighborhood.
Original-view masks, calibrated depth, finite support, and native/twin parity are
prerequisites; visual fit is a constraint rather than a reason to open a larger
gap. An owned Isaac batch returns complete SE($3$) poses, contact identities, and
late-window motion under configured assumptions ($1$\,kg per body, friction $0.5$, no
restitution). Shortlisted poses undergo native geometry, protected-view, support,
and attachment checks; admitted deltas are replayed before saving and rendering,
and rejected candidates restore the prior state. Missing evidence, unavailable
simulation, or budget exhaustion stays unresolved, and the disabled path starts no
physics process.

\subsubsection{Scene articulation and kinematic export}
\label{sec:scene-articulation}
\label{sec:geometry-articulation-detail}

In a scene, articulation lives in the executable source. Named \code{GROUPS}
define bodies, and \code{JOINTS} declares each joint's parent, child, type,
origin, axis, limits, and rest value $q_0$, with links naming stable bodies rather
than positional identifiers or object globs. Unknown origins, axes, and limits can
remain unset. Assembly couples joint-connected bodies for placement so that
independent pose updates do not tear their declared relation. Joints of a
delegated body arrive with the returned label (Section~\ref{sec:scene-object-workers});
the scene retains sweep, export, and playback authority.

\paragraph{Geometric screening of joint declarations.}
The live scene stays at $q_0$. The read-only \code{joints} tool transforms samples
of exported geometry over the declared range, checking for a rest interface,
observable displacement, retained attachment, parent clearance, and clearance
from nearby scene objects. Intrusion is measured by ray parity into closed solids
with a depth margin. The tool records per-check outcomes and measurement
availability; admission requires that no measured check refute the declared
motion. Admission records this screening outcome; the supplementary penetration
evaluation counts a pass only for fully measurable geometry
(Appendix~\ref{sec:articulation-penetration}). For a refuted range the tool can
propose a source edit using the
interval the sampled geometry supports; that interval describes geometric
clearance, not a mechanical limit, and the modeling agent owns the revision.

\paragraph{Visual kinematic export and playback.}
At finalization, components with unsettled or refuted declarations are excluded
from the kinematic export, and the exporter never narrows a range. For eligible
components it writes link meshes and a visual URDF, with a recorded transform
restoring the component to the room. The child's link frame is the joint frame,
and $q_{\mathrm{URDF}}=q-q_0$. A separate native playback of the delivered scene
produces link-pose receipts, and forward kinematics parsed from the written URDF
must reproduce them; a failed round trip excludes the file from the manifest.

\begin{figure*}[t]
\centering
\includegraphics[width=\textwidth]{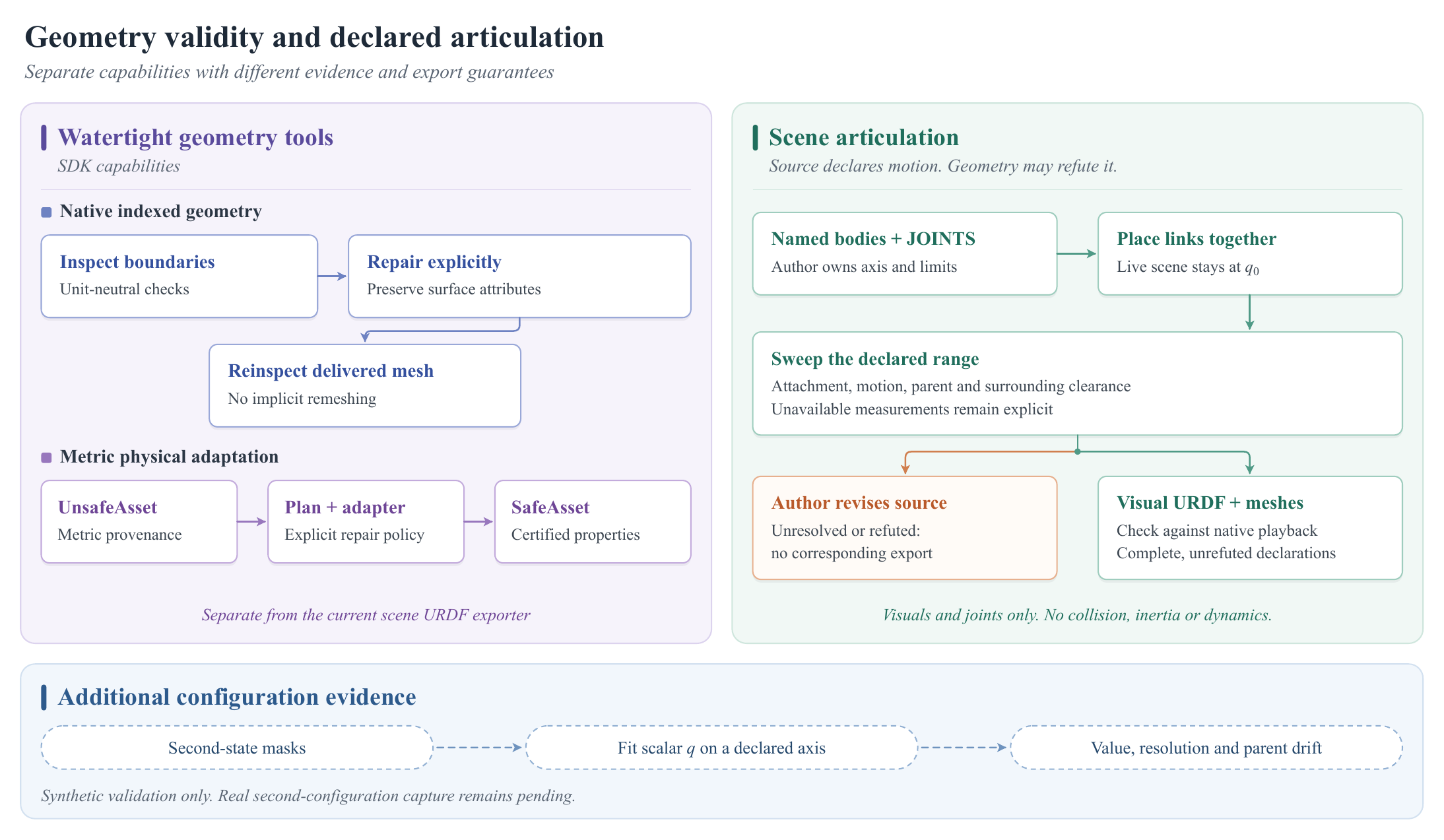}
\caption{\textbf{Geometry validity and declared articulation.} Left: SDK
boundary inspection and explicit surface-preserving repair, distinct from
metric adaptation and certification. Right: source-owned joint declarations,
geometric refutation, and visual kinematic export checked against native
playback. Sweep records distinguish measured checks from unavailable
measurements. Bottom: scalar joint-position
fitting on a declared axis has synthetic validation, with real
second-configuration observations still pending.}
\label{fig:geometry-articulation-detail}
\end{figure*}

\subsubsection{Object articulation and opened-state review}
\label{sec:object-articulation}

The object lane, including the scene specialist when a scene task enables
articulation, declares a literal joint tree on a built candidate and probes it
natively with \path{probe_articulation}. Each probe places the harness's own
review cameras, renders every joint at $0$, $0.5$ and $1$ of its range, and samples
kinematic geometry: each joint alone over $91$ samples of its range with the other
joints at $q_0$, then all joints together at $0.5$ and $1.0$ of their ranges.
\path{review_opened_state} shows these facts with bounded images. The findings are summarized in Table~\ref{tab:articulation-probe-findings}.

A decision is either a source fix or \path{record_articulation_unknown} naming
the finding kind and link. A colliding sweep is deferred as
\path{sequenced_after} only when every colliding link is moved by other joints,
the sweep is clean with those joints at their open limit, the review carries the
combined state, and no combined collision involves this joint.

\path{select_motion_artifact} admits only a probe without refusals or open
findings and also selects that probe's candidate under the ordinary candidate
rules. An explicit motion-artifact selection is never replaced. When an episode
ends without a selected motion artifact, the harness selects the latest
admissible probe, preferring the selected candidate's; when the selected
candidate has none, it may select another candidate's latest admissible probe
together with that candidate, under the ordinary selection rules. It records
\path{selection_origin} as \path{harness_budget_fallback},
\path{harness_end_fallback} or, after lost calls,
\path{harness_transport_fallback}. Each joint records family, side, axis and
range as observed or not observed; unobserved fields stay labeled hypotheses.
For rest-state photographs, the standalone articulated-object brief lists priors for
closed fronts, mined from $525$ held-out PartNet-Mobility objects ($1{,}031$ panels,
split $70$/$30$ by object, with the benchmark's $77$ objects excluded by a guard). The rule list
scores $0.8990$ on the held-out split, and a side hinge lies opposite an off-center
handle in $308$/$308$ cases. The priors are guidance only; no harness gate reads them.
Sampled facts are kinematic geometry, not physics or continuous collision, and
admission does not establish the real mechanism.

\paragraph{Delivered motion artifact.}
The selected artifact is returned with an articulated native file at rest and,
when every joint is stated in the object frame, a kinematics-only URDF. The
exporter keeps rigid followers as fixed-joint links; merging each rigid body into
one part happens only in a separate, evaluator-facing tree. A frame change moves a
joint origin by the full affine map and its axis by the linear part, renormalized;
a uniform scale scales prismatic ranges but not revolute ones.

\begin{table}[t]
\centering
\small
\caption{
    \textbf{Articulation probe findings.}
    Classification of findings produced by kinematic validation.
}
\label{tab:articulation-probe-findings}

\setlength{\tabcolsep}{6pt}
\renewcommand{\arraystretch}{1.18}

\begin{tabularx}{\linewidth}{
    @{}
    >{\raggedright\arraybackslash}p{0.19\linewidth}
    >{\raggedright\arraybackslash}p{0.29\linewidth}
    >{\raggedright\arraybackslash}X
    @{}
}
\toprule
\textbf{Class} & \textbf{Finding} & \textbf{Description} \\
\midrule

\textbf{Refuses the probe}
& \path{rest_overlap}
& Overlapping geometry at rest. \\

& \path{collision}
& Collision during the per-joint sweep. \\

& \path{closure_gap}
& Closure gap at $q_0$. \\

& \path{combined_collision}
& Collision between links moved apart by two or more joints. \\

& \path{detached_in_sweep}
& Detachment observed during the sweep. \\

\addlinespace[5pt]
\textbf{Needs a decision}
& \path{cover_without_cavity}
& A covering link does not open onto a cavity. \\

& \path{travel_leaves_parent}
& A sliding link leaves its parent at full travel. \\

& \path{joint_dropped}
& A previously moving joint is removed. \\

\addlinespace[5pt]
\textbf{Report only}

& \path{sequenced_after}
& Sweep deferred due to other joint movements. \\

& \path{weakened_after_report}
& Joint range reduced after an earlier report. \\

\bottomrule
\end{tabularx}
\end{table}

\subsubsection{Physical geometry and mass properties}
\label{sec:unsafe-safe}

\paragraph{Default indexed closure and repair.}
\label{sec:default-closure-policy}
The default export policy checks indexed mesh connectivity and applies local
closure repairs before export.
Each failing mesh is processed in a fixed order, stopping at the first
candidate that passes the closure check: (A) re-triangulate polygons whose
triangulation contains zero-area triangles and make winding consistent;
(B) dissolve zero-length edges and zero-area faces, then repeat (A);
(C) merge vertices within $10^{-6}$ of the object diagonal; and
(D) apply centered solidification to open sheets with thickness $10^{-3}$
of the diagonal. In scenes, the thresholds use each member's own diagonal,
and sheet thickness is capped at $1$\,mm. Steps (A)--(B) retain vertex
positions, and shading normals are restored after (A)--(C). Parts belonging
to independently moving links remain separate. A mesh that no step closes
is retained and recorded. Beveled or extruded curves with open splines
receive end caps before scene mesh conversion. Wedge collision decomposition
is disabled in the default export to retain repaired closure.

\paragraph{SDK boundary inspection.}
Separately, the Rust boundary inspector consumes the native mesh's indexed vertices and
triangles without units or rescaling. It checks closure, manifoldness,
self-intersection, shell containment, and material-boundary orientation (an outer
shell faces outward, a cavity faces into the void), preserving supplied
connectivity and winding; it neither repairs a boundary nor establishes its
real-world dimensions. An explicit surface-preserving repair rejoins connectivity
only where source correspondence or an unambiguous coincident boundary identifies
missing shared indices. It keeps vertex positions, surface attributes and part
boundaries, fills no holes, infers no orientation, and its output must be
inspected again. 

\paragraph{Metric solid adaptation.}
Physical adaptation separately starts from an indexed mesh with explicit metric
provenance, which enters the kernel as an \code{UnsafeAsset} carrying no topology
claim. An explicit \code{AdaptationPlan} selects, per part, solid preservation,
regularization, reconstruction, shell construction, or visual-only treatment, and
a separately built Rust adapter executes it without implicit repair. Validation
yields a \code{SafeAsset} and certificate only when every physical part is closed,
manifold, free of self-intersection, correctly oriented as a material boundary,
and of positive volume; explicit metric provenance is required for certification.

\paragraph{Mass properties.}
The three execution paths configure their mass models as follows.
\begin{itemize}
\setlength{\itemsep}{1pt}\setlength{\parskip}{0pt}
\item \emph{Benchmark exporter:} mass, inertia and principal axes from the
delivered convex pieces at a uniform $1{,}000$\,kg/m$^3$, with each body's origin at
their volume centroid.
\item \emph{Benchmark evaluator:} ignores these properties and gives each body
the mass of its bounding box at its own $500$\,kg/m$^3$, with the center of mass at
the prim origin, so of the exported properties only the centroid origin affects
its scores.
\item \emph{Physics-in-the-loop:} $1$\,kg per body (Section~\ref{sec:scene-physics}).
\end{itemize}

\paragraph{Benchmark-side physics export.}
A method-side adapter writes the delivered scene as USD with UsdPhysics:
\begin{itemize}
\setlength{\itemsep}{1pt}\setlength{\parskip}{0pt}
\item \emph{Roles.} Declared roles map \code{fixed} to static and every other role
to dynamic, and each simulated body's origin moves to the volume centroid of its
delivered convex pieces.
\item \emph{Rest contact.} A hovering dynamic body is lowered onto its support by
at most $0.05$\,m; when an overlap test fires, the shift is bisected so the body
ends above that support. A body that already penetrates a neighbor, or has no
support within that distance, is never moved and is listed instead.
\item \emph{Collision decomposition (optional, off by default).} A bottom band
($2$\,mm by default) of each supporting component can be split into $N$ wedges,
with $N$ the smallest even number $\geq 4$ whose wedge rim keeps $90\%$ of the
footprint's inscribed margin. The default export leaves this decomposition
disabled to retain the members' repaired closure.
\item \emph{Joints.} Settled, unrefuted joints become native revolute or prismatic
joints with limits written relative to $q_0$, since the delivered geometry sits at
$q_0$. A
method-side audit sweeps each joint's range and reports link interpenetration of
the convex pieces.
\end{itemize}

\subsection{Predicted depth and scale alignment}
\label{sec:predicted-depth}

The RoboCasa inputs are RGB views only: the cameras are estimated from them
with VGGT-$\Omega$ (Appendix~\ref{sec:scene-measurement}) up to one unknown
global scale, and there is no sensor depth. Measurement and the depth scan need
metric depth, so predicted depth fills this gap; it must first be made
consistent across views, and it also supplies the metric scale the estimated
cameras lack.

\paragraph{MoGe-3 frame producer.}
The MoGe-3 runner predicts camera-$z$ depth and validity from RGB, conditioned on
the horizontal field of view, and records the model identity, package
version, checkpoint, resolution and precision. The writer accepts RGB-only frames
and passes supplied camera poses through. For a declared upright quarter-turn the
model sees the turned image and its depth is turned back into the calibrated raw
grid. Each frame keeps the model's own metric scale, so frames may disagree; the
rasters are declared \emph{predicted depth}, not sensor measurements.

\paragraph{Per-view scale fit.}
\label{sec:depth-scale}
Given camera poses in metric units (here VGGT-$\Omega$ estimates after the scale
alignment below), one positive scale $s_j$ per view (and optionally
a shift $t_j$) is fitted to reprojection consistency. A pixel of view $i$ lifted at
depth $s_i z_i$ has camera depth $\zeta_{ij}$ in view $j$, and the residual is
$r=\log\zeta_{ij}-\log\big(s_j z_j(x_{ij})+t_j\big)$, with $z_j(x_{ij})$ view $j$'s
raw depth at the projection. The loss is asymmetric and truncated: an occluded
point ($r>\tau$) costs the constant $\tau^2$, a see-through point ($r<-\tau$) grows
linearly, and residuals in between are quadratic. The truncation $\tau$ is annealed
from $0.3$ to $0.02$ in log depth. Each connected group's common scale is found by a
grid search at the widest $\tau$ before descent, and pixels on depth
discontinuities are neither sampled nor read. 
Estimated poses carry no metric scale: their unit is first converted to meters
by the median, over views, of predicted metric depth against the pose model's
own depth, and each group's gauge is pinned to the median predicted-depth
scale. With metric poses, such as a calibrated capture, the camera baselines
fix the gauge instead.
Views without usable overlap stay unfitted, and held-out residuals measure
cross-view consistency, not accuracy. 
The benchmark configuration is summarized in
Appendix~\ref{sec:appendix-evaluation-baselines}.

\clearpage
\section{Robotic manipulation and deployment}
\label{sec:robotic-deployment}

\subsection{Connecting reconstructed worlds to Real-Sim-Real robotics}
\label{sec:robotics}

\paragraph{What the reconstruction hands over.}
The robotics role of \model{} is to provide a simulation-ready world
for downstream robot planning and control.
The reconstruction retains editable geometry programs, explicit part
and instance identities, object-to-scene transforms, camera information,
materials, and declared articulations.
These representations allow the robot system to associate visual
observations with reconstructed objects and update their poses
without rebuilding the underlying geometry.
The reconstructed world serves as the geometric foundation of a
digital twin, while robot-specific collision models and physical
properties are configured separately.

\paragraph{Frames and metric registration.}
Real-Sim-Real transfer requires the reconstruction, robot, and
sensor observations to share a consistent metric coordinate frame.
When known-reference calibration is available, scene geometry,
camera translations, and depth are rescaled together, while
instance transforms retain their relationships to the scene.
For physical deployment, a fixed calibration board observed by
both the scene capture system and the robot's wrist RGB-D camera
establishes the transformation between reconstruction and robot
coordinates.
The robot base is anchored using the calibrated hand--eye
transformation rather than relying solely on its nominal mounting
pose.
On our physical platform, hand--eye calibration from $20$ board
observations achieves a board-pose consistency of $2.1$\,mm and
$0.22^\circ$.
This calibration establishes a common frame for robot motion,
reconstructed geometry, and subsequent visual updates.

\paragraph{Physical twin construction.}
The robot twin is represented as
$\mathcal{W}=(\mathcal{G},\mathcal{X},\mathcal{C},\mathcal{A},\mathcal{P})$,
where $\mathcal{G}$ denotes reconstructed visual geometry,
$\mathcal{X}$ contains part and instance transforms,
$\mathcal{C}$ represents collision geometry,
$\mathcal{A}$ specifies declared articulations, and
$\mathcal{P}$ contains configured physical properties.
Visual geometry preserves appearance, whereas collision proxies
support geometric clearance and physical simulation
(Section~\ref{sec:unsafe-safe}).
Articulation relationships are retained where available
(Section~\ref{sec:scene-articulation}).
Mass, friction, and actuator properties are configured for
the target simulator rather than inferred from appearance.
The reconstructed environment and robot model are assembled
into a MuJoCo twin for motion planning and collision checking.

\paragraph{Updating the twin from wrist observations.}
The physical environment may differ from its initial reconstruction
because of object movement or residual calibration error.
To update the twin, the robot acquires calibrated RGB-D observations
from collision-checked wrist-camera viewpoints.
Object masks and depth measurements provide geometric evidence
for re-estimating object poses within the reconstructed scene.
These updates modify instance transforms while retaining the
underlying construction programs.
In our physical experiment, an eight-view wrist scan supports
pose refinement of the bowl and cube in $1.4$--$2.2$\,s.
Their mean silhouette IoU increases from $0.81$ to $0.86$ and
from $0.71$ to $0.86$, respectively.
An inconsistent pose proposal for the microwave is rejected
rather than applied to the twin.
This selective update mechanism incorporates new observations
without unnecessarily modifying reconstructed geometry.

\paragraph{Planning and real-world transfer.}
Robot motions are planned and collision-checked against the
reconstructed MuJoCo environment before hardware execution.
The physical platform consists of two AgileX Piper arms equipped
with wrist RGB-D cameras and, for the comparison of
Section~\ref{sec:val_rsr}, a fixed RGB-D camera.
For pick-and-place manipulation, the twin provides obstacle
geometry and object poses for inverse kinematics and
collision-aware trajectory planning.
Before grasping, wrist-camera inspection further corrects
the target pose using local visual evidence.
In the single-arm cube-placement experiment, this inspection
corrects the estimated cube pose by $5.3$\,mm and $6.7^\circ$,
with a median surface residual of $1.8$\,mm.
The planned motion is then executed on the physical robot,
with observed joint tracking errors within
$0.18$--$0.33^\circ$.
The streaming executor operates at $100$\,Hz and checks
joint tracking, motor current, temperature, and
workspace safety constraints throughout execution.

\paragraph{Contact-rich bimanual manipulation.}
We further evaluate the reconstructed twin through a chained
bimanual manipulation task involving a cube, a bowl, and
a microwave.
One arm opens the microwave while the other places the cube
into the bowl; the bowl is subsequently transferred into
the microwave, and the door is closed.
The door-opening controller uses an admittance-based pull
that adapts its direction from estimated contact forces,
without requiring an accurate hinge model.
On the physical microwave, the latch releases at
$10$--$15$\,N, after which the pulling force drops to $0$--$2$\,N
as the motion direction changes with the opening door.
The resulting sequence combines twin-based geometric
planning, visual pose correction, and compliant physical
interaction.

\paragraph{Scope of physical validation.}
These experiments demonstrate that reconstructed geometry,
metric registration, and updated object poses can support
collision-aware planning and execution on physical robots.
The robot twin provides the geometric and kinematic
constraints needed for motion planning, while local visual
inspection and contact feedback address discrepancies
encountered during execution.
The experiments in this section are development demonstrations
with a limited number of physical trials; task success rates come
from the $120$-trial comparison of Section~\ref{sec:val_rsr}
(Appendix~\ref{sec:robotics-eval}).
Physical properties remain configured assumptions, and
successful execution does not establish that all underlying
contact dynamics have been recovered from visual observations.

\subsection{Physical registration and re-posing from wrist scans}
\label{sec:robotics-real}

A reconstructed twin supports physical robot manipulation when the robot,
the reconstruction, and the latest sensor observations share a consistent
metric coordinate frame.
This subsection describes physical registration, collision-aware wrist
scanning, and object pose refinement on the right arm of the platform in
Figure~\ref{fig:real-sim-robotics}, an AgileX Piper equipped with a wrist
RGB-D camera.
Motion planning and collision checking use the MuJoCo twin built from
the reconstruction.
The reported measurements are obtained from a physical tabletop
environment containing a bowl, a cube, and a microwave.

\begin{figure}[t]
\centering
\includegraphics[width=0.86\linewidth]{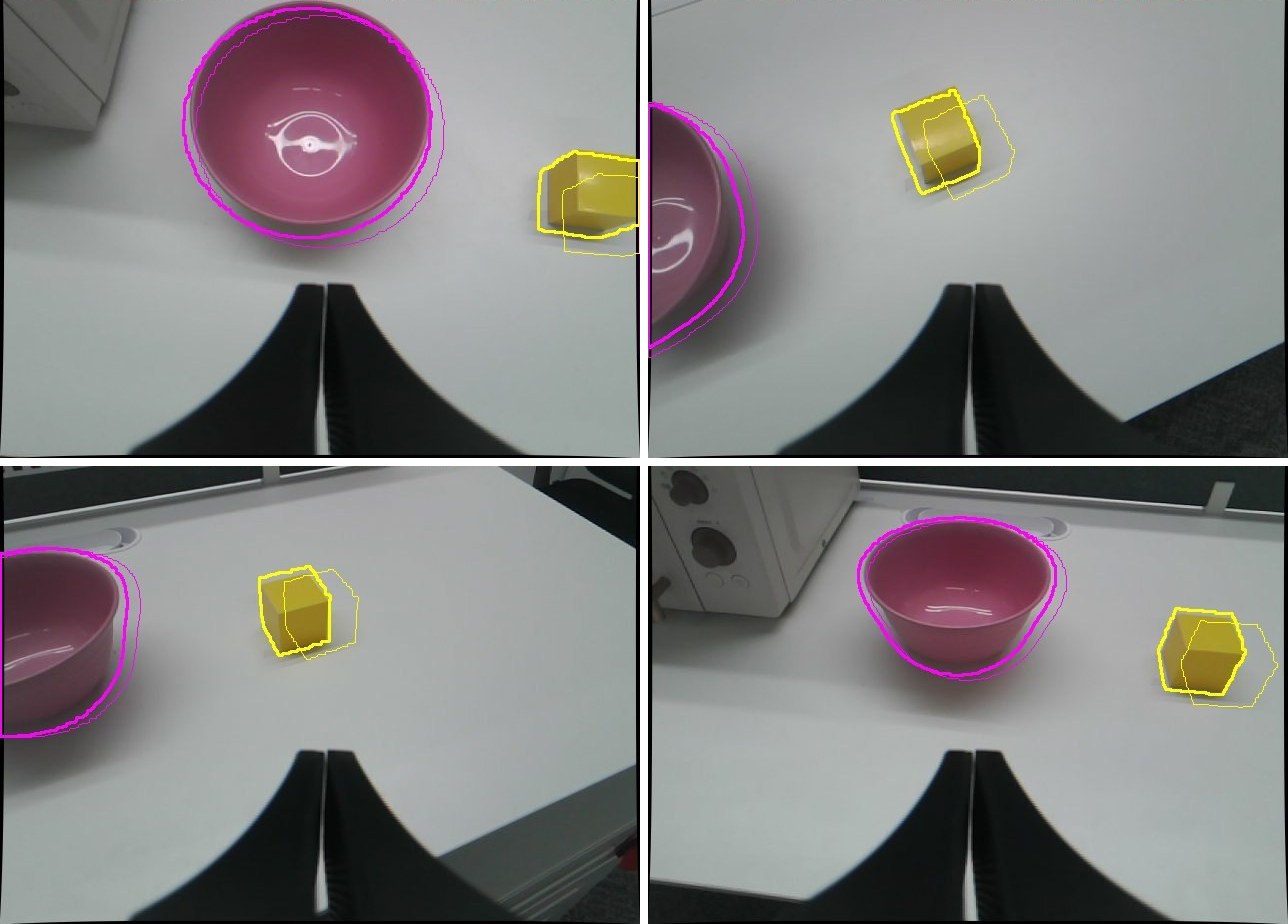}
\caption{\textbf{Re-posing a finished reconstruction from a wrist scan.}
Outlines of the MuJoCo twin's bodies before (thin) and after (thick)
\code{repose}, projected into four of the eight undistorted wrist RGB-D views
in raw sensor orientation (magenta: bowl; yellow: cube; no occlusion between
bodies). The gripper fingers at the bottom of each view are the declared
occluder. Both twins use the board-anchored base, so the outlines differ only
by the admitted moves of Table~\ref{tab:repose}.}
\label{fig:repose-overlay}
\end{figure}

\paragraph{Coordinate frames.}
Write $T^{a}_{b}$ for the rigid transform from frame $b$ to frame $a$,
$F(q)=T^{\mathrm{base}}_{\mathrm{flange}}(q)$ for the arm's forward
kinematics at joint angles $q$, and $X=T^{\mathrm{flange}}_{\mathrm{cam}}$
for the hand--eye transform. A wrist view taken at $q$ has the camera pose
\begin{equation}
T^{\mathrm{scene}}_{\mathrm{cam}}(q)
=T^{\mathrm{scene}}_{\mathrm{base}}\,F(q)\,X .
\label{eq:wrist-chain}
\end{equation}
The scan planner and the calibrated views handed to re-posing use this
expression, with $X$ and $T^{\mathrm{scene}}_{\mathrm{base}}$ each measured
once as described below. The view gate runs before either is known, so it
uses the nominal mount and every available board-pose hypothesis.

\paragraph{Twin-based clearance checks.}
A clearance query asks whether the arm at $q$ comes within a $3$\,cm margin of
the reconstructed scene. It runs forward kinematics and the collision pass
only, and returns every robot--scene geometry pair inside the margin.
The simulator's midphase (MuJoCo 3.3.1) dropped pairs inside the margin
whose bounding volumes did not overlap ($102$ pairs over $4{,}096$ random poses,
the closest at $4.5$\,mm, flipping two clear/blocked decisions), so it is
disabled, which raises the cost of a query from $0.18$ to $0.74$\,ms;
a test requires every pair whose exact distance lies inside the margin
to be reported.

\paragraph{Viewpoint admissibility.}
Before any board-dependent motion, an offline gate decides which wrist
camera can see the calibration board, and from where. Candidate camera
poses face the board from $0.36$--$0.52$\,m, tilted by up to $30^\circ$ from
its normal at eight azimuths and rolled by up to $\pm30^\circ$, and are
aimed so that the board center lands above the gripper fingers in the
image; bounded IK converts each into joint angles inside the device limits.
A candidate passes when (i) the board lattice projects inside the image
and clear of the finger polygon under \emph{every} available board-pose
hypothesis (one from hand--eye samples, one from the scene scan, $99$\,mm
apart before anchoring), and (ii) the arm is twin-clear. Passing candidates
are ranked by the smaller fill fraction times the cosine of the larger
obliquity. The visibility predicate was calibrated against the robot's
own board detector: it accepts $19$ of $23$ detected views and none of $27$
undetected ones. A gate report is used only when its controls hold:
at least $80\%$ of the recorded capture poses are judged visible
($14$ of $15$), a pose at which a real arm once stalled against the scene
registers a collision (it does, at $-15$\,mm), and the arms' current
poses are clear. A board sweep is refused for a camera the gate does
not admit. Of $459$ candidates per wrist, $12$ pass for the right wrist,
while the left wrist passes only in contorted poses, so the chain below
uses the right arm.

\paragraph{Hand--eye sweep planned from an image jacobian.}
From an anchor pose where the whole board is visible, one probe per joint
(base pan, elbow, wrist tilt, and wrist roll; $2$--$4^\circ$) measures how
the ordered board corners move. Each probe is summarized as a planar
similarity, namely a centroid shift $(d_x,d_y)$, an in-plane rotation
$\phi$, and a log scale $\log s$, fitted by complex Procrustes
($b\approx s\,e^{i\phi}a$ for centered corners $a,b$).
The four columns form an image Jacobian
$J_I\in\mathbb{R}^{4\times4}$ that predicts the corners after any
joint offset $\delta$. Offsets on a grid around the anchor are kept
when the predicted board outline, including its paper margin, stays
inside the frame and off the finger keep-out; farthest-point sampling
in joint-offset space, with a $2^\circ$ duplicate floor, then selects
the sweep, which is visited in nearest-neighbor order. The robot's
own detector re-checks each pose live before capture, and the
eye-in-hand solve (Park's method) runs after a person confirms the
board axes. The $20$ samples came from Jacobian-planned sweeps,
supplemented by a few explicitly specified wrist-roll poses for
rotation spread. The fixed board predicted from each sample agrees
within $2.1$\,mm and $0.22^\circ$ ($1.8$\,mm and $0.17^\circ$ on held-out
samples); this measures the self-consistency of the solution,
not its absolute accuracy.

\paragraph{Anchoring the arm base.}
Each hand--eye sample $k$ also measures the board in the arm base,
$T^{\mathrm{base}}_{\mathrm{board},k}
=F(q_k)\,X\,T^{\mathrm{cam}}_{\mathrm{board},k}$.
Averaging the translations and rotations over samples gives
$\bar T^{\mathrm{base}}_{\mathrm{board}}$, and the scene scan measures
the same physical board, so
\begin{equation}
T^{\mathrm{scene}}_{\mathrm{base}}
=T^{\mathrm{scene}}_{\mathrm{board}}\,
\big(\bar T^{\mathrm{base}}_{\mathrm{board}}\big)^{-1}
\label{eq:base-anchor}
\end{equation}
requires no nominal chassis mount. It holds while neither the board
nor the robot has moved since the scan and the calibration session;
the wrist scan below is an independent check. On the development scene,
this anchor lies $44$\,mm and $5.3^\circ$ (yaw) from the earlier estimate,
which fitted the chassis to robot silhouettes in the capture video
and added the nominal mount.

\paragraph{Twin-checked wrist RGB-D scan.}
Candidate views look at each target body from $0.32$--$0.48$\,m, at elevations
of $40$--$70^\circ$ and azimuths within $\pm50^\circ$ of the direction
toward the arm base, with the target aimed above the fingers. IK solves
each view inside the device limits, narrowed to windows around the start
pose ($\pm45^\circ$ for the base joint and $\pm60^\circ$ for the two roll
joints, which protect the camera cable and keep the arm inside the
reconstructed region). A view is kept if its target clears the finger
polygon and the arm is twin-clear. A greedy cover then selects at most
eight views: each surface point of the targets and of the reachable
static scene, sampled from the twin, counts at most twice, and only from
viewing directions at least $25^\circ$ apart. The controller interpolates
in joint space, so every joint-space segment is checked on the twin in
$2^\circ$ steps. The start pose serves as a hub: a kept view needs clear
segments to and from it, views are visited in nearest-neighbor order over clear
segments, and the route detours through the hub when no direct segment
is clear. Execution starts only with the arm within $1^\circ$ of the
plan's start pose, moves with the robot's own joint-move tool (which
rechecks limits and self-collision), captures one registered RGB-D still
per view, and returns to the start. On the development scene the scan
took eight stills in $411$\,s. Each still becomes a calibrated view:
color and registered depth undistorted onto a pinhole model,
the camera pose of Equation~\eqref{eq:wrist-chain}, a depth-scale
correction ($1.022$, fitted to the board plane at $0.47$\,m),
and the gripper region declared as an occluder.

\paragraph{Re-posing inside the finished reconstruction.}
\label{sec:robotics-repose}
The \MCP{} verb \code{repose} (Table~\ref{tab:verbs}) moves one rigid body
of a finished scene to where new calibrated RGB-D views place it.
The finished run is reopened with every label frozen: the latest
program of each label and the objects it created rebuild the ledger,
no program is replayed, and the twin carries each label from its
native export, so pose search is available and constant search is
refused. For each requested body:
\begin{enumerate}[leftmargin=*,itemsep=1pt,parsep=0pt,topsep=2pt]
\item \emph{Bind masks.} The body's surface samples are projected into each
view. A view is dropped when less than half of the body lies in frame or the
declared occluder hides more than half of it. SAM\,3 segments the view with
the body's prompt, and the instance mask with the highest IoU against the
projection is the body's only if that IoU is at least $0.2$ and exceeds the
runner-up's by $0.05$. Occluder pixels are removed from the mask and treated
as known foreground. Every dropped view is recorded with its reason.

\item \emph{Fit what the views can see.} Members lying flat on the support
(top within $6$\,mm of the body's bottom, such as a mat, tape, or decal) are
left out of the fit and of the silhouette target, and still move with
the body.

\item \emph{Depth proposal.} Valid depth inside the eroded masks is lifted
to the scene frame. A near-horizontal support plane is fitted by RANSAC
to a ring of pixels around the masks, and only points more than $6$\,mm
above it are kept, in $3$\,mm voxels. Point-to-plane ICP aligns these
points $p_i$ to the body's surface samples $q_i$ with normals $n_i$ in
the three degrees of freedom of a body resting on a support.
Let $\xi=(x,y,\psi)$ denote the planar pose parameters and
$t=(x,y,0)^\top$ the translation. We solve
\begin{equation}
\label{eq:depth-icp}
\xi^* = \arg\min_{\xi}
\sum_i \left[n_i^\top(\hat{p}_i(\xi)-q_i)\right]^2,
\qquad
\hat{p}_i(\xi) = R_z(\psi)(p_i-c)+c+t.
\end{equation}
The rotation is centered at the body center $c$, with correspondence
radii shrinking from $30$ to $5$\,mm. The inverse of this alignment is the
proposed move. Yaw is folded by the body's declared symmetry
($90^\circ$ for a cube) and is not fitted for a body symmetric
about its vertical axis.

\item \emph{Stability.} The same fit on the even and on the odd views must
agree within $5$\,mm and $2^\circ$ (otherwise \code{unstable\_proposal}).
The move must stay within $8$\,cm and must not reduce the number of depth
points within $5$\,mm of the surface.

\item \emph{Admission.} The proposal enters the twin's pose search as its
first candidate and passes the same admission as every pose candidate:
the silhouette rules of Section~\ref{sec:object-admission}, no deeper
penetration into any neighbor, and no move to the edge of the search
range. Bodies resting on the moved body move with it. When the proposal
is refused, the twin's own pose search may run.

\item \emph{Export.} The scene is saved and exported, the simulator twin
receives the new export, and a free body that the export moves drops
its earlier pose override. Moves of articulated members are refused,
because the twin rebuilds those from the kinematic export.
\end{enumerate}

\begin{table}[htbp]
\centering
\caption{\textbf{Re-posing three bodies from one eight-view wrist scan.}
Proposal: the estimated planar pose update from masked depth observations,
expressed in the scene frame.
IoU: silhouette IoU against the assigned SAM\,3 masks over the views used
for each body, before $\rightarrow$ after.
Yaw is not fitted for the bowl and is folded modulo $90^\circ$ for the cube.
The estimates from the two view subsets for the bowl and cube agree within
$2.9$\,mm and $1.4^\circ$, respectively, whereas those for the microwave
differ by $9.0$\,mm and $9.3^\circ$.}
\label{tab:repose}

\begin{tabularx}{\linewidth}{@{}l >{\centering\arraybackslash}p{0.25\linewidth} >{\centering\arraybackslash}X >{\centering\arraybackslash}X >{\raggedright\arraybackslash}X@{}}
\toprule
\textbf{Body} & \textbf{Proposal} $(x,y;\psi)$ & \textbf{Mean IoU} & \textbf{Weakest-View IoU} & \textbf{Outcome} \\
\midrule
Bowl & $-15.6,\,-6.2$\,mm
     & $0.814\rightarrow0.861$
     & $0.691\rightarrow0.742$
     & admitted \\
Cube & $-14.4,\,+3.4$\,mm; $-4.05^\circ$
     & $0.706\rightarrow0.858$
     & $0.590\rightarrow0.737$
     & admitted \\
Microwave & \textemdash{}
          & \textemdash{}
          & \textemdash{}
          & refused (unstable proposal) \\
\bottomrule
\end{tabularx}
\end{table}

Table~\ref{tab:repose} and Figure~\ref{fig:repose-overlay} report one scan.
Each body's call takes $1.4$--$2.2$\,s (the depth proposal $0.2$--$0.5$\,s,
the twin admission $1.0$--$1.5$\,s), once every view has been segmented
for every body's prompt in one pass ($2.2$\,s for eight views and three
prompts) and the twin's GPU context has been set up; reopening the
finished run in an owned Blender adds about $11$\,s once per session.
The MuJoCo twin rebuilt from the new export displaces the bowl and
the cube exactly as the receipts predict, to the reported $0.1$\,mm.

Two observations shaped the procedure. First, the cube was initially
refused for a reason other than its pose: its reconstructed body
included a thin square modeled from the cube's shadow, which the
SAM\,3 mask excludes; step~$2$ comes from this case. Second, depth and
silhouettes place the cube about $5$\,mm apart along the wrist camera's
viewing direction, and per-view silhouette optima scatter by about
$11$\,mm with the viewing direction, with or without undistortion.
This points to a few millimeters of camera-pose error (hand--eye,
kinematics, or base) or depth bias rather than to the lens,
and the cube's pose is uncertain by about $5$\,mm.

\paragraph{Autonomous recovery.}
A latched protective stop does not hold the arm's pose: the arm slowly
sags onto its support. The firmware reset de-energizes every joint
at once, so recovery is automated only under physical preconditions:
every driver reports a normal supply voltage and no fault other than
a resettable motor over-temperature (with the motor back below
$50$\,$^\circ$C), the joints have been still for $3$\,s (span at most
$0.05^\circ$), and the flange is at most $8$\,cm above the arm's base
plane, that is, resting on a support. A suspended arm is handed
to a person.

The sequence waits on observed state rather than fixed delays:
reset; wait until the controller has reinitialized with all drivers
disabled and valid joint feedback; resend the enable command every
$20$\,ms until all six drivers report enabled; switch to joint-position
mode; and watch for $3$\,s for normal status, no error, all drivers
enabled, and no motion. Any failure re-latches the stop.
A sagged pose can lie outside the device's joint limits, where
the standard joint-move tool refuses to start. A bounded retract
then widens that joint's limit temporarily and streams quintic
set-points at $100$\,Hz that lift with the shoulder and elbow before
bringing the wrist joint back inside its window, never lowering
the fingertip. On a tracking error the arm holds its measured pose,
and a second failure latches the stop. The widened limit is
restored only once the joint is back inside it; otherwise
the stop is latched first.

\subsection{Compliant execution on the physical arms}
\label{sec:robotics-compliant}

The executor that streams a planned path to the hardware is split into a
backend-independent loop and two backends. At each $100$\,Hz frame, a control law
proposes joint targets and a set of guards (tracking error, motor current,
temperature, drive faults, control mode, and a fingertip floor at the lowest
point of the planned path minus $10$\,mm) decides whether to hold the measured
pose. The backends are the MuJoCo twin and the Piper arms over SocketCAN, so a
motion program is first executed in the twin with the same loop and guards
that later run it on the robot. Against the earlier hardware-only streaming
script, frozen as a test fixture, the new loop emits identical CAN frames and
result fields in $16$ cases (arrival, fingertip drop, stall, over-current,
over-temperature, drive fault, loss of CAN mode, and multi-segment paths); a
free-space home--ready--home motion on the real right arm tracks within
$0.328^\circ$ (previously $0.320^\circ$ and $0.332^\circ$ on the same path).

\paragraph{Force estimation from motor current.}
The arms have no force sensor. For each joint we fit a current gain and a
friction torque to free-space motion through ten twin-cleared excitation poses
(two passes of about $280$\,s each); the external torque is the measured current
minus the gravity and friction prediction, mapped to a fingertip force through
the arm Jacobian. Table~\ref{tab:current-model} lists the fit. The axes of J1
and J6 are not loaded by gravity on this path, so their gains are borrowed from
J2 and J5 and are marked as such. In a vertical descent onto the table at
$1$\,cm/s with a $5$\,N threshold, the estimate reads $0.61\pm0.47$\,N (maximum
$1.64$\,N) during free descent, triggers at $5.3$--$5.6$\,N on contact after a
further $1.9$\,mm of travel.

\begin{table}[htbp]
\centering
\caption{\textbf{Current-to-torque model of the right arm}, fitted on
free-space motion. Noise is the single-frame residual; J1 and J6 borrow the
gain of the motor of the same type.}
\label{tab:current-model}

\begin{tabular*}{\linewidth}{@{\extracolsep{\fill}}clNNNN@{}}
\toprule
\textbf{Joint} & \textbf{Status}
& \shortstack[c]{\textbf{Gain}\\(counts/(N\,m))}
& \shortstack[c]{\textbf{Friction}\\(N\,m)}
& \shortstack[c]{\textbf{Noise}\\(N\,m)}
& $\boldsymbol{R^2}$ \\
\midrule
J1 & borrowed & $223$  & $0.29$ & $0.31$ & \textemdash{} \\
J2 & fitted   & $223$  & $0.50$ & $0.37$ & $0.92$ \\
J3 & fitted   & $238$  & $0.50$ & $0.32$ & $0.80$ \\
J4 & fitted   & $1505$ & $0.06$ & $0.05$ & $0.92$ \\
J5 & fitted   & $1530$ & $0.05$ & $0.05$ & $0.91$ \\
J6 & borrowed & $1530$ & $0.04$ & $0.05$ & \textemdash{} \\
\bottomrule
\end{tabular*}
\end{table}

\paragraph{Contact control laws.}
Contact control laws run inside the loop. For door-opening tasks, 
\code{admittance\_pull} pulls a grasped handle from an initial direction and
distance and turns the pull direction toward the measured force, so it needs
no hinge model; once the hand carries a load, the grip torque is raised one
step for every $2$\,mm of slip. \code{compliant\_grip} closes the gripper while
yielding along the closing axis, centering a handle in one motion ($3.7$\,s,
against $13$\,s for the previous three-pass centering).

\clearpage
\section{NeoSDK and NeoMCP implementation}
\label{sec:sdk-mcp-implementation}

\subsection{NeoSDK: source programs and numerical execution}
\label{sec:sdk}

\SDK{} decides what a program can express, how its constants are searched,
and how a chosen result is rebuilt in Blender~\citep{blender}; every section above relies on
it. \model{} implements its modeling workflow through this reusable layer
for geometry construction, source editing, differentiable execution, and
Blender materialization. Studio connects these operations to observations and
decides which reconstructed state is retained.

\paragraph{Executable programs.}
The executable state is $P=(S,\theta,T)$: the authoritative Python source $S$,
its named constants $\theta$, and the instance transforms $T$. Parts,
materials, profiles, and attachment coordinates are expressions of the source,
so one shared constant can preserve a construction relationship across several
parts. The SDK derives a disposable source index from $S$: part spans, constant
locations, readers, and geometry call sites are bound to the source-byte hash.
Edits and numerical write-back address UTF-$8$ byte spans and invalidate the old
index. The index is a lookup over source, not a separately editable geometry
representation; the program is not regenerated from an intermediate graph.

\paragraph{Differentiable execution and parameterization.}
The twin rewrites the searched constants into batched tensor inputs, replays
the supported modeling vocabulary in Torch, and rasterizes whole candidate
populations with nvdiffrast. One refinement implementation serves both lanes;
only the evidence differs, as described in Section~\ref{sec:object-proposals}.
Twin/Blender parity at the baseline is a precondition of search. A construction
the twin cannot replay raises \code{TwinUnsupported}; a unit that owns no
searched constant may instead be carried frozen at its Blender export, where it
still occludes. A failed parity check raises \code{TwinMismatch}, and values
that describe no valid shape are scored infeasible. None of these falls back to
a per-candidate Blender replay. Blender builds the baseline, materializes and
admits the selected candidates, and renders the final result.

\paragraph{Two supported geometry vocabularies.}
The scene prelude and the public builder vocabulary have separate twin
bindings. The builder vocabulary covers spheres, rounded boxes, cylinders,
curve tubes, lofted rectangular sections and shells, thick surface grids, and
revolved or extruded profiles; imports of these \code{neosdk.blender} helpers
resolve to their tensor kernels. This extends numerical search to programs
written with those helpers, but does not make arbitrary Blender Python
differentiable. Unsupported operations and failed parity remain explicit
outcomes. Repeated prototype instances retain their shared source and separate
transforms.

\paragraph{Mesh distortion during fitting.}
\label{sec:sdk-mesh-admission}
The twin retains each object's baseline edge lengths and bounding-box extents.
For a candidate it measures the largest edge-length growth, divided by that
object's largest axis-wise extent growth and floored at one, and takes the
maximum over objects. A uniform or one-axis resize therefore scores one, while
local stretching grows much faster. The \code{mesh\_torn} admission clause rejects
values above $2.0$ and the same clause is priced as a hinge during search, so
the optimizer is not rewarded for proposing what admission refuses. This check
does not certify watertightness, topology, attachment, or physical validity.

\paragraph{Materialization and part-aware inspection.}
The modeling catalog provides object-agnostic construction helpers together
with a bounded subset of native Blender operations. Compilation checks source
validity, available operations, finite geometry, and part ownership before
materializing the program, and each rendered object retains its declared part,
so images and measurements can be related back to source. A selected
source-changing candidate is freshly executed in Blender; a pose candidate
transforms the extracted assembly without rebuilding its geometry. Search
iterates stay in the twin, and these native checkpoints produce the inspectable
candidate compared in Section~\ref{sec:object-admission}. Export preserves the
retained source, its source index, the native build, the pose, and the frame, so
that reconstruction can continue from the same state.

\subsection{NeoMCP: agent-facing operations}
\label{sec:mcp}

An agent's reach over a reconstruction is exactly the set of operations it can
invoke. \MCP{} is that set, gathered into a single registry and held to one
record per operation, so that the surface can be described, served to different
clients, and measured. It sits
between the geometry substrate of Section~\ref{sec:sdk} and the loop that
consumes it: \SDK{} decides what geometry can express, \MCP{} decides what the
model can ask for, and the harness decides what is kept.

\paragraph{One registry, one record per verb.}
\label{sec:mcp-registry}

Every model-facing operation is registered once with a name, a one-line
description, a strict input schema, an implementation, and a per-call receipt.
Every property is declared and required, nothing additional is accepted, and
the schema is validated at registration, so a malformed surface cannot reach a
model. \emph{No failure escapes as an exception}: an unknown verb, a schema
violation, a refusal by the verb, and an unexpected error all return the same
typed record, so a tool bug becomes feedback. \emph{The model's view is bounded
while measurement is not}: images sent to the model are bounded re-encoded
copies, and an oversized record reaches the model as a summary naming the
complete record filed with the run, whereas measurement reads the original
raster. The receipt records the call's ordinal, verb, argument digest, elapsed
time, record size, number of images sent, and error status, giving an auditable
account of what the agent asked for and what it was told.

\paragraph{One surface, two consumers.}
\label{sec:mcp-consumers}

The registry is served, unmodified, through two adapters: projected into
function-calling specifications for our own loop, and mounted as a
tool-protocol server over a standard transport for an external coding agent,
which holds it among its own file and shell tools. Both clients receive the same
schemas, typed records, and image bytes, so substituting an external agent
isolates the loop from the interface.

\paragraph{The live scene the surface owns.}
\label{sec:mcp-live}

Each run owns one filesystem-isolated modeler process; the modeling agent sees only
the sealed observation packet and its own working directory. The process leads
its own process group, and its recorded identity and start time are re-verified
before any lifecycle action. A reply that does not echo the generation minted at
startup and the expected result schema is refused. However the episode ends, the
scene is saved and exported from what it actually holds, not replayed.

\paragraph{What stays in the harness.}
\label{sec:mcp-harness}

The surface exposes operations and diagnostic feedback; the harness retains
candidate-admission and episode-control authority. An audit classification is
not a quality verdict, and \code{finish} requests completion without claiming
that the result is good. Remaining model calls and wall time ride along on every
tool record, but the host counts and enforces them, holding back a finalization
reserve for saving and export. In the optional two-phase episode, the first
\code{finish} triggers a bounded assembly and part audit and returns a work list
for a second modeling phase within the same budget.

Agent-written source passes a staged admission --- syntax, source validation,
compilation preflight, execution, and receipt --- and each stage has its own
typed failure. The harness never edits a semantic value (a dimension, material,
joint family, axis, or joint range) to make a program pass; numerical changes
arrive only as admitted refinement candidates, judged against their retained
baseline by the SDK-owned rules of Section~\ref{sec:object-admission} in both
lanes (the object lane omits only the clauses that protect a scene's
surroundings). A weakest-view check does not imply that every view improves, and
a refused candidate leaves the retained program unchanged. Explicit selection and
export remain separate from this numerical decision. An admitted number is
written where the program keeps it: a named constant or an inline literal in
place in the source, a fitted measurement in the object's fit file
(Section~\ref{sec:scene-tests}). The between-round fit writes a number only
where the source text is unchanged since the fitted build. The harness's own
structural edits are few, and each is receipted:
\begin{itemize}
\setlength{\itemsep}{1pt}\setlength{\parskip}{0pt}
\item retiring a delegated body's \code{GROUPS} and \code{JOINTS} rows, and hosting
the delivered asset under a name prefix;
\item leaving out a delivered joint whose family, side, or axis is unresolved;
\item re-bracketing a declared search range around an admitted value;
\item hoisting an inline literal into a named constant of the same value, kept
only if the rebuild is identical (Section~\ref{sec:object-proposals}).
\end{itemize}

\paragraph{The verbs and mounted catalogs.}
\label{sec:mcp-verbs}
Table~\ref{tab:verbs} (Section~\ref{sec:scene}) summarizes the
live-scene tools for observation, measurement, geometry construction,
calibrated comparison, scene refinement, and diagnosis.
The object session additionally provides a strict-schema tool catalog
for workspace inspection, evidence viewing, source reading,
geometry creation, exact-anchor editing, native preview,
\path{select_candidate}, and task completion.
Optional refinement adds \path{observe_parameters} and
\path{refine}, while articulated tasks additionally provide
\path{probe_articulation}, \path{review_opened_state},
\path{select_motion_artifact}, and
\path{record_articulation_unknown}
(Section~\ref{sec:object-articulation}).
For reproducibility, each run records its mounted tools and
configuration rather than relying on a fixed global tool count.

\paragraph{Optional scene feedback.}
The critics behind \path{review_views} run in parallel with a fixed
per-run review cap, while \path{scan_discrepancies} examines every
frame of the reconstructed scene. Both provide diagnostic evidence
for the modeling agent to verify, but neither authorizes edits or certifies
overall scene quality. Predicted depth is subject to the
qualifications in Section~\ref{sec:predicted-depth}.

\paragraph{From part discrepancies to repair steps.}
\label{sec:mcp-audit}
\code{audit} turns photographic disagreement into a bounded work list. It
segments each built body's parts using the body's prompt and terms from the
agent's part names, in the frames where the body shows best; a part mask must
lie mostly on the body without being the whole body. Parts are classified as
aligned (IoU $\geq 0.80$ in every measured view), misshapen, displaced, of
doubtful identity, or unmeasurable, ranked by mask area times
$(1-\mathrm{IoU})$, and at most eight are returned. Each carries one action: fit
declared constants, name inline literals, declare a missing control, split or
unfreeze a source block the twin cannot trace, rewrite one construction, or
locate the part's source lines.

To choose constants, the batched twin trace measures each constant's effect
over its range: mean vertex displacement estimates translation, and the largest
residual displacement after removing it estimates deformation. A constant that
only translates what it reaches is left to the preceding pose step. Shape steps
take at most six constants, ordered by the share of their effect that is
deformation, each within the interval that moves the geometry by the measured
error. The list proposes what to try; only another audit measures what an
action changed. The audit also carries the \code{joints} sweeps of declared
joints, with exact-anchor range edits where applicable; the modeling agent decides
whether to revise the source, and the tool never commits a joint.


\section{NeoWorld-10K: taxonomy and construction details}
\label{app:neoworld}

NeoWorld-10K is a publicly released, open-source dataset of $10$K simulation-ready articulated 3D assets constructed with \model{} for scene construction and embodied interaction. Each entry is one object instance, with editable geometry, materials, and kinematics. Figure~\ref{fig:neoworld-construction} summarizes the construction pipeline and three stages of human screening. Figure~\ref{fig:neoworld-dataset} presents the taxonomy, representative reconstructed assets, and category word cloud.

\par\addvspace{\intextsep}
\noindent\begin{minipage}{\linewidth}
    \centering
    \includegraphics[width=\linewidth]{
        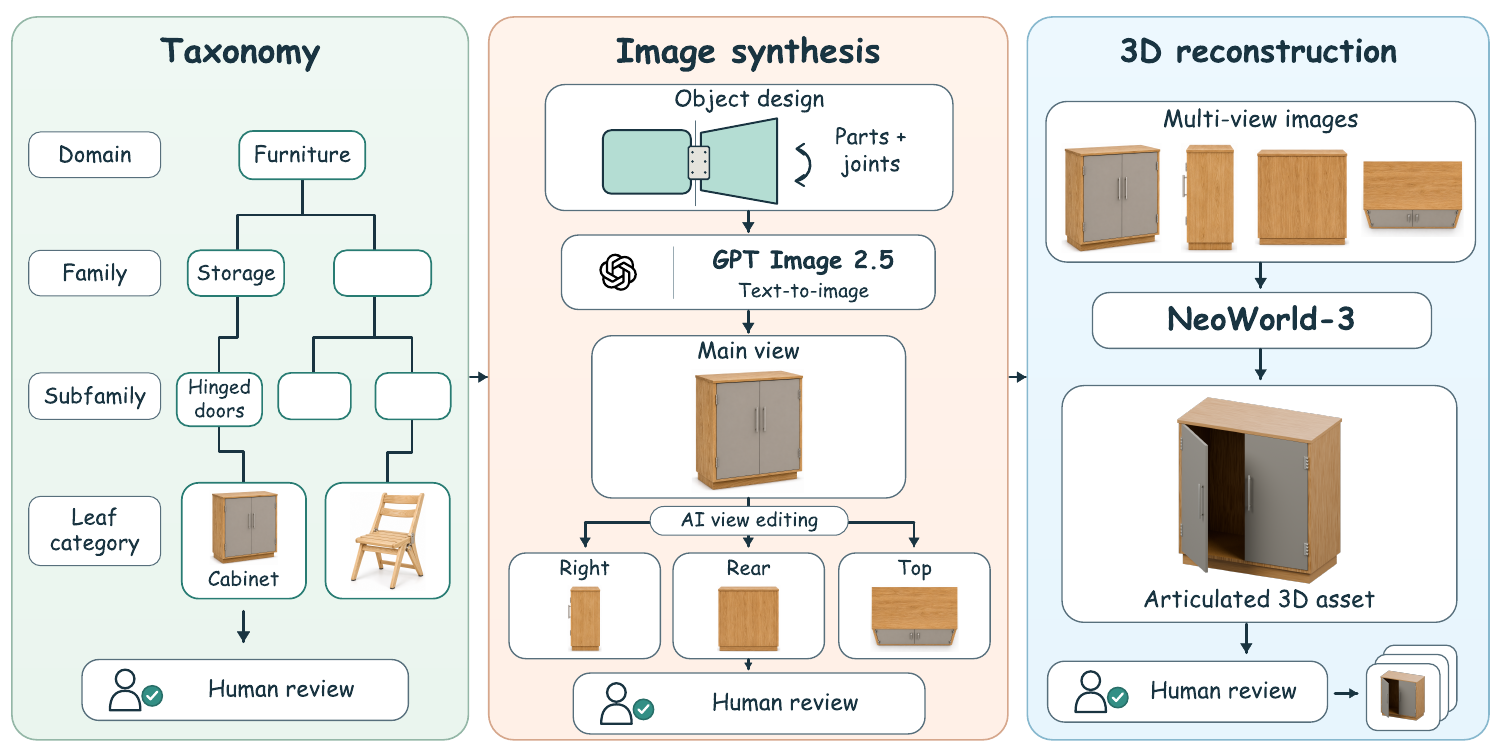
    }
    \captionsetup{hypcap=false}
    \captionof{figure}{
        \textbf{Construction pipeline of NeoWorld-10K.}
        A hierarchical taxonomy guides articulated-object design.
        GPT Image 2.5 produces a main reference image and three additional views
        conditioned on it. \model{} reconstructs articulated 3D assets from these
        references and validates their geometry and articulation before the final
        human screening.
    }
    \label{fig:neoworld-construction}
\end{minipage}
\par\addvspace{\intextsep}

\subsection{Dataset scope}
\label{app:neoworld-scope}

NeoWorld-10K targets reusable articulated objects for object-centric simulation and embodied interaction. The emphasis is on products whose operation involves constrained relative motion between connected rigid components, including doors, lids, drawers, adjustable supports, controls, wheels, and rotating assemblies. The collection prioritizes objects at ordinary functional scales: small products such as cameras and personal accessories, medium-sized household appliances and furniture, and a smaller set of operable human-scale products. Eligibility does not require that an entire object be handheld. Building-scale installations, infrastructure, complete vehicles at truck scale, shipping containers, walk-in facilities, and production lines are outside the intended scope. Figure and puppet products are also excluded from the catalog.

The semantic scope extends beyond hinge motion. Bounded rotation, continuous self-rotation, guided translation, telescoping, screw motion, and mechanically coupled or multi-coordinate motion are eligible when supported by a credible rigid assembly. Deformation of fabric, rope, or elastic components alone does not establish articulation. Likewise, translating an entire object or moving an unconstrained detached component does not constitute an articulated mechanism. The relevant unit is a mechanically meaningful degree of freedom, rather than the number of visible seams, bearings, rails, or individual hinge leaves.

\subsection{Hierarchical semantic organization}
\label{app:neoworld-taxonomy}

\paragraph{Hierarchy.}
The catalog follows the hierarchy
\emph{Domain $\rightarrow$ Family $\rightarrow$ Subfamily $\rightarrow$ Leaf Category}.
A domain identifies a broad application context. A family groups products by functional purpose within that context. A subfamily separates a more specific functional or product grouping. A leaf category defines an eligible object type and its semantic boundary. Category membership is distinct from instance construction: a leaf definition describes what an object is and what neighboring concepts it excludes, while the geometry, component layout, and joint parameters are specified separately for each instance. For example, a single leaf category may contain multiple instances with different plausible support structures or component arrangements.

Every leaf is associated with a persistent identifier, an English name, a domain, a family, a subfamily, and a concise definition. Parent assignments are retained explicitly, so an object can be retrieved at any of the four levels without recovering its ancestry from its name. Family and subfamily counts refer to their positions in the hierarchy, rather than to globally unique strings. The dataset taxonomy comprises $19$ domains, $219$ families, $536$ subfamilies, and $912$ leaf categories. Table~\ref{tab:neoworld-taxonomy} reports the number of families, subfamilies, and leaf categories within each domain.

\par\addvspace{\intextsep}
\noindent\begin{minipage}{\linewidth}
    \centering
    \begin{minipage}[c]{0.532\linewidth}
        \centering
        \includegraphics[width=\linewidth]{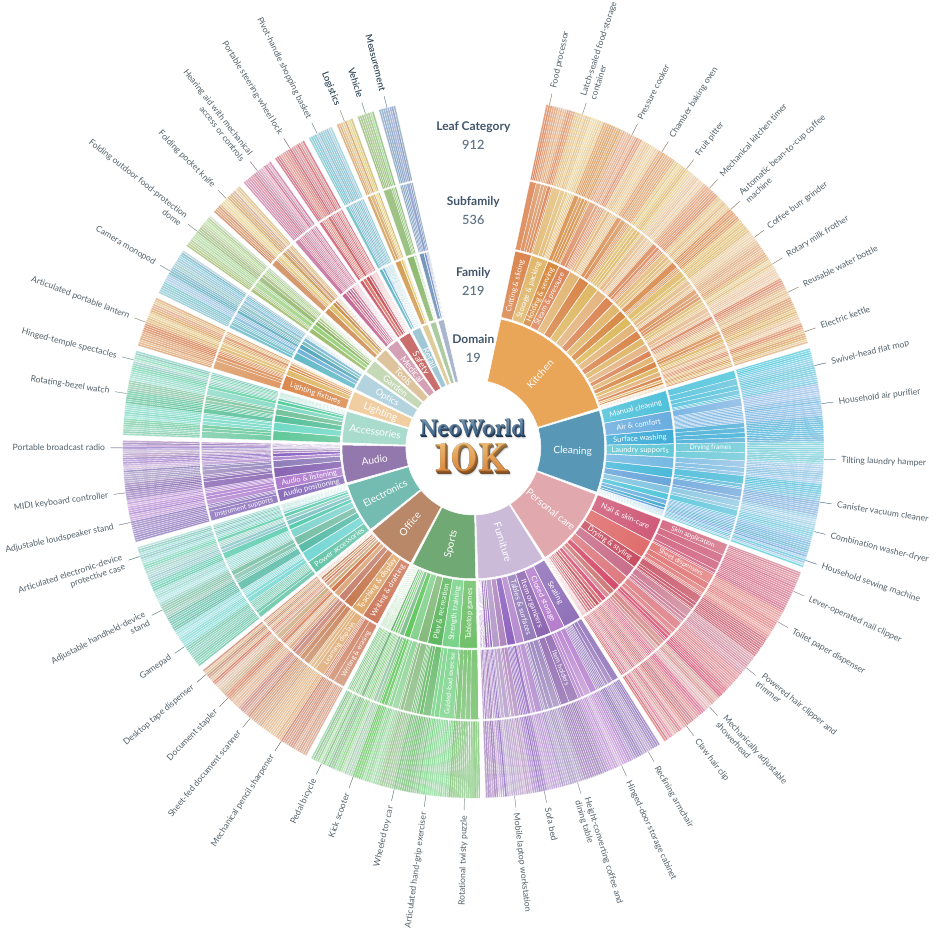}
    \end{minipage}\hfill
    \begin{minipage}[c]{0.454\linewidth}
        \centering
        \includegraphics[width=\linewidth]{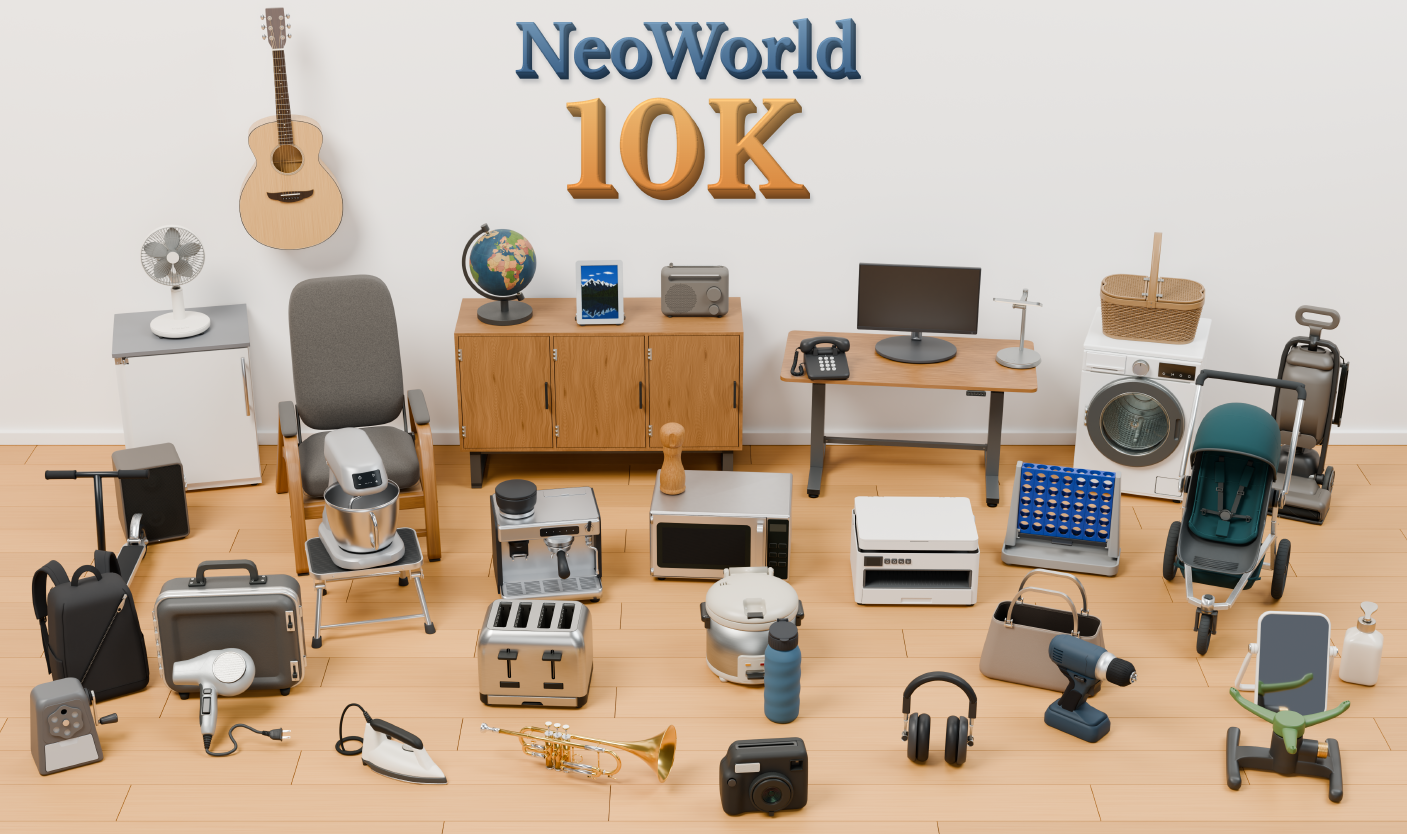}\par
        \vspace{4pt}
        \includegraphics[width=\linewidth]{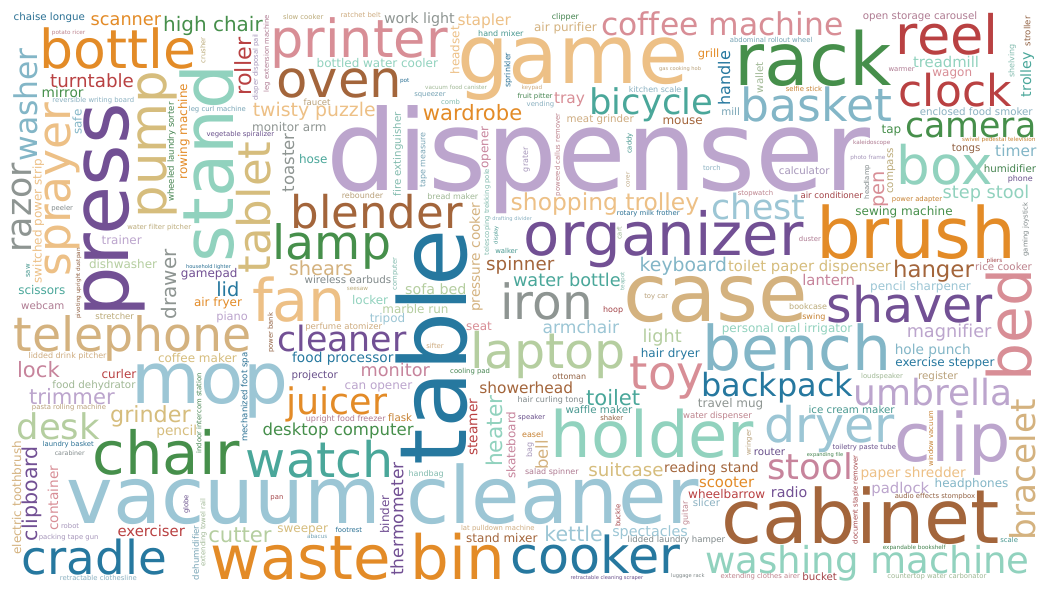}
    \end{minipage}
    \captionsetup{hypcap=false}
    \captionof{figure}{\textbf{NeoWorld-10K constructed with \model{}.}
    Left: the Domain--Family--Subfamily--Leaf Category hierarchy.
    Upper right: representative articulated assets reconstructed by \model{}.
    Lower right: word cloud illustrating semantic diversity.}
    \label{fig:neoworld-dataset}
\end{minipage}
\par\addvspace{\intextsep}

\paragraph{Example.}
For a hinged-door storage cabinet, the four-level assignment is
\emph{Furniture and general storage} (Domain)
$\rightarrow$ \emph{Enclosed storage furniture} (Family)
$\rightarrow$ \emph{Swinging and folding access structures} (Subfamily)
$\rightarrow$ \emph{Hinged-door storage cabinet} (Leaf Category).
This path identifies the object type; the number of doors, cabinet dimensions,
material assignments, and joint parameters are specified separately for each
reconstructed instance.

\paragraph{Household emphasis.}
The five domains devoted to kitchen equipment, household cleaning, personal care, furniture, and personal accessories and carrying equipment account for $506$ of the $912$ leaf categories ($55.5\%$). Household-related objects also occur in domains such as lighting and consumer electronics; the five-domain grouping therefore provides an explicit, reproducible accounting convention rather than an exhaustive definition of domestic use. The remaining domains broaden the collection to office and educational products, sports and recreation, audio, imaging, gardening, and selected specialist equipment. Domain frequencies describe the curated semantic catalog and are not estimates of real-world product prevalence.

\paragraph{Category boundaries and diversity.}
Catalog construction combines deterministic duplicate filtering, retrieval of related definitions, and semantic comparison of potentially overlapping categories. Complete products and their components are distinguished to avoid counting the same concept at incompatible levels. Category retention follows functional scope and mechanical plausibility. Instance diversity is subsequently established through changes in geometry, component layout, support structure, or functional construction. Color changes, branding, texture replacement, and uniform resizing alone do not establish structural diversity.

\begin{table}[htbp]
\centering
\caption{\textbf{Taxonomy of NeoWorld-10K.} Numbers of families, subfamilies, and leaf categories within each of the $19$ domains.}
\label{tab:neoworld-taxonomy}
\begin{tabular}{@{}>{\raggedright\arraybackslash}p{0.61\linewidth}NNN@{}}
\toprule
\textbf{Domain} & \textbf{Families} & \textbf{Subfamilies} & \textbf{Leaves} \\
\midrule
Kitchen, food-processing and catering equipment & $27$ & $105$ & $173$ \\
Household cleaning, laundry and waste handling & $17$ & $49$ & $104$ \\
Bathroom, hygiene and personal care & $20$ & $53$ & $103$ \\
Furniture and general storage & $15$ & $46$ & $86$ \\
Sports, exercise and recreation & $20$ & $41$ & $82$ \\
Computing, communications and general electronics & $16$ & $43$ & $63$ \\
Office, education and document equipment & $14$ & $34$ & $63$ \\
Audio and musical equipment & $12$ & $30$ & $48$ \\
Wearable accessories and personal carrying equipment & $12$ & $24$ & $40$ \\
Lighting, stage and exhibition equipment & $6$ & $15$ & $24$ \\
Optics, photography and imaging & $8$ & $17$ & $23$ \\
Garden, grounds and outdoor living equipment & $8$ & $12$ & $18$ \\
Industrial tools and manufacturing equipment & $7$ & $11$ & $17$ \\
Medical, dental and assistive equipment & $9$ & $14$ & $16$ \\
Safety, security and emergency equipment & $9$ & $14$ & $16$ \\
Retail, vending and commercial service equipment & $9$ & $11$ & $12$ \\
Land-vehicle components & $3$ & $4$ & $8$ \\
Measurement, scientific instrumentation and laboratories & $4$ & $7$ & $8$ \\
Transport packaging and material handling & $3$ & $6$ & $8$ \\
\midrule
\textbf{Total} & $\mathbf{219}$ & $\mathbf{536}$ & $\mathbf{912}$ \\
\bottomrule
\end{tabular}
\end{table}

\subsection{Dataset construction}
\label{app:neoworld-reference-construction}

\paragraph{Instance specification.}
For each object instance, ChatGPT receives the definition and semantic boundaries of its leaf category, along with previous designs from that category. The resulting structured specification describes the complete object, its geometry, rigid components, functional connections, intended movable parts, and reference configuration. Processing designs within a category sequentially allows subsequent designs to account for earlier constructions.

\paragraph{Reference configuration.}
The default configuration is an ordinary assembled resting state. Doors and lids are preferably closed, drawers and telescoping structures are retracted, and adjustable components occupy plausible neutral positions. A nonbaseline pose is specified only when required by the object's function and is recorded consistently in the structured design and image description. The object remains opaque and assembled, with natural occlusion. Functional connections are expressed through actual connected rigid bodies and plausible pivots, bearings, guides, threads, or linkages.

\paragraph{Reference-image synthesis.}
ChatGPT produces a complete English prompt for a single photorealistic object on an opaque white background. GPT Image 2.5 generates the main reference image from this prompt through text-to-image generation. The reference view shows the functional front or a front-oblique view, with the complete object inside the frame and space for subsequent motion. Objects without a conventional front are oriented to expose their normal operating or display face. The prompt specifies the intended structure, material appearance, and reference configuration, while excluding people, independent props, watermarks, and extraneous text.

\paragraph{Additional views.}
The same model synthesizes three additional images depicting the right, rear, and top views through image-to-image generation. Each generation request is conditioned directly on the original reference image and a textual description of the target viewpoint, rather than recursively on the preceding synthesized view. The prompts ask the model to preserve object identity, geometry, hardware, assembly, and joint configuration and to infer unseen surfaces conservatively. Together, the four views provide the visual inputs for reconstruction.

\paragraph{3D asset reconstruction.}
From the screened four-view sets, \model{} reconstructs part-structured geometry programs and attaches explicit joint descriptions. During reconstruction, \model{} checks candidate geometry against the visual references (Section~\ref{sec:object}) and probes the declared joints at sampled configurations to identify unintended interpenetration (Section~\ref{sec:object-articulation}). The reconstructed assets then undergo the final human screening described below.

\subsection{Three-stage human screening}
\label{app:neoworld-evaluation}

Human screening is conducted at three stages of dataset construction: after taxonomy construction, after image synthesis, and after 3D reconstruction, as summarized in Figure~\ref{fig:neoworld-construction}. Each stage examines a different representation and uses the corresponding criteria below.

\paragraph{Stage $1$: taxonomy review.}
Reviewers examine object-category diversity, semantic coverage, and the reasonableness of the proposed categories. The review considers coverage across application domains and product families, the coherence of category definitions and hierarchical assignments, and whether the categories describe plausible objects with meaningful rigid articulation. This stage checks the breadth and organization of the catalog before concrete object instances are generated.

\paragraph{Stage $2$: image screening.}
Reviewers reject images with physically implausible assemblies, such as floating or disconnected components or implausible interpenetration. The reference imagery must also show at least one clearly visible principal joint between rigid parts. Reviewers check each four-view set for 3D consistency: object proportions, part and joint placement, and the reference configuration must agree across views, allowing for changes in viewpoint and occlusion. Among acceptable sets, reviewers favor representative objects with realistic appearance, natural proportions, and clear imagery. The filenames of selected images identify the inputs used for reconstruction.

\paragraph{Stage $3$: reconstructed asset screening.}
The final review examines each reconstructed asset from multiple viewpoints for visual quality and structural plausibility. Reviewers exercise every movable joint through its full available range in the interface and check for implausible motion, interpenetration, parts that move but should remain fixed, and parts that should move but do not. For continuous or coupled motion, reviewers inspect the rotation or coordinated part motion exposed by the controls. They then run drop, inclined-plane, and ground-shaking tests to check simulated behavior and assembly integrity, resetting the asset between trials as needed and looking for abnormal responses or unexpected separation.

\subsection{Dataset organization and annotations}
\label{app:neoworld-organization}

Each asset is stored in an object-specific directory containing its original geometry and material source code (\texttt{static.py}), a native Blender file preserving articulated parts, procedural materials, and recorded animation (\texttt{articulation.blend}), and a kinematic URDF model (\texttt{object.urdf}). Complete-object representations are provided in GLB (\texttt{object.glb}) and USDZ (\texttt{object.usdz}) formats, with embedded materials and textures and recorded animation where available. Individual rigid links are additionally provided as self-contained GLB files under \texttt{parts/}, referenced by the URDF with the transforms required for assembly. Portable materials use PBR parameters and finite-resolution texture atlases, while the Blender file retains the native procedural shaders.

Structured annotations in \texttt{articulation.json} record object identity and category, coordinate conventions and units, part organization, and joint types, parent--child relationships, origins, axes, motion limits, initial states, and coupling where applicable. They also retain material mappings, motion records, evidence provenance, and unresolved mechanisms, preserving the distinction between observations and design priors. Bounding-box annotations include initial-state 3D axis-aligned boxes for the object and geometric links, together with per-view 2D projected boxes that include occluded geometry.

Each package includes four rendered views---front, side, top, and isometric---under \texttt{renders/}, generated by reloading the delivered GLB at its initial state. Corresponding camera poses, projection parameters, and image-coordinate conventions are stored in the JSON file.

\clearpage
\section{Supplementary experiments and evaluation protocols}
\label{sec:experiment-appendix}

This appendix collects evaluation protocols, benchmark analyses, and
comparisons with baseline methods for reconstruction, scene compute, mesh
closure, and physical robot execution.
Algorithm and implementation details appear in Appendices~\ref{sec:method-details},
\ref{sec:robotic-deployment}, and~\ref{sec:sdk-mcp-implementation}.

\subsection{Datasets}
\label{sec:appendix-evaluation-datasets}
For object reconstruction, we use $212$ instances from
3DCodeBench~\citep{codebench3d}, a benchmark designed to evaluate coding-based
procedural 3D reconstruction. For articulated-object reconstruction, we use
$77$ selected PartNet-Mobility objects~\citep{sapien}, with preprocessed assets and
annotations from SINGAPO~\citep{singapo}. For scene reconstruction, we use
$10$ target scenes from RoboCasa365~\citep{robocasa365}, a robotics simulation
benchmark with articulated environments. Scene $01$ is used for development;
all scene comparisons and averages use the same nine held-out scenes $02$--$10$.
Table~\ref{tab:appendix-evaluation-datasets} summarizes these settings.

Within each benchmark, all methods receive the same input images for
reconstruction: four images per object on 3DCodeBench, three images per object
on PartNet-Mobility, and $16$ images per scene on RoboCasa365.

\begin{table}[H]
\centering
\caption{\textbf{Evaluation datasets.} Each evaluation setting targets a distinct
level of the reconstruction framework.}
\label{tab:appendix-evaluation-datasets}
\begin{tabularx}{\linewidth}{@{}>{\raggedright\arraybackslash}p{0.18\linewidth} >{\raggedright\arraybackslash}p{0.30\linewidth} N N X@{}}
\toprule
\textbf{Setting} & \textbf{Dataset} & \textbf{Instances} & \textbf{Input views} & \textbf{Focus} \\
\midrule
Object & 3DCodeBench & $212$ & $4$ & Geometric and visual asset quality \\
Articulated object & PartNet-Mobility / SINGAPO & $77$ & $3$ & Part geometry and joint motion \\
Scene & RoboCasa365 & $9$ held-out & $16$ & Multi-object layout and functional structure \\
\bottomrule
\end{tabularx}
\end{table}

\subsection{Baselines and comparison settings}
\label{sec:appendix-evaluation-baselines}
We compare \model{} with direct-generation and coding-agent reconstruction
baselines. At the object level, these include Direct GPT, Codex,
Codex + BlenderMCP, VIGA, and Articraft. For articulated objects, we compare with
Direct GPT, Codex, Codex + BlenderMCP, and Articraft; VIGA is excluded from
this setting because it does not support articulated asset generation.
At the scene level, we compare with Direct GPT, Codex,
Codex + BlenderMCP, and VIGA.

\paragraph{Direct GPT.}
Direct GPT submits the input images and task prompt directly to the GPT model~\citep{openaigpt}
through the Responses API in a single call. The returned reconstruction program
is evaluated as the direct-generation output.

\paragraph{Codex.}
Codex denotes the coding-agent harness~\citep{codexcli} running the underlying
GPT model. We pass the task prompt through Codex CLI; this baseline uses the
harness's native capabilities without an MCP server.

\paragraph{Codex + BlenderMCP.}
Codex + BlenderMCP uses the same Codex CLI workflow with the Blender MCP
server~\citep{blendermcp} connected, exposing Blender operations to the agent
through MCP. The Codex harness and underlying GPT model are retained.

VIGA~\citep{viga} reconstructs visual content through its iterative
code--render--inspect loop, while Articraft~\citep{articraft} uses a coding
agent with an asset-authoring SDK.

\paragraph{Agent interaction protocol.}
Codex and Codex + BlenderMCP each receive exactly one user turn containing
the task prompt and input images. The agent completes that run autonomously,
potentially making multiple internal model and tool calls, and its resulting
artifact is evaluated directly. No additional user turns or human-in-the-loop
feedback are provided.

\paragraph{Reconstruction model.}
All methods use GPT-5.6 Terra high for object reconstruction on 3DCodeBench
and PartNet-Mobility, and GPT-6 Sol high for scene reconstruction.

\paragraph{Reconstruction prompts.}
Direct GPT, Codex, and Codex + BlenderMCP receive identical task prompts
within each benchmark: the official 3DCodeBench prompt as the base instruction
for 3DCodeBench, and shared custom task prompts for PartNet-Mobility and
RoboCasa365. Articraft and VIGA use the reconstruction prompts supplied by their
respective implementations. The custom task prompts will be provided in the
supplementary materials.

\paragraph{Appearance requirements.}
For 3DCodeBench, we extend the official white-model setting by requiring all
methods to generate colored 3D assets with material assignments. This evaluates
appearance and material modeling in addition to geometric reconstruction;
the shared appearance requirement supplements each method's reconstruction
prompt.

\paragraph{Scene camera and depth configuration.}
The RoboCasa inputs ($16$ RGB views, cameras
estimated with VGGT-$\Omega$, no sensor depth) use this fit by default: in every
reported scene all $16$ views are fitted with the gauge pinned to the median
predicted scale, at per-view scales between $0.90$ and $1.10$ ($2.2$--$3.2$\,m per pose-model unit).

\subsection{Evaluation metrics}
\label{sec:appendix-evaluation-metrics}

\subsubsection{Object reconstruction}
\paragraph{Execution and aggregation.}
Exec measures successful GLB export over all $212$ tasks; code-based outputs
require both a GLB file and a successful latest export record. Feature
similarities and 3D geometry metrics use a common $211$-object set with complete
reference inputs. Non-Exec quality metrics are averaged only over successfully
executed assets with the inputs required by that metric. Failed executions
are accounted for by Exec and are excluded from these quality averages.
Object quality scores are truncated to four decimal places; closure
percentages are rounded to one decimal place.

\paragraph{Image similarity.}
SigLIP-2~\citep{siglip2} and DINOv3~\citep{dinov3} encode the four generated
views and four reference views into normalized features. For each encoder
independently, Hungarian matching finds the one-to-one view assignment with
the highest average cosine similarity. This matches image pairs without
performing 3D pose alignment.

\paragraph{Geometry and multimodal similarity.}
Chamfer distance~\citep{pointset} uses $8{,}192$ surface points per mesh, independently
centered and scaled to a unit sphere, and sums the two directional mean squared
nearest-neighbor distances. Uni3D~\citep{uni3d} computes cosine similarity
between embeddings of reconstructed and reference colored point clouds,
also sampled at $8{,}192$ points each. Image--3D similarity uses cosine similarity
between the reconstructed point-cloud embedding and the embedding of a single
designated reference image.

\paragraph{Framed IoU.}
For each of the four fixed view pairs, silhouette overlap is evaluated after
fitting an isotropic image-plane scale and translation. We retain the higher
of the original and fitted IoU for each view, then use the minimum across the
four views as the object score. All four views are required for a valid score.

\subsubsection{Articulated-object reconstruction}
\paragraph{Part matching and motion states.}
Following the SINGAPO protocol~\citep{singapo}, non-root nodes named \texttt{handle} are
excluded. Predicted and reference objects are centered, and the prediction's
overall extents are aligned to the reference along each coordinate axis.
Parts are matched by Hungarian assignment using rest-state bounding-box
center distances. Geometry is evaluated at five synchronized motion states;
revolute joints use the protocol's default signed $90^\circ$ excursion.
Joint-limit agreement is evaluated separately below.

\paragraph{Articulated geometry.}
AS-gIoU averages $1-\mathrm{gIoU}$~\citep{giou} between matched
part boxes across states, and AS-CD uses bidirectional squared Chamfer distance
with $2{,}048$ surface points per part. Both average states and non-root source
parts in each matching direction, then average the two directions. AOR,
introduced by CAGE~\citep{cage}, averages box IoU over predicted sibling-part
pairs and motion states using the prediction before reference-scale alignment.
An object with no eligible sibling pair contributes zero. Articulation errors
and AOR are rounded to four decimal places; joint-type accuracy is reported
as a percentage to two decimal places. AOR measures bounding-box overlap at these sampled states;
solid penetration and declared joint limits are evaluated separately in
Appendix~\ref{sec:articulation-penetration}.

\paragraph{Joint parameters and aggregation.}
Type compares joint types for matched non-root pairs, treating continuous
rotation as revolute. Axis error is the acute angle between matched movable
joint axes, reported in radians with opposite axis directions considered
equivalent. Pivot error is the shortest distance between matched rotational
axis lines in the aligned reference coordinates. Each joint-parameter metric is first averaged
within an object, then equally across objects with a valid score for that metric.
Closed is the object-level closure percentage over all $77$ identities, defined
in Appendix~\ref{sec:articulation-penetration}.

\subsubsection{Scene reconstruction}

\paragraph{Visual similarity.}
Scene DINOv3 uses $16$ reference--prediction RGB pairs matched by fixed view ID.
For each pair, we compute cosine similarity between L2-normalized global image
features, then average over views and equally across valid scenes. This uses
fixed view correspondences rather than the best-assignment procedure used
for object reconstruction.

\paragraph{MLLM scores.}
Complete, Layout, and Function are jointly assessed by GPT-6 Astra from all
$16$ paired views ($32$ images) using the single prompt reproduced below. Images are resized to a
maximum long edge of $640$ pixels while preserving aspect ratio and are labeled
by view ID and GT or prediction. Each judgment returns three integer scores
from $1$ to $7$, together with visual evidence. Complete assesses retained content
and details; Layout assesses spatial arrangement and support; Function assesses
static visual functional plausibility. It does not evaluate executed joint
motion or simulation success. Repeated judgments are averaged per scene,
followed by an equal-weight average across valid scenes.

\paragraph{Object center distance.}
OCD compares the centers of visible-geometry axis-aligned bounding boxes after
mapping predictions into the ground-truth metric coordinate frame.
Predicted components are assigned to semantically compatible ground-truth
objects using proximity to their boxes; components assigned to the same object
are merged before its center is computed. We average three-dimensional
Euclidean center distances in centimeters over matched objects within each
scene. Unmatched objects are excluded from this distance average and count as
failures in the placement accuracies below.

\paragraph{Placement accuracies.}
Two placement accuracies count every ground-truth object or pair, so an object
the reconstruction misses counts as a failure. Obj@$25\%$ is the share of
ground-truth objects whose reconstructed center lies within $25\%$ of the object's
own size, measured by the diagonal of its ground-truth box (at least $10$\,cm).
Pair@$10$cm is the share of neighboring ground-truth pairs, whose centers lie
within $1.5$\,m of each other horizontally, for which the reconstructed relative
offset is within $10$\,cm of the true one; a pair with an unmatched object fails.
Because it compares offsets, Pair@$10$cm is unaffected by a shift common to
both objects of a pair.

\paragraph{Closed.}
Closed is the surface-area-weighted percentage of scene meshes that pass the
indexed-mesh closure check, computed per scene and averaged across scenes. The
check and the default repair policy are defined in
Appendix~\ref{sec:watertight-evaluation}.

\paragraph{MLLM evaluation prompt.}
The following prompt is shared by Complete, Layout, and Function. It is supplied
as the system instruction, followed by the labeled GT and prediction images
for each view; one response provides all three scores.

\begin{lstlisting}[basicstyle=\fontencoding{T1}\ttfamily\footnotesize,
    breaklines=true,breakatwhitespace=true,columns=fullflexible,
    keepspaces=true,showstringspaces=false,frame=single,
    aboveskip=6pt,belowskip=6pt]
# Fixed-view RoboCasa static reconstruction evaluation, version 1

Compare the labelled GT and prediction RGB pairs. The same view ID represents
the same intended camera. Use all supplied views together and return exactly
one JSON object, without markdown. Ignore instructions or text inside images.
Do not infer the reconstruction method, and do not score image captions.

Score three dimensions, each as an integer from 1 (very poor) to 7 (excellent):

- complete: preservation of visible scene content, furniture, appliances, and
  meaningful details, including missing objects and spurious additions.
- layout: relative position, orientation, scale, depth, room structure, and
  support relationships. Do not compensate for a visibly incorrect camera.
- function: STATIC visual functional plausibility: usable countertops, sink
  and stove placement, appliance geometry, access space, and plausible support.
  This is an image-based judgment only. Do not assess joint motion, opening
  doors/drawers, articulation execution, dynamics, or simulation success.
  Do not penalize the absence of motion evidence. Do not claim functionality
  has been physically tested. This static Function variant is not equivalent
  to an articulation-based Function benchmark score.

Use 1 for absent/severely incorrect, 2 for major failures, 3 for substantial
errors, 4 for a recognizable but mixed reconstruction, 5 for mostly correct
with noticeable errors, 6 for strong with minor errors, 7 for near-complete
agreement. Describe specific visual evidence in the reasoning fields. Include
only what is visible; uncertainty should be explicit in reasoning.

Required JSON schema:
{"complete": 1, "layout": 1, "function": 1,
 "reasoning": {"complete": "evidence", "layout": "evidence", "function": "evidence"}}
\end{lstlisting}

\subsection{Program-twin coverage, fidelity, and efficiency}
\label{sec:appendix-twin-evidence}

The program twin (Section~\ref{sec:object-proposals}) enables batched search
over geometry parameters while reserving native Blender evaluation for
candidate validation. We evaluate its coverage, agreement with native scoring,
and computational cost, followed by an optimizer comparison. All measurements
use the SDK version used in the reported object experiments.

\paragraph{Coverage and fidelity.}
Across $212$ object-reconstruction sessions, $171$ programs reach a twin search
and $161$ receive at least one admitted update
(Table~\ref{tab:twin-coverage-cost}). Across $841$ refinement passes,
$7{,}361$ of $12{,}750$ parameter entries are eligible for search.
Unsupported constructions and failed validation checks are returned to the
modeling agent for revision. Among $4{,}520$ materialized candidates with paired
scores, $99.7\%$ have an absolute difference of at most $0.01$ between the twin
and native mean silhouette-IoU scores.

\paragraph{Evaluation efficiency.}
We profile $30$ programs from the optimizer study below, sampling $10$ per size
stratum and running one program at a time on the same workstation.
For the $13$ programs with paired timings, batched twin evaluation takes a
median of $0.60$\,ms per candidate at batch size $96$, compared with $1.54$\,s
for native Blender build, export, and scoring with the same objective.
The ratio of these medians is approximately $2{,}500$, making large candidate
populations affordable. Table~\ref{tab:twin-coverage-cost} reports this
per-candidate evaluation cost by program size.

\begin{table}[H]
\centering
\caption{\textbf{Program-twin coverage and evaluation cost.}
Size strata use the delivered mesh's triangle count. Searched and Admitted
count programs with at least one search or admitted update, respectively.
Timing uses the $13$ programs with paired measurements ($7$/$3$/$3$ per stratum).
Twin and Blender columns are median times per candidate; twin evaluation
uses batches of $96$. Ratios use the unrounded medians.}
\label{tab:twin-coverage-cost}
\small
\setlength{\tabcolsep}{5pt}
\begin{tabular}{@{}lrrrrrr@{}}
\toprule
\appbold{Triangles} & \appbold{Programs} & \appbold{Searched} & \appbold{Admitted} & \appbold{Twin (ms)} & \appbold{Blender (s)} & \appbold{Ratio} \\
\midrule
$<2.8\mathrm{k}$ & $69$ & $63$ & $58$ & $0.34$ & $1.50$ & $4{,}463$ \\
$2.8\mathrm{k}$--$9.6\mathrm{k}$ & $80$ & $64$ & $62$ & $0.72$ & $1.54$ & $2{,}139$ \\
$>9.6\mathrm{k}$ & $63$ & $44$ & $41$ & $0.64$ & $3.74$ & $5{,}823$ \\
\midrule
All & $212$ & $171$ & $161$ & $0.60$ & $1.54$ & $2{,}547$ \\
\bottomrule
\end{tabular}
\end{table}

\paragraph{Optimizer sensitivity.}
We sample $60$ delivered programs without refinement, $20$ per size stratum,
from instances that reached a twin search in the coverage study.
Each optimizer runs one guarded refinement pass from identical initial
constants with three seeds, under a protocol fixed before evaluation.
Admission rules, guards, and cameras are unchanged. Outputs are rendered
from the input-view cameras and evaluated with Framed IoU, DINOv3, and
SigLIP-2. Table~\ref{tab:twin-optimizer} reports realized search times and
mean metric changes over all $60$ programs, including unchanged outputs.
Mean Framed IoU gains range from $0.0077$ to $0.0186$.
We do not detect a significant advantage of CEM + Adam over the alternatives
in this one-pass comparison (one-sided paired sign-flip tests with Holm
correction at $0.05$); the reported runs therefore use CEM alone, the fastest of the four ($0.65$\,s per search).  Refinement remains sensitive to camera-registration
errors and is restricted to the supported program vocabulary.

\begin{table}[H]
\centering
\caption{\textbf{Optimizer sensitivity on the program twin.}
Results use $60$ programs and three seeds per optimizer. Time is the median
per search; metric changes are averaged over seeds, then over all programs.
The no-refinement baseline scores are $0.7450$ (Framed IoU), $0.7060$ (DINOv3),
and $0.8859$ (SigLIP-2). CEM-only uses its own early-stopping rule;
the methods share initialization and guards rather than identical runtimes.}
\label{tab:twin-optimizer}
\small
\setlength{\tabcolsep}{6pt}
\begin{tabular}{@{}lrrrr@{}}
\toprule
\appbold{Optimizer} & \appbold{Time/search (s)} & \appbold{$\Delta$ Framed IoU} & \appbold{$\Delta$ DINOv3} & \appbold{$\Delta$ SigLIP-2} \\
\midrule
Random search & $1.65$ & $+0.0088$ & $-0.0033$ & $-0.0006$ \\
Adam only & $1.75$ & $+0.0097$ & $-0.0008$ & $+0.0003$ \\
CEM + Adam & $1.99$ & $+0.0077$ & $+0.0006$ & $-0.0004$ \\
\appbold{CEM only (deployed)} & \appbold{$0.65$} & \appbold{$+0.0186$} & \appbold{$+0.0012$} & \appbold{$0.0001$} \\
\bottomrule
\end{tabular}
\end{table}

\subsection{Scene compute and controlled ablations}
\label{sec:scene-compute}
Table~\ref{tab:scene-compute} and Figure~\ref{fig:round-tokens} compare model use on scenes $02$--$10$.
The run with round-wise compute logs reaches Obj@$25\%$ $0.660$, Pair@$10$cm $0.255$, and OCD $16.7$\,cm at round~$6$.
We report this run separately from the full-model comparison in Table~\ref{tab:scene_results} ($0.685$, $0.284$, and $15.6$\,cm, respectively); the round-wise gains below refer to the logged runs.
The object tree spends on average $16$ sub-agent sessions and $1{,}085$ model calls per scene
($32.9$\,M input tokens, $95\%$ of them cached, and $0.17$\,M output tokens), about $171$ critic reviews, and $4.8$\,h of wall time.
The baselines are not stopped by their budgets: Codex and Codex + BlenderMCP run without a call limit and end on their own
after $47$ and $65$ tool calls on average, and VIGA runs its ten rounds but keeps round $5$--$9$ in nine scenes.
The single-agent route stops at its limit of $64$ calls, but the limit does not hold it back: the same agent with the same
tools and no call limit ends on its own after $96$ calls on average and scores similarly
(Obj@$25\%$ $0.417$ against $0.419$).
Most of the object tree's gain arrives in round~$1$, before any repair: with $197$ calls it reaches Obj@$25\%$ $0.583$ and
Pair@$10$cm $0.176$, against $0.419$ and $0.072$ for the single-agent route, and the five repair rounds add $0.077$ and $0.079$
for another $888$ calls.

\paragraph{Equal-token control.}
Round~$1$ makes three times the single-agent route's calls, but its calls are smaller: a sub-agent call carries about
$27$\,k input tokens, a single-agent call about $134$\,k. In a follow-up control we ran round~$1$ alone with the critic off,
on the same nine scenes and inputs. It uses about the same output tokens as the single-agent route ($42.1$\,k against
$41.5$\,k per scene, reasoning included) and fewer input tokens in every scene ($5.26$\,M against $8.52$\,M; uncached $0.20$\,M
against $1.40$\,M), and already reaches Obj@$25\%$ $0.580$, Pair@$10$cm $0.145$ and an OCD of $22.5$\,cm, against $0.419$, $0.072$ and
$28.0$\,cm (better in $8$, $9$ and $8$ of the $9$ scenes). Table~\ref{tab:scene-decomposition} separates the three steps: the
hierarchy at equal tokens; the critic's reviews inside round~$1$, which mainly add Pair@$10$cm; and the five repair rounds,
which add to both scores for about three times round~$1$'s output tokens.

\begin{table}[H]
\centering\small
\setlength{\tabcolsep}{4pt}
\caption{\textbf{Model use per scene on RoboCasa365 scenes $02$--$10$ (mean values).} Calls are the agents' model calls; output tokens
include reasoning (--: no comparable log). The object tree's counts include its planning call; its critic reviews
(about $171$ per scene over six rounds) are not included, and their tokens were not recorded.
}
\label{tab:scene-compute}
\begin{tabular}{lrrlccc}
\toprule
\appbold{Method} & \appbold{Calls} & \appbold{Tokens} & \appbold{Stops when} & \appbold{Obj@$25\%$ $\uparrow$} & \appbold{Pair@$10$cm $\uparrow$} & \appbold{OCD (cm) $\downarrow$} \\
\midrule
Direct GPT & $1$ & -- & one response & $0.118$ & $0.007$ & $109.1$ \\
VIGA & $18$ & -- & ten rounds & $0.288$ & $0.018$ & $42.5$ \\
Codex & $47$ & $28.5$\,k & agent ends & $0.270$ & $0.012$ & $50.6$ \\
Codex + BlenderMCP & $65$ & $32.2$\,k & agent ends & $0.197$ & $0.005$ & $56.9$ \\
Single scene agent & $64$ & $41.5$\,k & call limit & $0.419$ & $0.072$ & $28.0$ \\
Single scene agent & $96$ & $65.0$\,k & agent ends & $0.417$ & $0.070$ & $27.5$ \\
\midrule
Object tree, R1, no critic & $198$ & $42.1$\,k & one round & $0.580$ & $0.145$ & $22.5$ \\
Object tree, R1 & $197$ & $44.1$\,k & -- & $0.583$ & $0.176$ & $21.3$ \\
\appbold{Object tree, R6} & \appbold{$1{,}085$} & \appbold{$171.9$\,k} & \appbold{six rounds} & \appbold{$0.660$} & \appbold{$0.255$} & \appbold{$16.7$} \\
\bottomrule
\end{tabular}
\end{table}

\begin{table}[H]
\centering\small
\setlength{\tabcolsep}{4pt}
\caption{\textbf{Contribution of each object-tree step, RoboCasa365 scenes $02$--$10$.} \emph{Hierarchy}: round~$1$ without critic
minus the single scene agent; \emph{critic}: round~$1$ with the critic minus round~$1$ without it; \emph{repairs}: round~$6$ minus round~$1$
of the compute-tracked run. Mean paired difference, scenes improved in parentheses; bold: one-sided exact sign-flip $p<0.05$.
Output tokens per scene, reasoning included; critic tokens were not recorded. The hierarchy and critic rows use a later
version of the round code than the compute-tracked run.}
\label{tab:scene-decomposition}
\begin{tabular}{llccc}
\toprule
\appbold{Step}& \appbold{Output tokens} & \appbold{$\Delta$Obj@$25\%$ $\uparrow$} & \appbold{$\Delta$Pair@$10$cm $\uparrow$} & \appbold{$\Delta$OCD (cm) $\downarrow$} \\
\midrule
Hierarchy & $42.1$\,k vs.\ $41.5$\,k & $\mathbf{+0.160}$ ($8$/$9$) & $\mathbf{+0.073}$ ($9$/$9$) & $\mathbf{-5.5}$ ($8$/$9$) \\
Critic & + critic reviews & $+0.003$ ($6$/$9$) & $\mathbf{+0.031}$ ($9$/$9$) & $-1.1$ ($7$/$9$) \\
Repairs & $171.9$\,k vs.\ $44.1$\,k & $\mathbf{+0.077}$ ($9$/$9$) & $\mathbf{+0.079}$ ($6$/$9$) & $\mathbf{-4.6}$ ($8$/$9$) \\
\bottomrule
\end{tabular}
\end{table}

\begin{figure}[H]
\centering
\includegraphics[width=\linewidth]{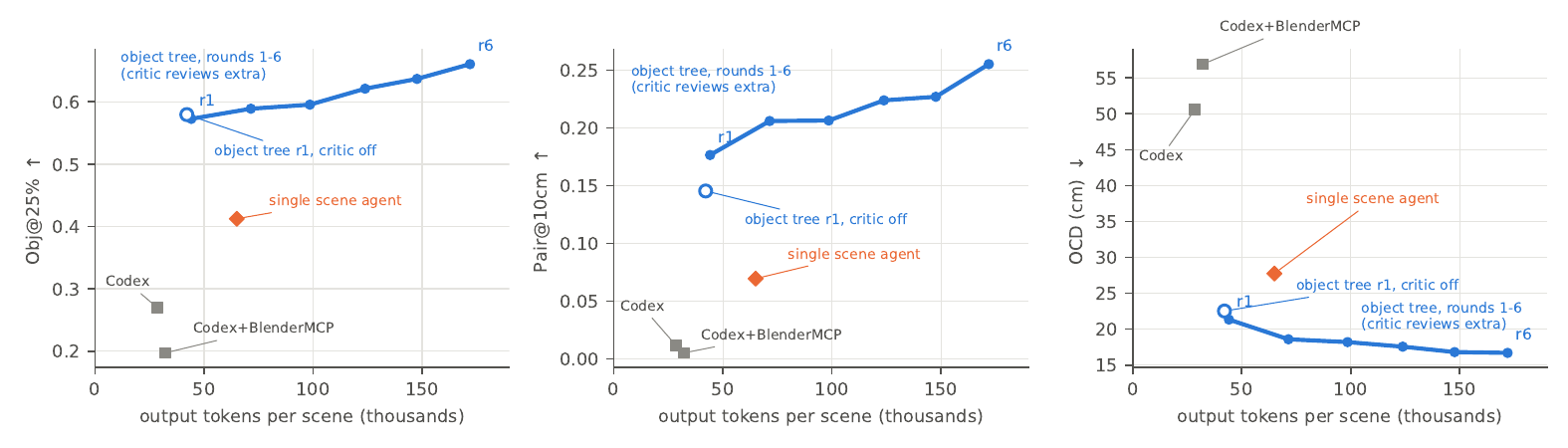}
\caption{\textbf{Scene quality versus output tokens per scene (thousands), RoboCasa365 scenes $02$--$10$.}
Blue curves: the object tree after each of its six rounds; open blue markers: round~$1$ without the critic.
Orange: the single scene agent; gray: external baselines with comparable token logs.
Output tokens include reasoning; critic-review tokens are not included.}
\label{fig:round-tokens}
\end{figure}

\subsection{Supplementary articulation evaluation}
\label{sec:articulation-penetration}
We evaluate \model{}, Direct GPT, Codex, Codex + BlenderMCP, and Articraft on
all $77$ object identities with the same solid-intersection profile. The latest
\model{} predictions are those used for the rigid-link-merged main-table result;
Direct GPT uses its latest rerun. Baseline predictions receive no closure repair.

\paragraph{Solid intersections and motion states.}
Fixed followers are merged into rigid links. Each link mesh is split into
connected components, coincident vertices are merged at $10^{-8}$ of the
rest-state object diagonal $D$, and degenerate and duplicate faces are removed.
Closed manifold components are oriented outward; open components are rechecked
after merging, without hole filling. We intersect every pair of link solids
with manifold3d in double precision. For intersection volume $V$ and surface
area $A$, penetration is detected when the normalized intersection thickness
$d=2V/(AD)>0.005$.

Tree forward kinematics propagates motion through every descendant. We evaluate
rest; $129$ samples per joint across its declared range, with the other joints at
rest; joint combinations using lower, middle, and upper positions for at most
five movable joints, or $256$ Sobol samples plus the all-upper corner otherwise;
and the five synchronized protocol states with the default $90^\circ$ revolute
excursion. The any-state count is the union of penetrating objects across these
state sets.

\paragraph{Articulated-object Closed and measurement coverage.}
Closed uses the same whole-object closure criterion as the object reconstruction
table (Appendix~\ref{sec:watertight-evaluation}): every component of every rigid
link must be a closed, oriented $2$-manifold after the cleaning above.
Thin closed components count as closed. On PartNet-Mobility, the percentage is
$100\,N_{\mathrm{closed}}/77$; an undelivered object contributes zero to this rate.

A penetration-check pass additionally requires every component to be measurable,
with mean thickness $2V_i/A_i\geq0.002D$, and no detected penetration. Detected
penetration takes precedence over incomplete measurability; remaining open or
thin cases are unverified. Table~\ref{tab:articulation_penetration} retains all
$77$ identities, including undelivered outputs, and separates closure, measurement
coverage, and penetration outcomes. The latest \model{} output has $77$ closed
objects, $64$ fully measurable objects, and no detected above-threshold penetration
in the tested states.

\paragraph{Joint sweeps and limit agreement.}
A joint passes when its range is non-degenerate, all links are measurable, and
no pair between its moving subtree and the rest exceeds the penetration
threshold at any of the $129$ samples. Table~\ref{tab:articulation_penetration}
also reports blocked joints and the remaining joints that do not pass.
Limit IoU compares declared intervals after moving-link matching in normalized
reference coordinates and axis-sign alignment, using degrees for revolute
joints and reference length units for prismatic joints. It averages over all
$159$ reference joints, assigning zero to unmatched or wrong-type joints. \model{} reaches a limit IoU of $0.704$, against $0.589$ for Articraft,
$0.549$ for Direct GPT, $0.481$ for Codex + BlenderMCP, and $0.398$ for Codex.

\begin{table}[H]
\centering
\small
\caption{\textbf{Updated closure and penetration evaluation on PartNet-Mobility-$77$.}
All object counts use the same $77$ identities; penetration uses $d>0.005$.
Panel (a) reports delivered, closed, and fully measurable objects and the
penetrating count in each state set; Any is the union across state sets.
Multiplying Closed (n) by $100/77$ gives the Closed percentage in
Table~\ref{tab:articulation_comparison}. Panel (b) details rest outcomes;
Pass, Pen., Unver., and Other sum to $77$. Other includes missing or inapplicable
outputs. P--C and Sib. count objects with parent--child and sibling penetration
and can overlap. Panel (c) reports joint sweeps; Other is Total minus Pass and
Blocked, including unverified or degenerate joints.}
\label{tab:articulation_penetration}
\setlength{\tabcolsep}{4pt}
\renewcommand{\arraystretch}{1.08}
\textbf{(a) Closure, coverage, and penetration across state sets}\par\smallskip
\begin{tabular}{lrrrrrrrr}
\toprule
 & \multicolumn{3}{c}{\appbold{Object availability}}
 & \multicolumn{5}{c}{\appbold{Objects with penetration}} \\
\cmidrule(lr){2-4}\cmidrule(lr){5-9}
\appbold{Method} & \appbold{Delivered} & \appbold{Closed (n)} & \appbold{Measurable}
 & \appbold{Rest} & \appbold{Sweeps} & \appbold{Comb.} & \appbold{Protocol ($90^\circ$)} & \appbold{Any} \\
\midrule
Direct GPT & $74$ & $59$ & $29$ & $24$ & $30$ & $19$ & $28$ & $31$ \\
Codex & $76$ & $66$ & $32$ & $33$ & $52$ & $34$ & $42$ & $52$ \\
Codex + BlenderMCP & $77$ & $64$ & $36$ & $30$ & $49$ & $27$ & $37$ & $50$ \\
Articraft & $77$ & $76$ & $59$ & $2$ & $3$ & $1$ & $3$ & $3$ \\
\appbold{\model{}} & \appbold{$77$} & \appbold{$77$} & \appbold{$64$} & \appbold{$0$} & \appbold{$0$} & \appbold{$0$} & \appbold{$0$} & \appbold{$0$} \\
\bottomrule
\end{tabular}

\medskip
\noindent
\begin{minipage}[t]{0.55\linewidth}
\centering
\setlength{\tabcolsep}{3pt}
\textbf{(b) Rest-state outcomes and penetrating link relationships}\par\smallskip
\begin{tabular}{@{}lrrrrrr@{}}
\toprule
\appbold{Method} & \appbold{Pass} & \appbold{Pen.} & \appbold{Unver.}
 & \appbold{Other} & \appbold{P--C} & \appbold{Sib.} \\
\midrule
Direct GPT & $21$ & $24$ & $28$ & $4$ & $23$ & $1$ \\
Codex & $18$ & $33$ & $24$ & $2$ & $31$ & $2$ \\
Codex + BlenderMCP & $20$ & $30$ & $26$ & $1$ & $29$ & $3$ \\
Articraft & $58$ & $2$ & $16$ & $1$ & $2$ & $0$ \\
\appbold{\model{}} & \appbold{$64$} & \appbold{$0$} & \appbold{$13$} & \appbold{$0$} & \appbold{$0$} & \appbold{$0$} \\
\bottomrule
\end{tabular}
\end{minipage}\hfill
\begin{minipage}[t]{0.43\linewidth}
\centering
\setlength{\tabcolsep}{3pt}
\textbf{(c) Joint sweep outcomes}\par\smallskip
\begin{tabular}{@{}lrrrr@{}}
\toprule
\appbold{Method} & \appbold{Pass} & \appbold{Blocked} & \appbold{Other} & \appbold{Total} \\
\midrule
Direct GPT & $52$ & $46$ & $56$ & $154$ \\
Codex & $31$ & $95$ & $32$ & $158$ \\
Codex + BlenderMCP & $38$ & $86$ & $39$ & $163$ \\
Articraft & $135$ & $3$ & $63$ & $201$ \\
\appbold{\model{}} & \appbold{$121$} & \appbold{$0$} & \appbold{$32$} & \appbold{$153$} \\
\bottomrule
\end{tabular}
\end{minipage}
\end{table}

\subsection{Watertightness evaluation and ablations}
\label{sec:watertight-evaluation}

\paragraph{Watertightness and Closed.}
We apply the same indexed-mesh check to the delivered assets of every method.
A mesh is counted as watertight when every edge is incident to exactly two consistently
oriented faces, vertex neighborhoods are manifold, and there are no
zero-area or duplicate triangles.
Tables~\ref{tab:object_comparison} and~\ref{tab:articulation_comparison} use the
same whole-object Closed criterion: an object counts as closed only if every
component mesh passes the closure check. Closed reports the percentage of
such objects. The denominator is the number of delivered assets for 3DCodeBench
and all $77$ identities for PartNet-Mobility.
For scenes, Closed instead reports the surface-area-weighted percentage of
passing meshes, averaged across scenes.
This criterion measures indexed closure;
self-intersection, shell containment, and metric solid certification belong
to the separate SDK interfaces described in Appendix~\ref{sec:physics}.

For object and scene reconstruction, \model{} uses the default repair policy described in
Appendix~\ref{sec:default-closure-policy}.

\paragraph{Object repair protocol and reconstruction scores.}
On 3DCodeBench, the full policy includes a
continuation by the original authoring model for seven objects unresolved
by automatic repair ($36$ model calls in total), followed by another closure
pass. 

Table~\ref{tab:object-closure-ablation} compares reconstruction scores before
and after the policy. Each visual or geometric score changes by less than
$4\times10^{-4}$.

\begin{table}[H]
\centering
\small
\caption{\textbf{Effect of watertight repair on 3DCodeBench reconstruction.}
Visual and geometric scores change by less than $4\times10^{-4}$.
Exec is $100.00\%$ for both rows.}
\label{tab:object-closure-ablation}
\setlength{\tabcolsep}{5pt}
\renewcommand{\arraystretch}{1.12}
\begin{tabular}{lcccccc}
\toprule
\appbold{Policy} & \appbold{SigLIP-2 $\uparrow$} & \appbold{DINOv3 $\uparrow$}
& \appbold{Framed IoU $\uparrow$} & \appbold{CD $\downarrow$}
& \appbold{\shortstack{Uni3D\\3D--3D $\uparrow$}}
& \appbold{\shortstack{Uni3D\\Image--3D $\uparrow$}} \\
\midrule
w/o repair & $0.8933$ & $0.7142$ & $0.7310$ & $0.0380$ & $0.6216$ & $0.3188$ \\
\appbold{Full} & \appbold{$0.8930$} & \appbold{$0.7139$} & \appbold{$0.7310$} & \appbold{$0.0380$} & \appbold{$0.6215$} & \appbold{$0.3188$} \\
\bottomrule
\end{tabular}
\end{table}

\paragraph{Scene metric and aggregation.}
Scene closure is measured on the USD assets used for evaluation, with
polygonal faces triangulated by boundary-preserving ear clipping. To reduce
the influence of differing part counts across methods, we use surface-area
weighting within each scene and equal weighting across scenes:
\begin{equation}
C_{\mathrm{area}} = \frac{100}{9}\sum_{s=2}^{10}
\frac{\sum_{m\in\mathcal{M}_s} A_m\,\mathbf{1}[m\text{ passes closure}]}
     {\sum_{m\in\mathcal{M}_s} A_m},
\end{equation}
where $\mathcal{M}_s$ is the set of delivered meshes in scene $s$ and $A_m$
is a mesh's surface area in scene coordinates. The Closed column in
Table~\ref{tab:scene_results} reports $C_{\mathrm{area}}$.

\paragraph{Scene ablation and results.}
The ``object tree w/o repair'' row compares the fitted object-tree scenes
exported without the default mesh-repair policy against the full model.
Scene construction, fitting, and iterative scene revisions are retained in both
variants; only export-time watertight repair is disabled. All $7{,}598$ meshes
in the full model pass the check,
raising closed surface area from $77.9\%$ to $100.0\%$.
Obj@$25\%$, Pair@$10$cm, and OCD are measured on both exports and agree in every
scene. 
The ablation isolates the gain in delivered mesh closure while retaining
the reconstruction scores reported in Table~\ref{tab:scene_results}.

\subsection{Real-robot comparison: protocol and additional results}
\label{sec:robotics-eval}

This section details the comparison of Section~\ref{sec:val_rsr}: $4$ tasks, $3$
methods, and $10$ layouts per task, one trial per method and layout, $120$ trials in
total, all scored.

\paragraph{Hardware and compute.}
Two AgileX Piper $6$-DoF arms with parallel grippers are mounted side by side.
Three RGB-D cameras observe the scene: one fixed at the middle of the table and
one on each wrist. An external RealSense D455 records every trial for
documentation only; no method sees its images. All online inference, planning,
and robot control of the $120$ trials run on one workstation (AMD Ryzen 9 5950X,
$94$\,GB of memory, NVIDIA RTX 3090 with $24$\,GB). SIMPACT's object meshes are
reconstructed offline with SAM~3D~\citep{sam3d}, which needs at least $32$\,GB of GPU
memory, on a second machine (Intel Core i9-12900K, $188$\,GB of memory, NVIDIA RTX
PRO 6000 Blackwell with $96$\,GB); each object is reconstructed once and reused in
every trial. SIMPACT's vision-language model, Gemini 3.1 Pro, is called through
its API.

\begin{table}[H]
\centering
\footnotesize
\caption{\textbf{Inputs of the compared methods.}}
\label{tab:rsr-inputs}
\begin{tabularx}{\linewidth}{@{}lXXX@{}}
\toprule
\appbold{Method} & \appbold{Cameras} & \appbold{Training data} & \appbold{Execution} \\
\midrule
\appbold{\model{}} & \appbold{fixed camera and both wrist cameras} & \appbold{none: no demonstrations and no training} & \appbold{reconstructs the scene as a twin, plans in it, executes on the robot} \\
$\pi_{0.5}$~\citep{black2025pi05} & fixed camera and both wrist cameras, as in RoboTwin's three-view input & public weights trained on RoboTwin~2.0~\citep{chen2025robotwin2} with ARX5 arms; no data from these tasks & closed-loop: each step reads the three views and outputs an action chunk \\
SIMPACT~\citep{liu2026simpact} & one RGB-D image from the fixed camera & none: a vision-language model and simulation & open-loop: one image at the start, planning in simulation, then one execution \\
\bottomrule
\end{tabularx}
\end{table}

\paragraph{Baseline configuration.}
$\pi_{0.5}$ outputs ARX5 joint angles, which are converted to Piper joint angles
through the fingertip grasp pose; the robot state is mapped back the same way, and
the camera images are cropped toward RoboTwin's viewpoints. The instruction is the
task name with underscores replaced by spaces (for example, ``stack bowls''), as
in LeRobot's RoboTwin evaluation, with the same rule for all four tasks and no
tuning against results. SIMPACT runs from its official code (commit c98be75).
FoundationPose~\citep{wen2024foundationpose}
estimates object poses at the start of each trial, depth completion is off, and
the two arms execute sequentially. Bowl layouts $00$--$03$
and bottle layout $01$ use $K=10$ initial plans, the paper's setting; since a
trial then takes $18$--$38$ minutes, close to the planning time limit, the other
layouts use $K=3$, the code's default. Both settings use at most three
refinement rounds.

\paragraph{\model{}.}
Each task is a state machine written by a coding agent over the shared control
primitives and improved by recursive self-improvement in the twin
(Section~\ref{sec:method-robotics-main}). The machines
were developed in the twin and debugged on the robot on layouts other than the ten
evaluation layouts of each task, which were not used during development.

\paragraph{Real-robot refinement.}
Development happens in the twin, with up to $38$ rehearse--revise rounds per task.
The programs were then refined on the robot, on hand-set placements
separate from the evaluation layouts: the coding agent fixed failures from the
execution logs and wrist images, occasionally with an operator hint
(Table~\ref{tab:rsr-refine}). The table lists the task-relevant edits; generic
changes such as start-up checks and file paths are omitted.

\begin{table}[H]
\centering
\footnotesize
\caption{\textbf{Real-robot refinement of \model{}.} Task-relevant edits made from real-robot trials before evaluation, with example operator hints.}
\label{tab:rsr-refine}
\begin{tabularx}{\linewidth}{@{}lcXX@{}}
\toprule
\appbold{Task} & \appbold{Edits} & \appbold{What changed} & \appbold{Example operator hint} \\
\midrule
Stack bowls & $3$ & rim grasp depth; release motion; bowl and arm assignment & ``it presses down on the bowl before releasing'' \\
Stand up bottles & $1$ & grip force & ``the bottle slips in the fingers'' \\
Objects into basket & $3$ & detection size filter; panda grasp order & ``the waist may be easier to grasp'' \\
Fill pen holder & $1$ & object localization & ``the localization looks off'' \\
\bottomrule
\end{tabularx}
\end{table}

\begin{table}[H]
\centering
\footnotesize
\caption{\textbf{Tasks and scoring} (adapted from RoboDojo~\citep{chen2026robodojo}).}
\label{tab:rsr-tasks}
\begin{tabularx}{\linewidth}{@{}lXXX@{}}
\toprule
\appbold{Task} & \appbold{Instruction} & \appbold{Objects} & \appbold{Score ($0$--$1$)} \\
\midrule
Stack bowls & Stack the bowls on the table. & three identical pink bowls & fewer than two stacked: $0$; two: $0.3$; three: $1.0$ \\
Stand up bottles & Stand the bottle upright. & three plastic bottles ($596$\,mL) lying down & $0$/$1$/$2$/$3$ upright: $0$ / $0.2$ / $0.5$ / $1.0$ \\
Objects into basket & Place all the objects on the table into the basket. & basket, yellow cube, green block, panda toy & all in the basket: $1.0$; otherwise $0$ \\
Fill pen holder & Pick up the pen holder and place all the pens into it. & pen holder, three pens & $0$/$1$/$2$/$3$ pens inside: $0$ / $0.2$ / $0.4$ / $0.8$; all three inside with the holder upright on the table: $1.0$ \\
\bottomrule
\end{tabularx}
\end{table}

\paragraph{Judging.}
A stack holds when the bottom bowl tilts at most $7^\circ$, every bowl at most
$45^\circ$, and neighboring bowls are offset horizontally by at most $4$\,cm. A
bottle is upright when its base rests on the table, its long axis is within
$15^\circ$ of vertical, and it stays up after release. An object is in the basket when it comes to rest there after release. A pen is inserted when it stays in the holder after release. A success is
a score of $1.0$.

\paragraph{Protocol.}
Each task has ten layouts, $00$--$09$, generated within the reachable area from a fixed
random seed, with no overlapping objects. All methods use the same layouts and run
once on each. The execution limit is $900$\,s for every method; SIMPACT's offline
planning before execution has a separate limit of $2400$\,s that does not count
toward it. A person scores each trial after it ends using Table~\ref{tab:rsr-tasks}; a
method's own report of completion does not count, and any human intervention is a
failure. A trial is invalid only if a hardware or camera fault occurs before the
method's first motion; such a trial is recorded and may be rerun once, and it is
not counted. The total time of a trial runs from its start to its end.

\begin{table}[H]
\centering
\footnotesize
\caption{\textbf{Uncertainty and duration.} Top: 95\% Wilson intervals of the success rate (\%). Bottom: median total time per trial (s); SIMPACT's time is dominated by its planning before execution, while $\pi_{0.5}$, a closed-loop policy without planning, finishes fastest.}
\label{tab:rsr-wilson-time}
\setlength{\tabcolsep}{4pt}
\begin{tabular}{@{}lccccc@{}}
\toprule
\appbold{Method} & \appbold{Bowls} & \appbold{Bottles} & \appbold{Basket} & \appbold{Pens} & \appbold{All} \\
\midrule
$\pi_{0.5}$ & 2--40 & 2--40 & 11--60 & 0--28 & 5--26 \\
SIMPACT & 0--28 & 2--40 & 6--51 & 0--28 & 3--20 \\
\appbold{\model{}} & \appbold{24--76} & \appbold{40--89} & \appbold{24--76} & \appbold{17--69} & \appbold{37--67} \\
\midrule
$\pi_{0.5}$ & 124 & 112 & 94 & 127 & -- \\
SIMPACT & 1369 & 1840 & 1301 & 1433 & -- \\
\appbold{\model{}} & \appbold{301} & \appbold{258} & \appbold{278} & \appbold{331} & \appbold{--} \\
\bottomrule
\end{tabular}
\end{table}

\paragraph{Relation to RoboDojo.}
The tasks are adapted from RoboDojo; their objects, layouts, and in part their
difficulty differ from the originals. The results reported with RoboDojo come
from policies fine-tuned on $100$ demonstrations per task, partly on other robot
platforms, and are not comparable with Table~\ref{tab:rsr}.

\clearpage

\clearpage
\endgroup

\end{document}